\documentclass{article}

\usepackage{iclr2023_conference, times}
\iclrfinaltrue
\pdfoutput=1

\usepackage[utf8]{inputenc} % allow utf-8 input
\usepackage[T1]{fontenc}    % use 8-bit T1 fonts
\usepackage{hyperref}       % hyperlinks
\usepackage{url}            % simple URL typesetting
\usepackage{booktabs}      % professional-quality tables
\usepackage{amsfonts}       % blackboard math symbols
\usepackage{nicefrac}       % compact symbols for 1/2, etc.
\usepackage{microtype}      % microtypography
\usepackage{xcolor}         % colors
\usepackage{textcomp}

\usepackage{hyperref}
\usepackage{tabulary}
\usepackage{amsmath}
\usepackage{bbm}
\usepackage{amsthm}
\usepackage{amssymb}
\usepackage{mathtools}
\usepackage{xspace}
\usepackage{enumitem}
\usepackage[hang,flushmargin]{footmisc}
\usepackage{algorithm}% http://ctan.org/pkg/algorithm
\usepackage{algpseudocode}% http://ctan.org/pkg/algorithmicx
\usepackage{subfigure}
\setlist{leftmargin=*}
\setlist[itemize]{noitemsep}
\setlist[enumerate]{noitemsep}
\usepackage{tikz-network}
\usepackage{multirow}
\definecolor{falured}{rgb}{0.7, 0.15, 0.15}

\newcommand\accfail[1]{{\color{falured}#1}}

\newcommand\fontred[1]{{\color{red}#1}}
\newcommand{\methodname}{KCal\xspace}
\newcommand{\calibrationset}{\mathcal{S}_{\text{cal}}} 
\newcommand{\trainingset}{\mathcal{S}_{\text{train}}} 
\newcommand{\embedder}{\mathbf{f}}
\newcommand{\phat}{\hat{\mathbf{p}}}
\newcommand{\ptrue}{\mathbf{p}}
\newcommand{\defeq}{\vcentcolon=}
\newcommand{\proj}{\mathbf{\Pi}}

\DeclareMathOperator*{\argmax}{arg\,max}

\newcommand{\baselineUnCal}{\texttt{UnCal}\xspace}
\newcommand{\baselineTS}{\texttt{TS}\xspace}
\newcommand{\baselineSpline}{\texttt{Spline}\xspace}
\newcommand{\baselineDirCal}{\texttt{DirCal}\xspace}
\newcommand{\baselineIMax}{\texttt{I-Max}\xspace}
\newcommand{\baselineIOP}{\texttt{IOP}\xspace}
\newcommand{\baselineGP}{\texttt{GP}\xspace}
\newcommand{\baselineMMCE}{\texttt{MMCE}\xspace}
\newcommand{\baselineFocal}{\texttt{Focal}\xspace}

\newtheorem{theorem}{Theorem}[section]

\newtheorem{lemma}[theorem]{Lemma}
\newtheorem{definition}{Definition}

\usepackage{wrapfig}

\def \TableData{
\begin{table}[ht]
\vspace{-0.2in} 
\caption{
Dataset summary: Splits and number of classes ($K$).
}
\label{tab:main:data}
\begin{center}
\begin{footnotesize}

\scalebox{0.8}{
\begin{tabular}{l|ccccc|ccr}
\toprule
Dataset & IIIC & IIIC(pat) & ISRUC & ISRUC(pat) & PN2017 & C10 & C100 & SVHN\\
\midrule
Train & 103,818 & 1,936 & 61,841 & 69 & 15,087 & 45,000, & 45,000 & 65,931\\
Calibration & 1,787 & 77 & 1,372 & 6 & 253 & 5,000 & 5,000 & 7,326\\
Test & 33,953 & 684 & 26,070 & 24 & 4,813 & 10,000 & 10,000 & 26,032\\
$K$ & 6 & 6 & 5 & 5 & 5 & 10 & 100 & 10\\
\bottomrule
\end{tabular}
}
\end{footnotesize}
\end{center}
\vskip -0.2in
\end{table}
}

\def \AppendixTabHealthData{
\begin{table}[ht]
\caption{
Additional information about the healthcare datasets used in this paper.
}
\label{appendix:tab:main:healthdata}
\begin{center}
\begin{footnotesize}
\scalebox{1.0}{
\begin{tabular}{l|ccc|ccc|ccr}
\toprule
& \multicolumn{3}{c}{IIIC}& \multicolumn{3}{c}{ISRUC}& \multicolumn{3}{c}{PN2017}\\
Dataset & Name &  Train & Cal+Test & Name & Train & Cal+Test & Name & Train & Cal+Test\\
\midrule
Class 0 & Other & 42228 & 6852 & Wake & 14325 & 6433 & Normal & 8877 & 2893 \\
Class 1 & Seizure & 3305 & 549 & N1 & 7589 & 3798 & Other & 4524 & 1579\\
Class 2 & LPD & 17338 & 7589 & N2 & 19501 & 8505 & AF & 1345 & 449\\
Class 3 & GPD & 16983 & 9737 & N3 & 12012 & 5254 & Noisy & 341 & 145 \\
Class 4 & LRDA & 12515 & 5946 & REM & 8414 & 3452 & \textendash & \textendash & \textendash\\
Class 5 & GRDA & 11449 & 5067 & \textendash & \textendash & \textendash & \textendash & \textendash & \textendash\\
\bottomrule
\end{tabular}
}
\end{footnotesize}
\end{center}
\end{table}
}

\def \TableExpRanks{
\begin{table}[ht]
\caption{
Ranks for different evaluation metrics.
The best rank is \underline{underscored}.
In general, \methodname consistently outperforms baselines on Accuracy, CECE and Brier, and the difference between most methods on ECE is small. 
}
\label{tab:main:ranks}
\centering
\begin{small}
\scalebox{0.75}{
\begin{tabular}{p{0.14\textwidth}p{0.1\textwidth}|p{0.1\textwidth}p{0.1\textwidth}p{0.1\textwidth}p{0.1\textwidth}p{0.1\textwidth}p{0.1\textwidth}p{0.1\textwidth}|p{0.1\textwidth}}
\toprule
Ranking & \baselineUnCal & \baselineTS & \baselineDirCal & \baselineIMax & \baselineFocal & \baselineSpline & \baselineIOP & \baselineGP & \methodname\\ 
\midrule
Accuracy & 4.74$\pm$1.18 & 4.74$\pm$1.18 & 3.74$\pm$2.68 & 5.83$\pm$2.56 & 8.99$\pm$0.10 & 4.97$\pm$1.56 & 4.71$\pm$1.19 & 5.37$\pm$1.68 & \underline{1.91$\pm$1.98}\\
CECE & 6.04$\pm$1.96 & 6.84$\pm$1.49 & 2.48$\pm$1.52 & 6.93$\pm$1.93 & 8.27$\pm$1.14 & 4.51$\pm$1.73 & 4.45$\pm$1.57 & 3.73$\pm$1.43 & \underline{1.75$\pm$1.17}\\
Brier & 7.57$\pm$1.09 & 6.27$\pm$1.15 & 3.01$\pm$1.68 & 6.69$\pm$1.12 & 8.09$\pm$2.30 & 3.92$\pm$0.92 & 4.50$\pm$1.43 & 3.52$\pm$1.83 & \underline{1.43$\pm$1.02}\\
ECE & 7.68$\pm$1.22 & 5.87$\pm$1.54 & 2.68$\pm$1.39 & 6.95$\pm$1.37 & 8.43$\pm$0.93 & 3.27$\pm$1.40 & 3.70$\pm$1.83 & \underline{2.65$\pm$1.76} & 3.77$\pm$1.64\\
\bottomrule
\end{tabular}
}
\end{small}
\end{table}
}

\def \TableExpMainAcc{
\begin{table}[ht]
\caption{Accuracy in \% ($\uparrow$ means higher=better).
Accuracy numbers lower than the uncalibrated predictions are in \accfail{dark red} and the best are in \textbf{bold} (both at p=0.01).
\methodname always surpasses or maintains the accuracy.
}
\label{tab:main:acc}
\centering
\begin{footnotesize}
\scalebox{0.7}{
\begin{tabular}{p{0.14\textwidth}c|ccccccc|r}
\toprule
  Accuracy $\uparrow$  & \baselineUnCal & \baselineTS & \baselineDirCal & \baselineIMax & \baselineFocal & \baselineSpline & \baselineIOP & \baselineGP & \methodname\\ 
\midrule
IIIC (pat) & 58.68$\pm$1.42 & 58.68$\pm$1.42 & \textbf{63.17$\pm$1.42} & 57.20$\pm$1.32 & \accfail{54.35$\pm$1.64} & 58.51$\pm$1.32 & 58.68$\pm$1.42 & 58.68$\pm$1.42 & \textbf{61.67$\pm$2.22}\\
IIIC & 58.53$\pm$0.06 & 58.53$\pm$0.06 & 63.80$\pm$0.10 & \accfail{56.96$\pm$0.14} & \accfail{54.41$\pm$0.05} & \accfail{58.36$\pm$0.20} & 58.53$\pm$0.06 & 58.52$\pm$0.06 & \textbf{66.32$\pm$0.21}\\
ISRUC (pat) & \textbf{75.11$\pm$0.77} & \textbf{75.11$\pm$0.77} & \textbf{75.57$\pm$0.91} & \textbf{75.54$\pm$0.68} & \accfail{73.79$\pm$0.72} & \textbf{75.11$\pm$0.79} & \textbf{75.11$\pm$0.77} & \textbf{75.11$\pm$0.76} & \textbf{76.13$\pm$0.89}\\
ISRUC & 74.66$\pm$0.08 & 74.66$\pm$0.08 & 76.08$\pm$0.16 & 75.15$\pm$0.07 & \accfail{73.34$\pm$0.09} & 74.69$\pm$0.09 & 74.66$\pm$0.08 & 74.66$\pm$0.09 & \textbf{77.45$\pm$0.16}\\
PN2017 & 54.67$\pm$0.14 & 54.67$\pm$0.14 & \textbf{60.00$\pm$0.22} & 57.55$\pm$0.39 & \accfail{13.78$\pm$0.13} & 55.11$\pm$0.84 & 55.15$\pm$1.48 & 54.69$\pm$0.15 & \textbf{60.36$\pm$0.61}\\
\midrule
C10 (ViT) & \textbf{98.94$\pm$0.05} & \textbf{98.94$\pm$0.05} & \textbf{98.94$\pm$0.05} & \textbf{98.94$\pm$0.05} & \accfail{98.76$\pm$0.06} & \textbf{98.94$\pm$0.05} & \textbf{98.94$\pm$0.05} & \textbf{98.94$\pm$0.06} & \textbf{98.98$\pm$0.09}\\
C10 (Mixer) & \textbf{98.17$\pm$0.08} & \textbf{98.17$\pm$0.08} & \accfail{98.03$\pm$0.09} & \textbf{98.13$\pm$0.08} & \accfail{96.98$\pm$0.08} & \textbf{98.17$\pm$0.08} & \textbf{98.17$\pm$0.08} & \textbf{98.16$\pm$0.08} & \textbf{98.14$\pm$0.06}\\
C100 (ViT) & 92.09$\pm$0.16 & 92.09$\pm$0.16 & 92.08$\pm$0.14 & 91.95$\pm$0.17 & \accfail{91.21$\pm$0.12} & 92.09$\pm$0.16 & 92.09$\pm$0.16 & 92.09$\pm$0.16 & \textbf{92.37$\pm$0.15}\\
C100 (Mixer) & \textbf{87.53$\pm$0.20} & \textbf{87.53$\pm$0.20} & \accfail{87.24$\pm$0.22} & \accfail{87.10$\pm$0.21} & \accfail{86.49$\pm$0.23} & \textbf{87.53$\pm$0.20} & \textbf{87.53$\pm$0.20} & \textbf{87.51$\pm$0.20} & \textbf{87.55$\pm$0.16}\\
SVHN (ViT) & 95.93$\pm$0.05 & 95.93$\pm$0.05 & 95.93$\pm$0.05 & \accfail{95.85$\pm$0.06} & \accfail{95.70$\pm$0.08} & 95.93$\pm$0.05 & 95.93$\pm$0.05 & 95.93$\pm$0.05 & \textbf{96.42$\pm$0.05}\\
SVHN (Mixer) & 95.85$\pm$0.04 & 95.85$\pm$0.04 & 95.98$\pm$0.04 & 95.85$\pm$0.05 & \accfail{95.24$\pm$0.04} & 95.85$\pm$0.04 & 95.85$\pm$0.04 & 95.85$\pm$0.05 & \textbf{96.10$\pm$0.04}\\
\bottomrule
\end{tabular}
}
\end{footnotesize}
\end{table}
}

\def \TableExpMainCECEt{
\begin{table}[ht]
\caption{Class-wise ECE in $10^{-2}$ ($\downarrow$ means lower=better).
The best accuracy-preserving method is in \textbf{bold} (p=0.01).
\methodname almost always achieves the lowest class-wise ECE, while maintaining accuracy.
}
\label{tab:main:cecer}
\centering
\begin{small}
\scalebox{0.75}{
\begin{tabular}{p{0.14\textwidth}p{0.1\textwidth}|p{0.1\textwidth}p{0.1\textwidth}p{0.1\textwidth}p{0.1\textwidth}p{0.1\textwidth}p{0.1\textwidth}p{0.1\textwidth}|p{0.1\textwidth}}
\toprule
  CECE $\downarrow$  & \baselineUnCal & \baselineTS & \baselineDirCal & \baselineIMax & \baselineFocal & \baselineSpline & \baselineIOP & \baselineGP & \methodname\\ 
\midrule
IIIC (pat) & 7.94$\pm$0.27 & 8.93$\pm$0.85 & \textbf{5.08$\pm$1.49} & 9.17$\pm$0.99 & 8.95$\pm$0.51 & 8.52$\pm$0.63 & 8.30$\pm$0.54 & 7.92$\pm$0.67 & \textbf{4.60$\pm$1.28}\\
IIIC & 7.84$\pm$0.02 & 8.96$\pm$0.49 & \textbf{2.12$\pm$0.12} & 8.77$\pm$0.24 & 8.77$\pm$0.02 & 8.40$\pm$0.23 & 7.98$\pm$0.26 & 7.48$\pm$0.24 & \textbf{1.97$\pm$0.27}\\
ISRUC (pat) & \textbf{4.44$\pm$0.24} & \textbf{4.63$\pm$0.77} & \textbf{4.15$\pm$0.90} & 8.65$\pm$0.99 & 9.22$\pm$0.20 & \textbf{4.64$\pm$0.47} & \textbf{4.55$\pm$0.57} & \textbf{4.61$\pm$0.41} & \textbf{3.77$\pm$1.25}\\
ISRUC & 4.45$\pm$0.02 & 5.13$\pm$0.79 & 2.65$\pm$0.40 & 9.29$\pm$0.86 & 9.08$\pm$0.03 & 4.68$\pm$0.16 & 4.60$\pm$0.38 & 4.66$\pm$0.25 & \textbf{1.84$\pm$0.31}\\
PN2017 & 12.20$\pm$0.07 & 12.31$\pm$0.21 & \textbf{3.97$\pm$0.49} & 9.70$\pm$1.19 & 16.69$\pm$0.12 & 8.40$\pm$0.84 & 12.10$\pm$0.35 & 12.19$\pm$0.07 & \textbf{3.66$\pm$1.27}\\
\midrule
C10 (ViT) & 3.40$\pm$0.01 & 1.33$\pm$0.08 & 1.17$\pm$0.08 & 1.15$\pm$0.06 & 5.17$\pm$0.03 & 1.27$\pm$0.06 & 1.14$\pm$0.06 & \textbf{1.09$\pm$0.06} & \textbf{1.06$\pm$0.04}\\
C10 (Mixer) & 3.33$\pm$0.02 & 2.01$\pm$0.13 & 1.49$\pm$0.12 & 1.76$\pm$0.24 & 7.00$\pm$0.03 & 1.58$\pm$0.06 & 1.60$\pm$0.10 & 1.54$\pm$0.08 & \textbf{1.42$\pm$0.08}\\
C100 (ViT) & 5.97$\pm$0.04 & 5.91$\pm$0.31 & 4.91$\pm$0.16 & 5.96$\pm$0.21 & 5.75$\pm$0.06 & \textbf{4.66$\pm$0.14} & 5.09$\pm$0.16 & 4.98$\pm$0.10 & \textbf{4.64$\pm$0.11}\\
C100 (Mixer) & 5.36$\pm$0.06 & 6.08$\pm$0.21 & 5.28$\pm$0.16 & 6.64$\pm$0.29 & 5.83$\pm$0.05 & 4.93$\pm$0.10 & 5.44$\pm$0.29 & 5.19$\pm$0.15 & \textbf{4.67$\pm$0.11}\\
SVHN (ViT) & 3.50$\pm$0.02 & 2.48$\pm$0.60 & \textbf{1.25$\pm$0.07} & 2.98$\pm$0.22 & 6.11$\pm$0.02 & 1.45$\pm$0.06 & 1.34$\pm$0.06 & 1.36$\pm$0.06 & \textbf{1.32$\pm$0.10}\\
SVHN (Mixer) & 3.34$\pm$0.02 & 3.31$\pm$0.69 & \textbf{1.22$\pm$0.10} & 3.00$\pm$0.16 & 5.79$\pm$0.02 & 1.56$\pm$0.05 & 1.45$\pm$0.04 & 1.40$\pm$0.05 & 1.42$\pm$0.10\\
\bottomrule
\end{tabular}
}
\end{small}
\end{table}
}

\def \TableExpMainECE{
\begin{table}[ht]
\caption{ECE in $10^{-2}$ ($\downarrow$ means lower=better).
The best accuracy-preserving method is in \textbf{bold} (p=0.01).
\methodname is usually on par or better than the best baseline.
}
\label{tab:main:ece}
\begin{center}
\begin{small}
\scalebox{0.75}{
\begin{tabular}{p{0.14\textwidth}p{0.1\textwidth}|p{0.1\textwidth}p{0.1\textwidth}p{0.1\textwidth}p{0.1\textwidth}p{0.1\textwidth}p{0.1\textwidth}p{0.1\textwidth}|p{0.1\textwidth}}
\toprule
 ECE $\downarrow$  & \baselineUnCal & \baselineTS & \baselineDirCal & \baselineIMax & \baselineFocal & \baselineSpline & \baselineIOP & \baselineGP & \methodname\\ 
\midrule
IIIC (pat) & 9.13$\pm$1.07 & \textbf{4.94$\pm$2.77} & \textbf{2.79$\pm$1.66} & 10.56$\pm$4.05 & 7.41$\pm$0.61 & \textbf{4.48$\pm$2.08} & \textbf{4.50$\pm$2.17} & \textbf{3.77$\pm$1.69} & \textbf{4.19$\pm$1.46}\\
IIIC & 9.10$\pm$0.05 & 4.44$\pm$1.53 & \textbf{1.15$\pm$0.17} & 10.17$\pm$0.81 & 7.22$\pm$0.04 & 3.05$\pm$0.70 & 3.44$\pm$0.38 & 1.66$\pm$0.56 & 2.55$\pm$0.61\\
ISRUC (pat) & 3.59$\pm$0.32 & \textbf{2.66$\pm$1.59} & 2.92$\pm$1.02 & 8.82$\pm$1.41 & 14.98$\pm$0.39 & \textbf{1.91$\pm$0.39} & \textbf{2.42$\pm$1.41} & \textbf{1.94$\pm$0.55} & \textbf{2.70$\pm$1.31}\\
ISRUC & 3.46$\pm$0.06 & 3.81$\pm$1.70 & 2.18$\pm$0.66 & 9.58$\pm$1.26 & 14.82$\pm$0.06 & \textbf{1.37$\pm$0.55} & 2.66$\pm$0.97 & 2.02$\pm$0.79 & \textbf{1.25$\pm$0.44}\\
PN2017 & 16.99$\pm$0.11 & 16.94$\pm$0.58 & 5.42$\pm$0.76 & 8.97$\pm$1.85 & 24.64$\pm$0.13 & 5.71$\pm$1.91 & 15.96$\pm$2.39 & 17.00$\pm$0.11 & \textbf{3.94$\pm$0.86}\\
\midrule
C10 (ViT) & 9.16$\pm$0.05 & 0.76$\pm$0.11 & 0.44$\pm$0.08 & 0.61$\pm$0.06 & 7.18$\pm$0.07 & 0.47$\pm$0.09 & 0.34$\pm$0.06 & \textbf{0.25$\pm$0.08} & 0.39$\pm$0.10\\
C10 (Mixer) & 9.05$\pm$0.07 & 1.07$\pm$0.13 & 0.50$\pm$0.06 & 1.04$\pm$0.17 & 12.54$\pm$0.06 & 0.47$\pm$0.11 & 0.55$\pm$0.11 & \textbf{0.31$\pm$0.10} & 0.60$\pm$0.09\\
C100 (ViT) & 11.65$\pm$0.14 & 2.78$\pm$0.46 & \textbf{0.71$\pm$0.11} & 3.39$\pm$0.23 & 9.97$\pm$0.09 & 0.99$\pm$0.29 & 1.11$\pm$0.29 & \textbf{0.82$\pm$0.16} & 1.51$\pm$0.34\\
C100 (Mixer) & 13.71$\pm$0.15 & 3.11$\pm$0.36 & 1.08$\pm$0.28 & 4.82$\pm$0.25 & 14.35$\pm$0.21 & \textbf{1.22$\pm$0.38} & 1.78$\pm$0.71 & \textbf{0.95$\pm$0.19} & 3.09$\pm$0.50\\
SVHN (ViT) & 10.11$\pm$0.05 & \textbf{2.43$\pm$2.72} & \textbf{0.55$\pm$0.08} & 2.08$\pm$0.18 & 12.17$\pm$0.08 & \textbf{0.64$\pm$0.13} & \textbf{0.50$\pm$0.13} & \textbf{0.53$\pm$0.10} & 0.66$\pm$0.12\\
SVHN (Mixer) & 10.30$\pm$0.04 & 3.19$\pm$2.56 & \textbf{0.49$\pm$0.08} & 2.21$\pm$0.10 & 11.09$\pm$0.06 & 0.65$\pm$0.13 & \textbf{0.45$\pm$0.12} & \textbf{0.51$\pm$0.08} & 0.72$\pm$0.11\\
\bottomrule
\end{tabular}
}
\end{small}
\end{center}
\end{table}
}

\def \TableExpMainBrier{
\begin{table}[ht]
\caption{
Brier Score in $10^{-2}$ ($\downarrow$ means lower=better).
The best accuracy-preserving methods are in \textbf{bold} (p=0.01).
}
\label{tab:main:brier}
\centering
\begin{small}
\scalebox{0.75}{
\begin{tabular}{p{0.14\textwidth}p{0.1\textwidth}|p{0.1\textwidth}p{0.1\textwidth}p{0.1\textwidth}p{0.1\textwidth}p{0.1\textwidth}p{0.1\textwidth}p{0.1\textwidth}|p{0.1\textwidth}}
\toprule
 Brier $\downarrow$ & \baselineUnCal & \baselineTS & \baselineDirCal & \baselineIMax & \baselineFocal & \baselineSpline & \baselineIOP & \baselineGP & \methodname\\ 
\midrule
IIIC (pat) & 21.30$\pm$0.25 & 20.70$\pm$0.69 & \textbf{18.94$\pm$0.55} & 21.09$\pm$1.29 & 21.48$\pm$0.19 & 20.43$\pm$0.50 & 20.52$\pm$0.58 & 20.33$\pm$0.42 & \textbf{19.33$\pm$0.78}\\
IIIC & 21.35$\pm$0.01 & 20.62$\pm$0.27 & 18.33$\pm$0.04 & 20.83$\pm$0.19 & 21.46$\pm$0.01 & 20.21$\pm$0.09 & 20.39$\pm$0.09 & 20.05$\pm$0.08 & \textbf{17.54$\pm$0.10}\\
ISRUC (pat) & \textbf{15.26$\pm$0.25} & \textbf{15.20$\pm$0.31} & 15.37$\pm$0.38 & 16.25$\pm$0.49 & 18.55$\pm$0.18 & \textbf{15.11$\pm$0.26} & \textbf{15.16$\pm$0.31} & \textbf{15.16$\pm$0.29} & \textbf{14.97$\pm$0.29}\\
ISRUC & 15.46$\pm$0.03 & 15.50$\pm$0.19 & 15.07$\pm$0.09 & 16.62$\pm$0.33 & 18.77$\pm$0.01 & 15.31$\pm$0.05 & 15.39$\pm$0.10 & 15.35$\pm$0.06 & \textbf{14.28$\pm$0.08}\\
PN2017 & 26.61$\pm$0.05 & 26.74$\pm$0.27 & \textbf{22.44$\pm$0.15} & 24.58$\pm$0.59 & 17.79$\pm$0.03 & 23.28$\pm$0.37 & 26.39$\pm$0.69 & 26.61$\pm$0.05 & \textbf{22.56$\pm$0.28}\\
\midrule
C10 (ViT) & 1.76$\pm$0.03 & 0.89$\pm$0.06 & \textbf{0.78$\pm$0.04} & 0.84$\pm$0.04 & 1.75$\pm$0.03 & \textbf{0.79$\pm$0.04} & \textbf{0.79$\pm$0.04} & \textbf{0.78$\pm$0.04} & \textbf{0.75$\pm$0.05}\\
C10 (Mixer) & 2.29$\pm$0.03 & 1.48$\pm$0.07 & 1.42$\pm$0.05 & 1.46$\pm$0.08 & 4.16$\pm$0.04 & 1.39$\pm$0.04 & 1.40$\pm$0.05 & \textbf{1.37$\pm$0.04} & \textbf{1.34$\pm$0.04}\\
C100 (ViT) & 6.94$\pm$0.08 & 5.35$\pm$0.15 & 5.17$\pm$0.10 & 5.48$\pm$0.14 & 6.93$\pm$0.07 & 5.19$\pm$0.09 & 5.18$\pm$0.10 & 5.14$\pm$0.09 & \textbf{5.01$\pm$0.08}\\
C100 (Mixer) & 10.15$\pm$0.11 & 7.94$\pm$0.17 & 7.82$\pm$0.12 & 8.23$\pm$0.17 & 10.91$\pm$0.08 & 7.76$\pm$0.12 & 7.82$\pm$0.15 & \textbf{7.72$\pm$0.13} & \textbf{7.61$\pm$0.09}\\
SVHN (ViT) & 3.99$\pm$0.03 & 3.03$\pm$0.34 & 2.78$\pm$0.04 & 2.99$\pm$0.07 & 5.03$\pm$0.03 & 2.80$\pm$0.03 & 2.79$\pm$0.04 & 2.79$\pm$0.04 & \textbf{2.49$\pm$0.03}\\
SVHN (Mixer) & 4.03$\pm$0.03 & 3.21$\pm$0.36 & 2.77$\pm$0.03 & 3.04$\pm$0.04 & 5.06$\pm$0.04 & 2.84$\pm$0.03 & 2.81$\pm$0.04 & 2.81$\pm$0.04 & \textbf{2.68$\pm$0.03}\\
\bottomrule
\end{tabular}
}
\end{small}
\end{table}
}

\def \AppendixFigIIICReliability{
\begin{figure}[H]
\begin{center}
\centerline{\includegraphics[width=1\columnwidth]{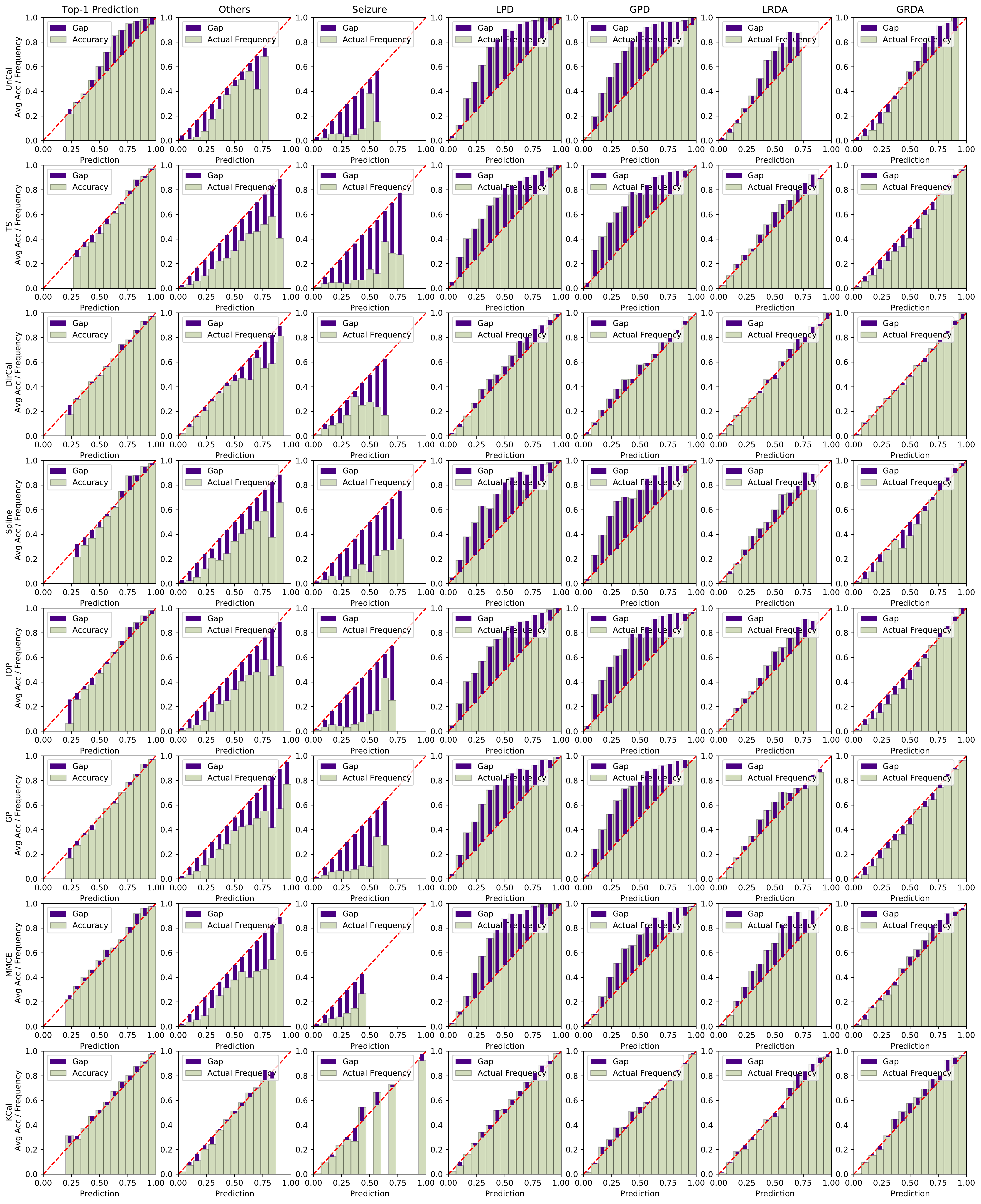}}
\caption{
Reliability diagrams for the IIIC dataset.
}
\label{fig:appendix:reliability:iiic}
\end{center}
\end{figure}
}
\def \AppendixFigISRUCReliability{
\begin{figure}[H]
\begin{center}
\centerline{\includegraphics[width=1\columnwidth]{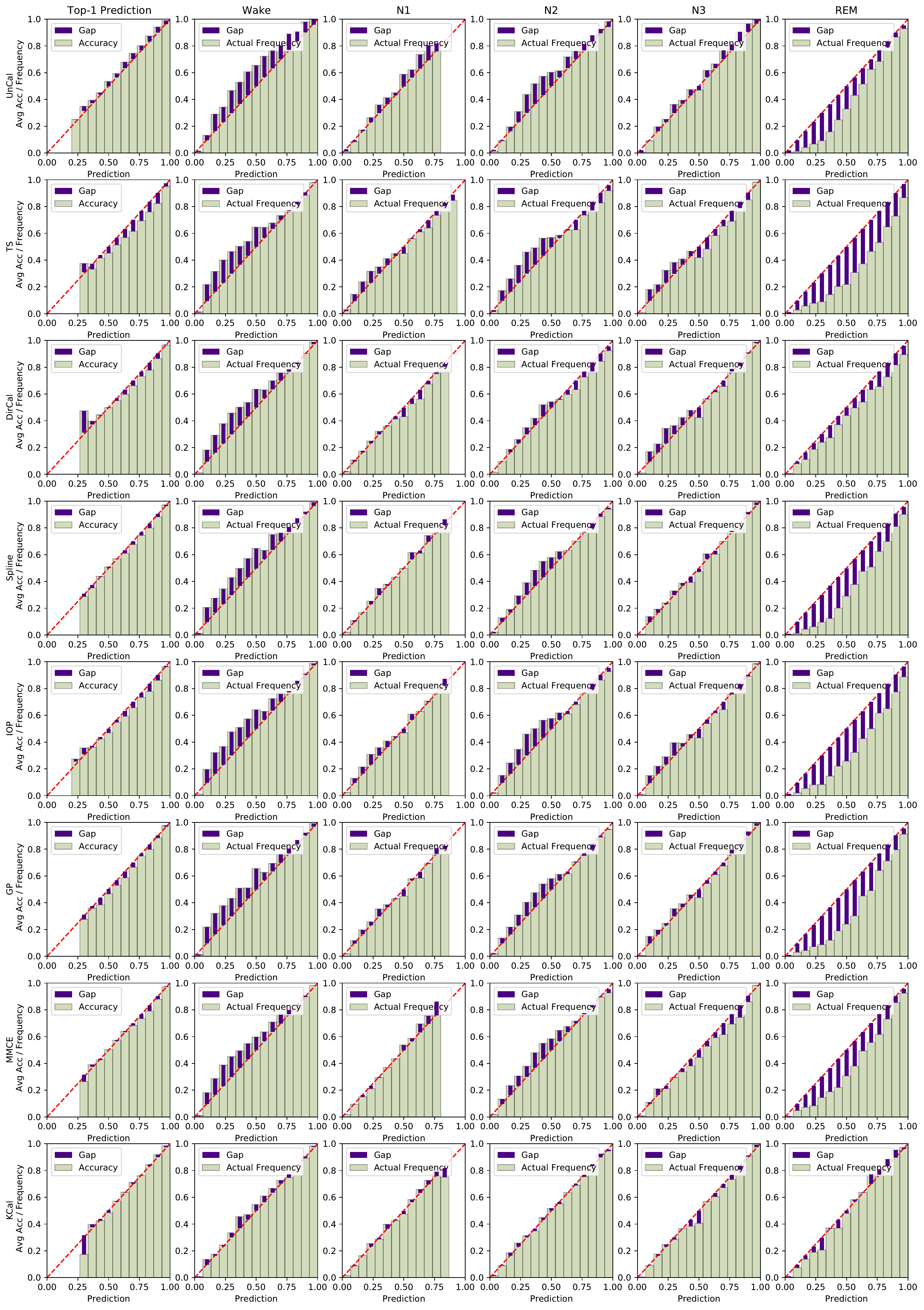}}
\caption{
Reliability diagrams for the ISRUC dataset.
}
\label{fig:appendix:reliability:isruc}
\end{center}
\end{figure}
}

\def \AppendixFigECGReliability{
\begin{figure}[H]
\begin{center}
\centerline{\includegraphics[width=1\columnwidth]{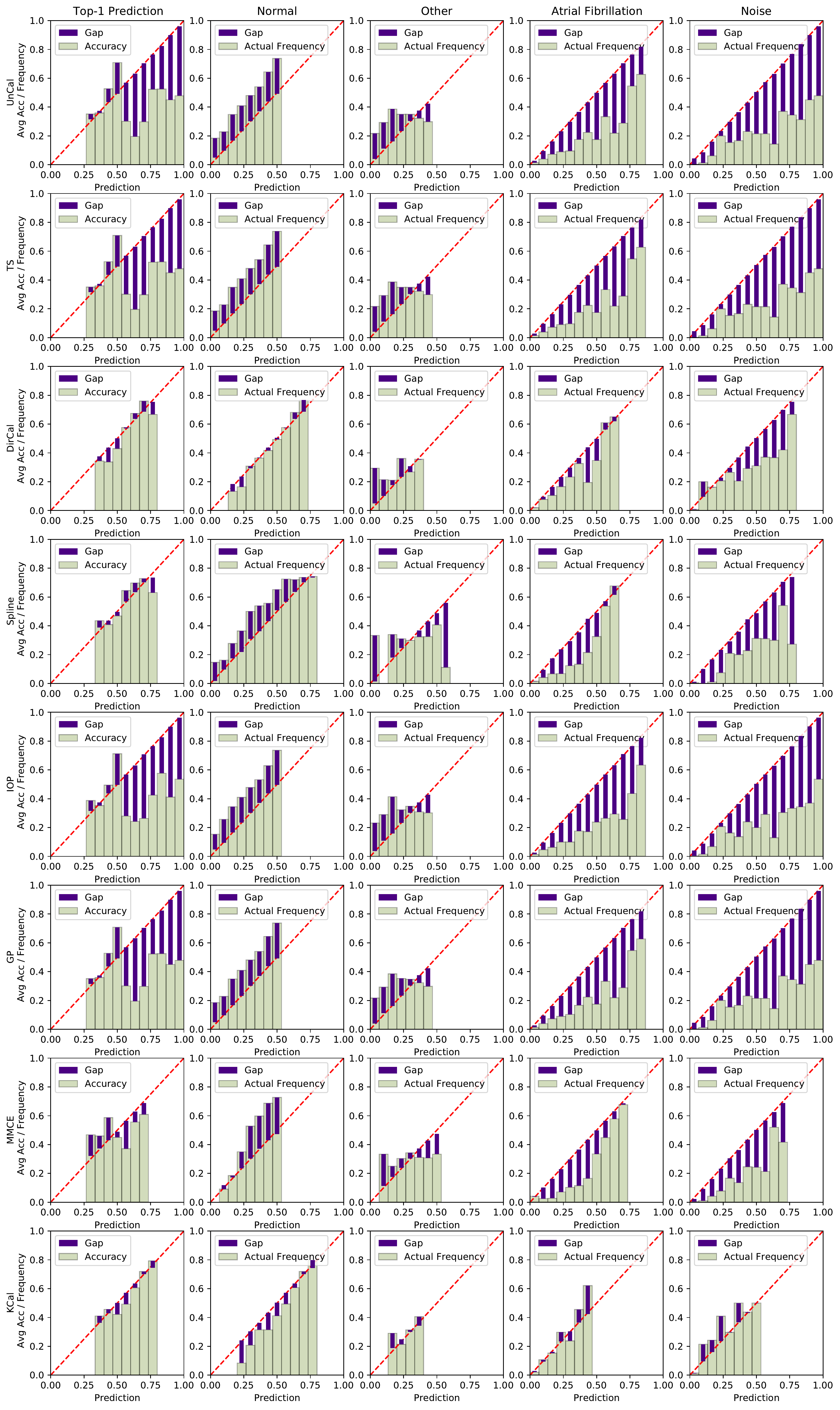}}
\caption{
Reliability diagrams for the PN2017 dataset.
}
\label{fig:appendix:reliability:ecg}
\end{center}
\end{figure}
}

\def \KernelArchitecture{
\begin{figure}
    \centering
    \begin{tikzpicture}[node distance=0.2cm,
    simple/.style={draw},
    emptyAnchor/.style={circle, fill=none}, %remove draw
    X/.style={draw, fill=none},
    Linear/.style={draw, fill=yellow!50},
    DNN/.style={outer sep=1pt,text width=2cm,minimum height=2cm,minimum width=2cm, fill=gray!60, fill opacity=0.3},
    OUT/.style={draw, fill=none}
    ]
    \footnotesize
    \node[X] (X) at (-4,0) {$\embedder(X)$};
    \node[Linear] (linear1) [right=0.5 cm of X] {Linear};
    \node[emptyAnchor] (elu) [above right=0.0 cm and -0.5cm of linear1] {ELU};
    \node[Linear] (linear2) [above=0.8 cm of linear1] {Linear};
    \node[emptyAnchor, draw] (add) [right=0.4cm of linear2] {   $+$};
    \node[DNN, fit=(elu) (linear1) (linear2) (add)] (proj) {};
    \node[OUT] (out) [right=0.5cm of add] {$\proj(\embedder(X))$};
    \draw [->] (X) -- (linear1);
    \draw [->] (linear1.north) -- (linear2);
    \draw [->] (linear2.east) -- (add.west);
    \draw [->] (add.east) -- (out.west);
    
    \draw[->] plot [smooth] coordinates { (linear1.east)  (add.south) };
    
    \end{tikzpicture}
    \caption{Structure of the learnable projection $\proj$ (in gray).}
    \label{fig:mlp2}
    \vskip -0.2in
\end{figure}
}

\def \FigSeizureREMReliability{

\begin{figure}[H]
\centering
\includegraphics[width=1\columnwidth]{Figures/Appendix/IIIC_Seizure_illustrate.pdf}
\includegraphics[width=1\columnwidth]{Figures/Appendix/ISRUC_REM_illustrate.pdf}
\caption{
Reliability diagrams for Seizure in IIIC and REM in ISRUC.
Only \methodname achieves reasonable calibration quality consistently.
}
\label{fig:reliability:seizure_rem}
\end{figure}
}

\def \AppendixTabLinear{

\begin{table*}[ht]
\caption{
Comparison between the architecture described in Figure~\ref{fig:mlp2} (\methodname) and a simple linear projection with the same input and output dimensions (\methodname-Linear).
On average, \methodname adapts to different datasets and architectures better than (\methodname-Linear), although the performance is generally similar. 
}
\label{tab:appendix:linear}
\begin{center}
\begin{small}
\scalebox{0.8}{
\begin{tabular}{l|cc|cc|cc|cc}
\toprule
   & \multicolumn{2}{c}{Accuracy$\uparrow$}& \multicolumn{2}{c}{CECE$\downarrow$}& \multicolumn{2}{c}{ECE$\downarrow$}&\multicolumn{2}{c}{Brier $\downarrow$}\\
   & \methodname & \methodname-Lienar & \methodname & \methodname-Lienar& \methodname & \methodname-Lienar & \methodname & \methodname-Lienar\\
\midrule
IIIC (pat) & \textbf{61.67$\pm$2.22} & 61.51$\pm$2.46 & \textbf{4.60$\pm$1.28} & 4.64$\pm$1.42 & \textbf{4.19$\pm$1.46} & 4.38$\pm$2.11 & 19.33$\pm$0.78 & \textbf{19.28$\pm$0.82} \\
IIIC   & \textbf{66.32$\pm$0.21} & 65.59$\pm$0.20 & \textbf{1.97$\pm$0.27} & 2.03$\pm$0.26 & \textbf{2.55$\pm$0.61} & 3.06$\pm$0.69 & \textbf{17.54$\pm$0.10} & 17.88$\pm$0.09 \\
ISRUC (pat) & \textbf{76.13$\pm$0.89} & 76.02$\pm$1.08 & \textbf{3.77$\pm$1.25} & 3.90$\pm$1.33 & \textbf{2.70$\pm$1.31} & 2.84$\pm$1.58 & \textbf{14.97$\pm$0.29} & 15.04$\pm$0.30\\
ISRUC   & \textbf{77.45$\pm$0.16} & 77.19$\pm$0.19 & \textbf{1.84$\pm$0.31} & 2.00$\pm$0.32 & \textbf{1.25$\pm$0.44} & 1.61$\pm$0.51 & \textbf{14.28$\pm$0.08} & 14.37$\pm$0.08\\
PN2017 & \textbf{60.36$\pm$0.61} & 60.15$\pm$0.56 & 3.66$\pm$1.27 & \textbf{3.53$\pm$1.29} & \textbf{3.94$\pm$0.86} & 4.55$\pm$0.85 & \textbf{22.56$\pm$0.28} & 22.69$\pm$0.32\\
\midrule
C10 (ViT) & \textbf{98.98$\pm$0.09} & 98.96$\pm$0.07 & 1.06$\pm$0.04 & \textbf{1.04$\pm$0.09} & 0.39$\pm$0.10 & \textbf{0.36$\pm$0.11} & \textbf{0.75$\pm$0.05} & \textbf{0.75$\pm$0.05}\\
C10 (Mixer) & \textbf{98.14$\pm$0.06} & 98.12$\pm$0.09 & \textbf{1.42$\pm$0.08} & 1.47$\pm$0.06 & \textbf{0.60$\pm$0.09} & \textbf{0.60$\pm$0.12} & \textbf{1.34$\pm$0.04} & \textbf{1.34$\pm$0.04}\\
C100 (ViT) & 92.37$\pm$0.15 & \textbf{92.47$\pm$0.14} & \textbf{4.64$\pm$0.11} & 4.68$\pm$0.06 & 1.51$\pm$0.34 & \textbf{1.49$\pm$0.32} & 5.01$\pm$0.08 & \textbf{4.93$\pm$0.08}\\
C100 (Mixer) & 87.55$\pm$0.16 & \textbf{88.00$\pm$0.24} & \textbf{4.67$\pm$0.11} & 4.75$\pm$0.13 & 3.09$\pm$0.50 & \textbf{2.83$\pm$0.47} & 7.61$\pm$0.09 & \textbf{7.39$\pm$0.07} \\
SVHN (ViT) & \textbf{96.42$\pm$0.05} & 96.36$\pm$0.06 & \textbf{1.32$\pm$0.10} & 1.40$\pm$0.08 & \textbf{0.66$\pm$0.12} & 0.67$\pm$0.12 & \textbf{2.49$\pm$0.03} & \textbf{2.49$\pm$0.03}\\
SVHN (Mixer) & 96.10$\pm$0.04 & \textbf{96.13$\pm$0.04} & \textbf{1.42$\pm$0.10} & 1.52$\pm$0.09 & 0.72$\pm$0.11 & \textbf{0.65$\pm$0.11} & \textbf{2.68$\pm$0.03} & 2.69$\pm$0.03\\
\bottomrule
\end{tabular}
}
\end{small}
\end{center}
\end{table*}
}

\def \AppendixFigBW{
\begin{figure}[ht]
\vskip 0.2in
\begin{center}
\includegraphics[width=\columnwidth]{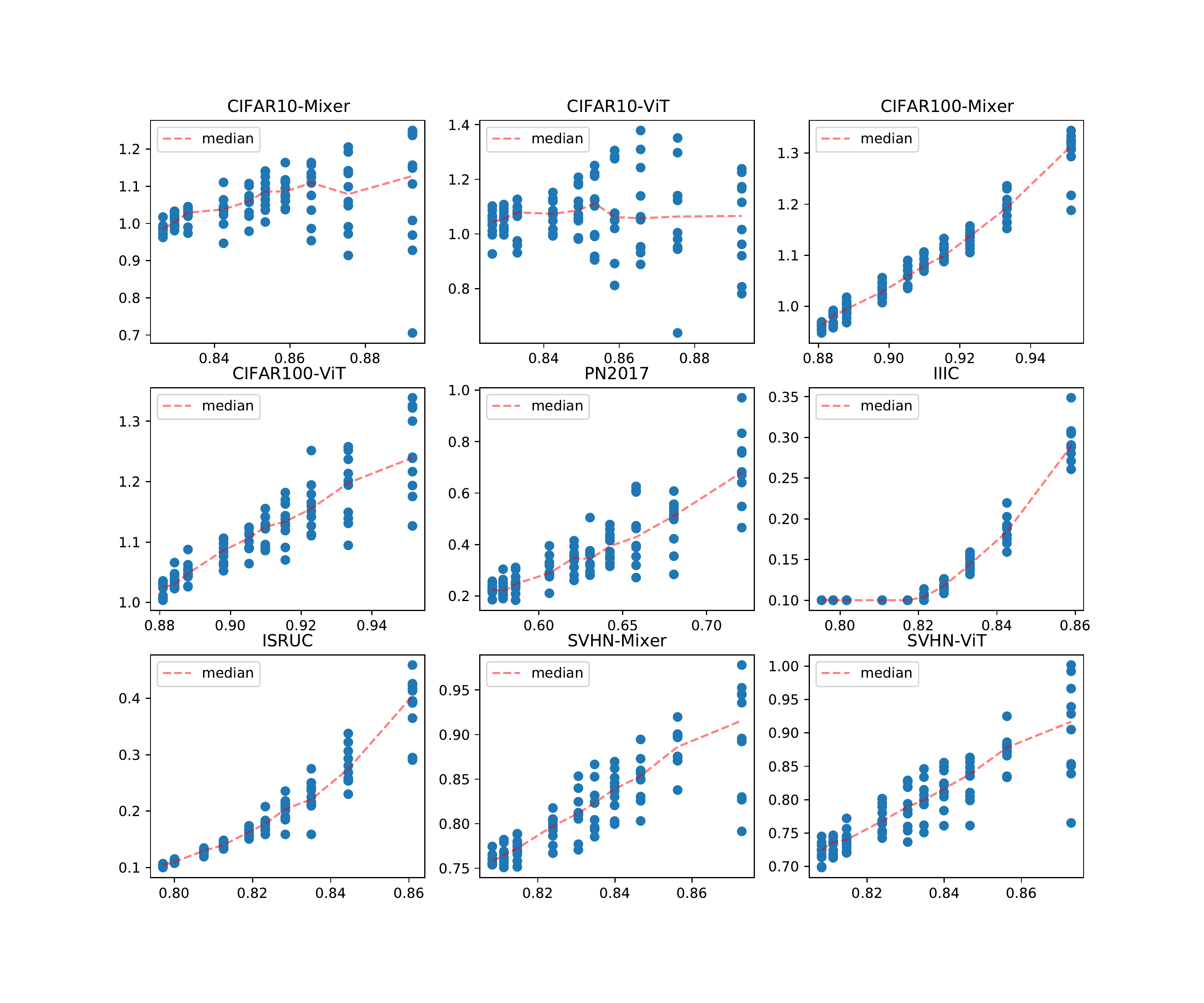}
\caption{
Empirically-selected bandwidth ($b^*$) on the y-axis, and predicted bandwidth ($\Theta(m^{-\frac{1}{d+4}})$) on the x-axis. 
For each calibration set size we have 10 experiments like in the main text, and we plot the scatter plot and median of the experiments.
As expected, we see a nearly linear relationship for most data, except for IIIC, which exhibits a  piece-wise linear pattern.
This suggests that in practice, as new samples are added into the calibration set in an online manner, we could \textit{compute} the bandwidth $b$ and only re-do the cross validation sparingly. 
}
\label{fig:bandwidth}
\end{center}
\vskip -0.2in
\end{figure}
}

\def \AppendixFigDim{
\begin{figure}[ht]
\vskip 0.2in
\begin{center}
\includegraphics[width=0.48\columnwidth]{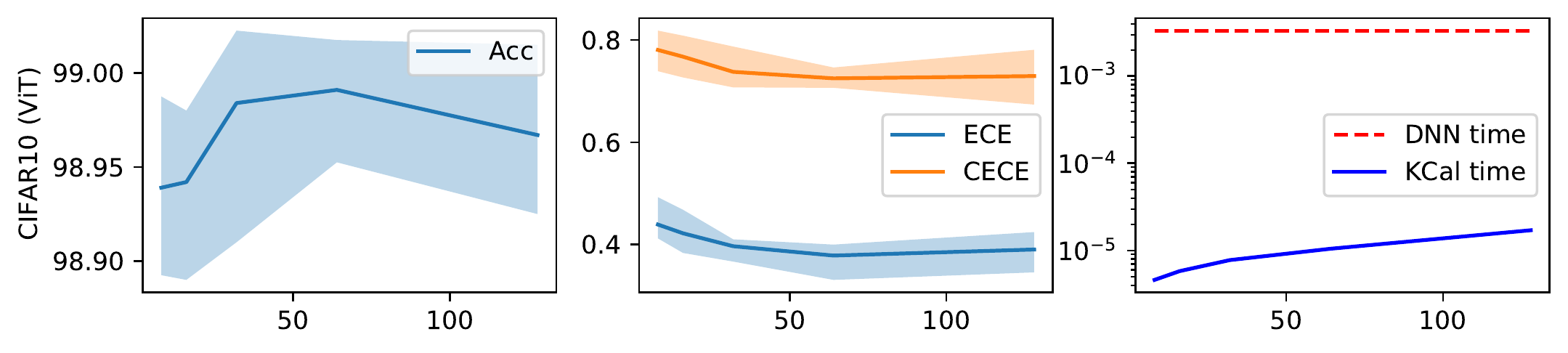}\hspace{5pt}
\includegraphics[width=0.48\columnwidth]{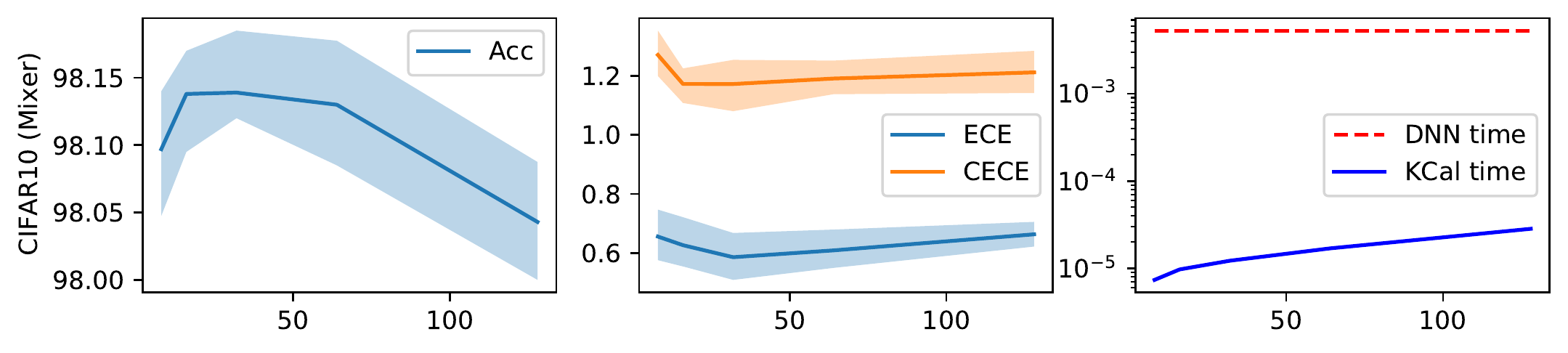}
\includegraphics[width=0.48\columnwidth]{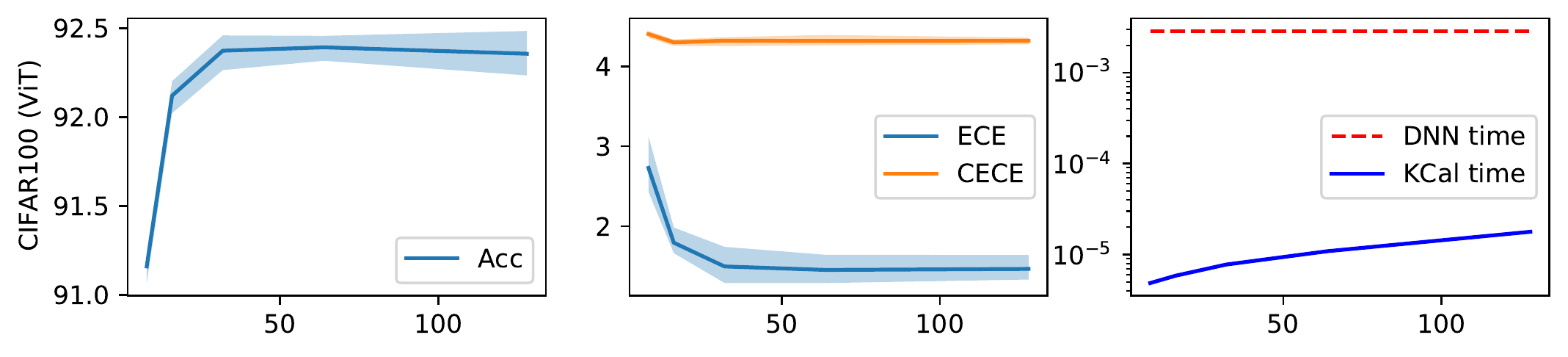}\hspace{5pt}
\includegraphics[width=0.48\columnwidth]{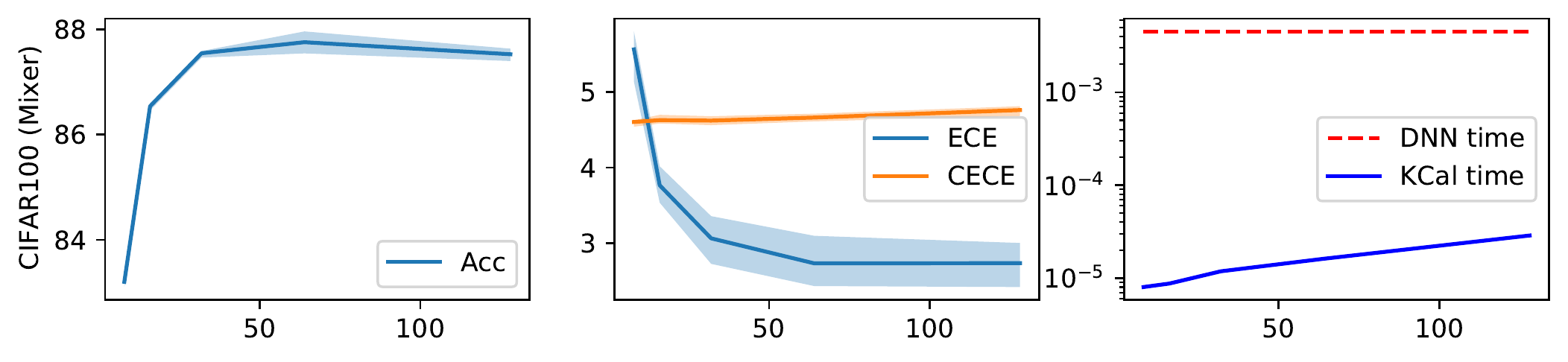}
\includegraphics[width=0.48\columnwidth]{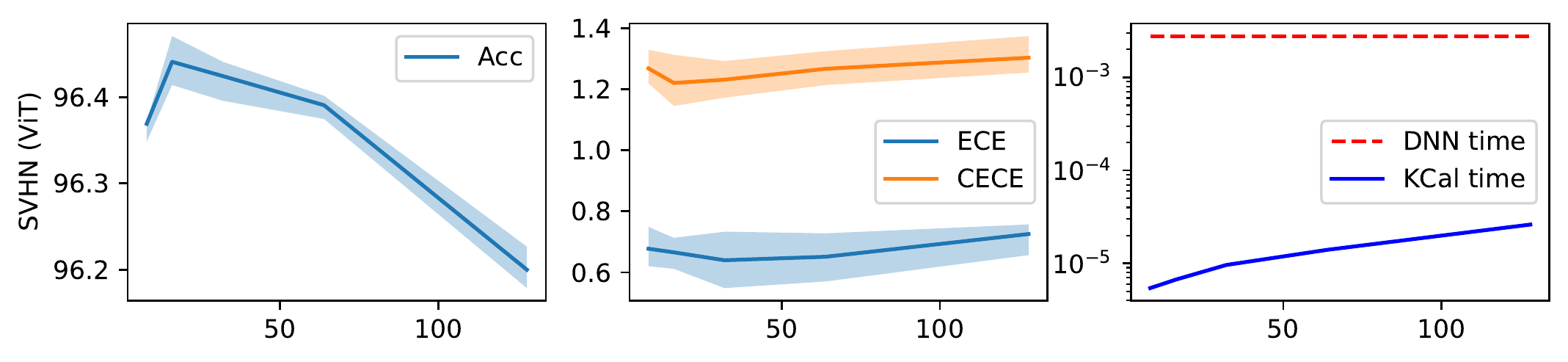}\hspace{5pt}
\includegraphics[width=0.48\columnwidth]{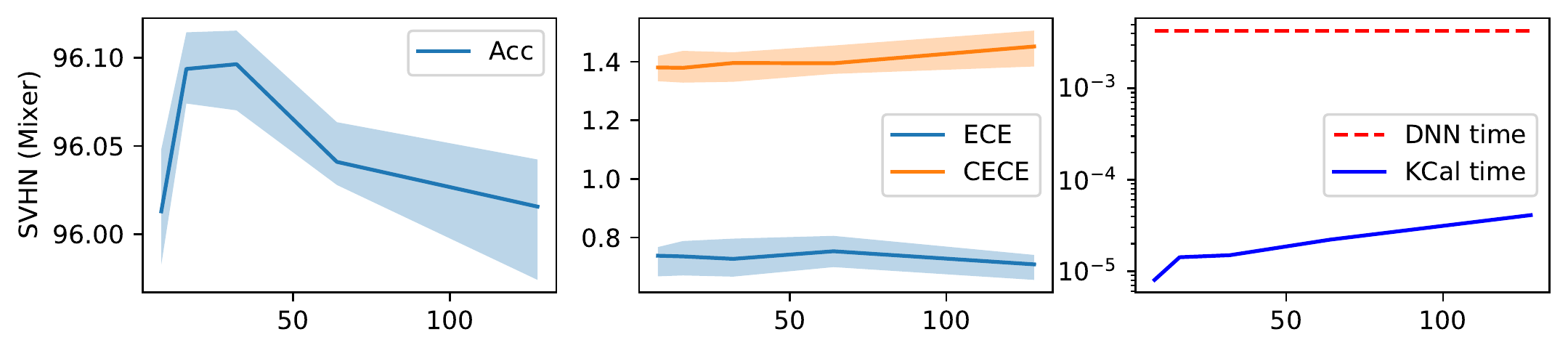}
\includegraphics[width=0.48\columnwidth]{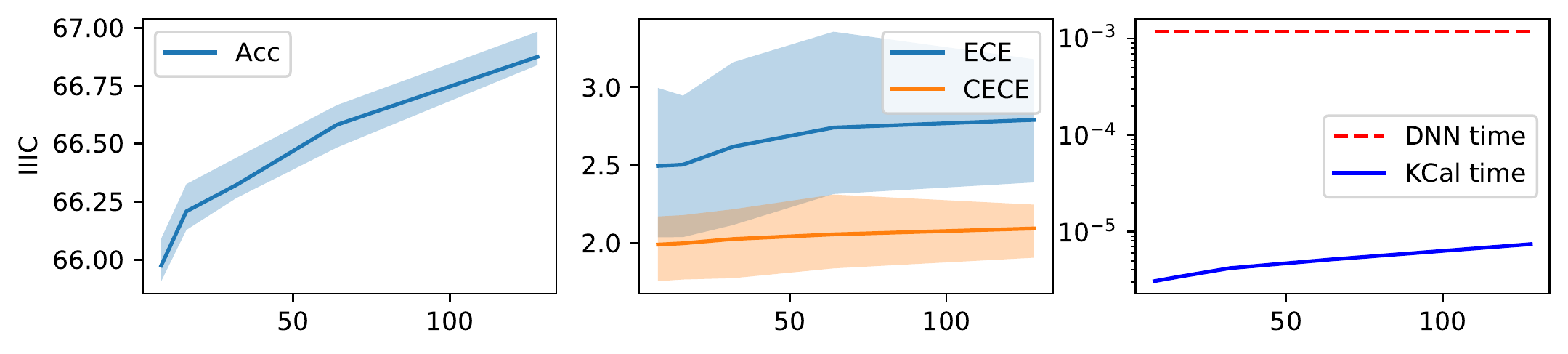}\hspace{5pt}
\includegraphics[width=0.48\columnwidth]{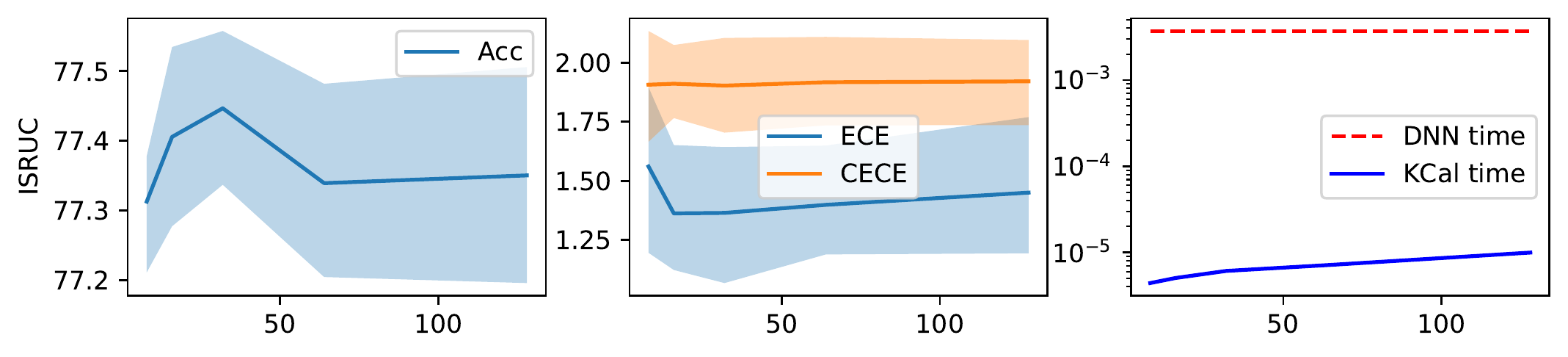}
\includegraphics[width=0.5\columnwidth]{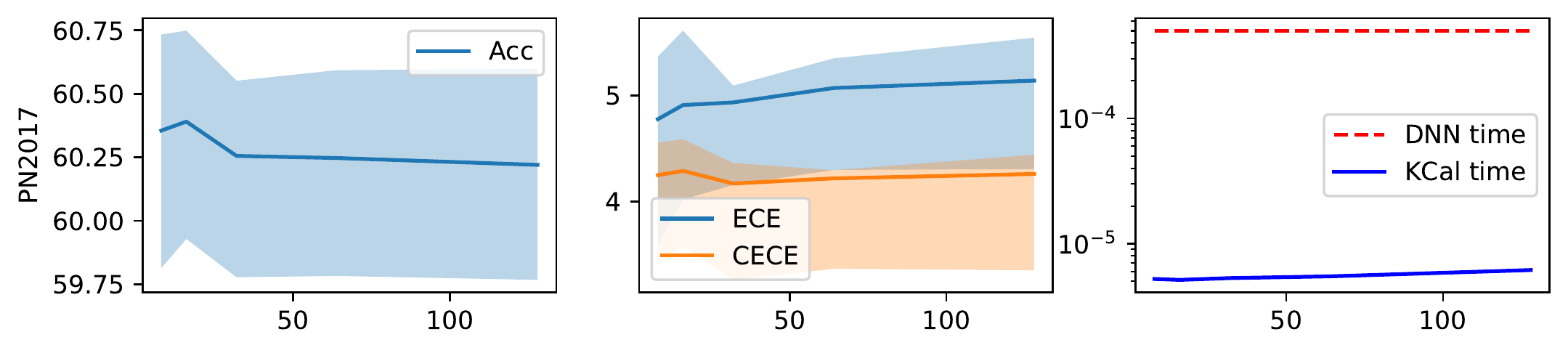}
\caption{
Change in performance and inference time if we we change $d$ (the output embedding size of $\proj$. 
``DNN time'' refers to the average time running $\embedder$ for one input $x$, and ``\methodname time'' refers to the average time transforming $\embedder(x)$ to $\phat(x)$ using \methodname. 
For Accuracy, ECE and CECE, the unit is percentage. 
The band represents the median 50\% among 10 experiments.
For time, the unit is second.
Performance is not always improving as $d$ increases, but a larger $d$ naturally leads to larger overhead. 
It is however worth noting that in all experiment, the overhead (``\methodname time'') is negligible compared with the `DNN time''. 
}
\label{appendix:fig:dim}
\end{center}
\vskip -0.2in
\end{figure}
}

\def \AppendixFigBWConvexity{
\begin{figure}[ht]
\vskip 0.2in
\begin{center}
\includegraphics[width=0.48\columnwidth]{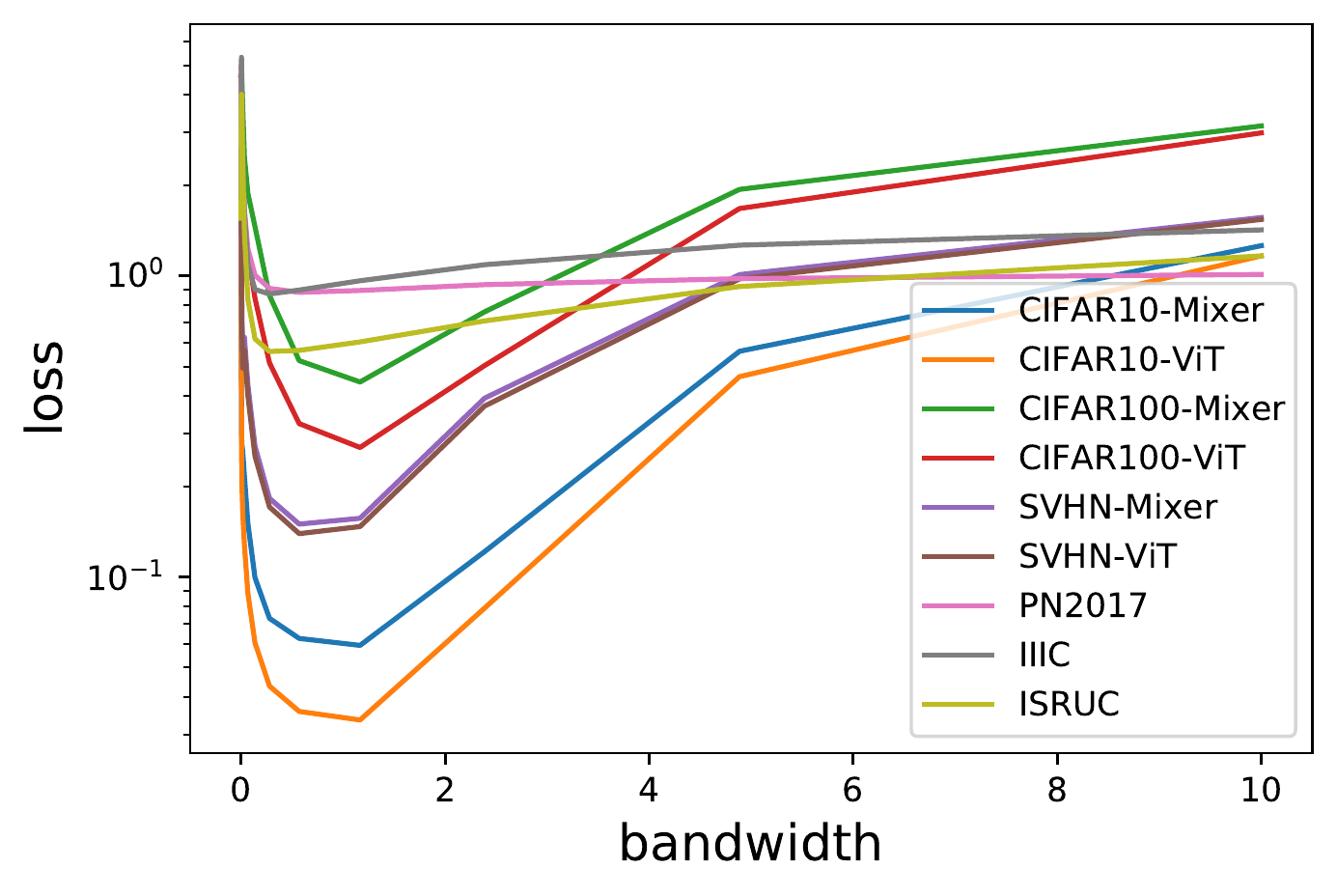}
\caption{
Change in cross validation loss used for the Golden-Section Search mentioned in Section~\ref{sec:method:impl} as a function of bandwidth.
As we can see, the loss is indeed roughly convex in the bandwidth for all datasets.
}
\label{appendix:fig:bw_convexity}
\end{center}
\vskip -0.2in
\end{figure}
}

\def \AppendixFigSizeSCal{
\begin{figure}[ht]
\vskip 0.2in
\begin{center}
\includegraphics[width=0.98\columnwidth]{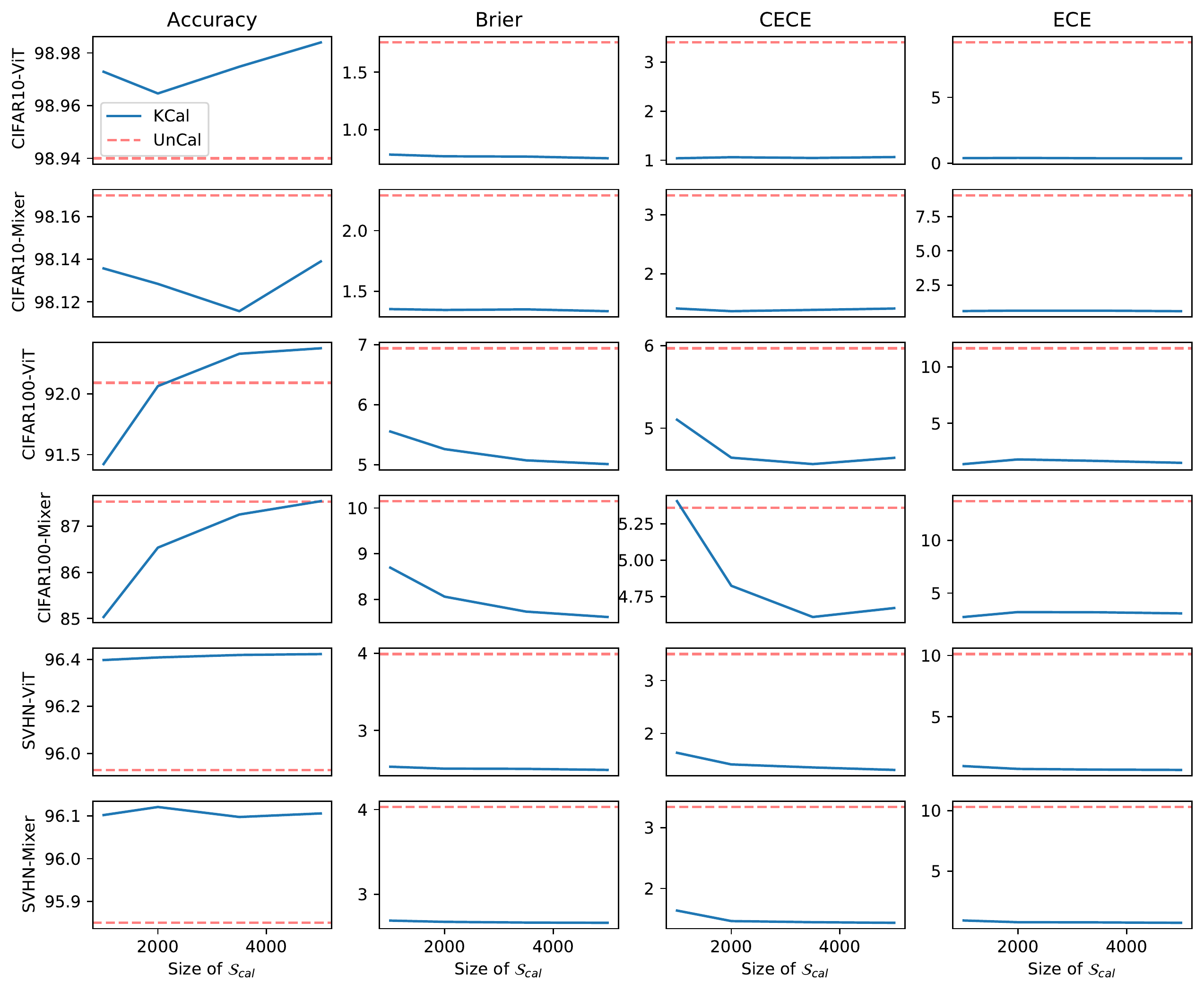}%\hspace{5pt}

\includegraphics[width=0.98\columnwidth]{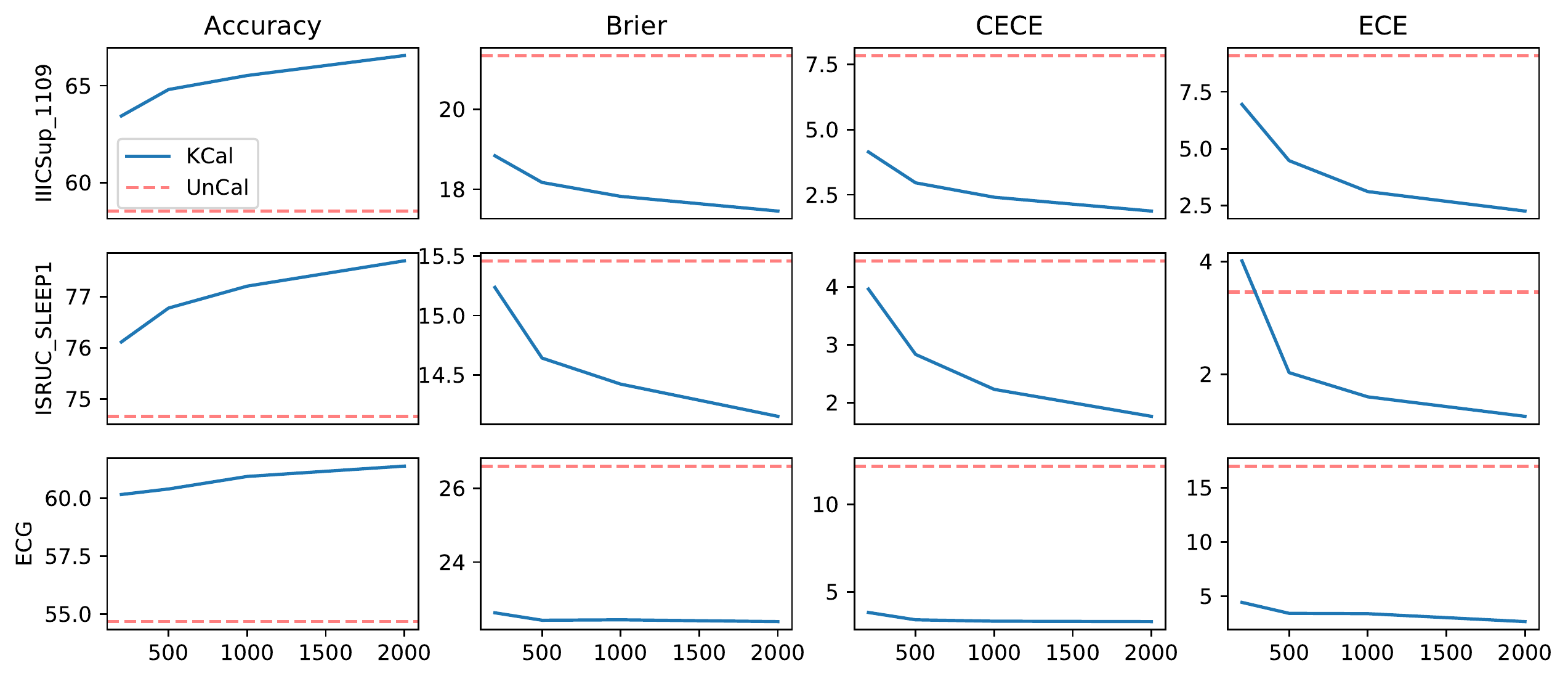}%\hspace{5pt}
\caption{
We change the size of $\calibrationset$ and repeat the experiment for \methodname. 
The red dashed line denotes \baselineUnCal.
We see that, as expected, all metrics improves as the size of the calibration set increases, and the performance is generally stable.
}
\label{appendix:fig:size_of_scal}
\end{center}
\vskip -0.2in
\end{figure}
}

\begin{document}

\title{Taking a Step Back with \methodname: Multi-Class Kernel-Based Calibration for Deep Neural Networks}

\author{%
  Zhen Lin \\%\thanks{Contact: zhenlin4@illinois.edu} \\
  University of Illinois at Urbana-Champaign\\
  Urbana, IL 61801 \\
  \texttt{zhenlin4@illinois.edu} \\
  % examples of more authors
  \AND
  Shubhendu Trivedi \\
  \texttt{shubhendu@csail.mit.edu} \\
  \AND 
  Jimeng Sun \\
  University of Illinois at Urbana-Champaign\\
  Urbana, IL 61801 \\
  \texttt{jimeng@illinois.edu} \\
}

\maketitle
\begin{abstract}
\looseness=-1 Deep neural network (DNN) classifiers are often overconfident,  producing miscalibrated class probabilities. 
In high-risk applications like healthcare, practitioners require \textit{fully calibrated} probability predictions for decision-making.
That is, conditioned on the prediction \textit{vector}, \textit{every} class' probability should be close to the predicted value. 
Most existing calibration methods either lack theoretical guarantees for producing calibrated outputs, reduce classification accuracy in the process, or only calibrate the predicted class. 
This paper proposes a new Kernel-based calibration method called \methodname.
Unlike existing calibration procedures, \methodname does not operate directly on the logits or softmax outputs of the DNN.
Instead, \methodname learns a metric space on the penultimate-layer latent embedding and generates predictions using kernel density estimates on a calibration set. 
We first analyze \methodname theoretically, showing that it enjoys a provable \textit{full} calibration guarantee.
Then, through extensive experiments across a variety of datasets, we show that \methodname consistently outperforms baselines as measured by the calibration error and by proper scoring rules like the Brier Score.
\end{abstract}

%===========================================================================================
\section{Introduction}\label{sec:intro}
%Deep learning's issue aka accuracy is not good enough
\looseness=-1 The notable successes of Deep Neural Networks (DNNs) in complex classification tasks, such as object detection~\citep{PedestrianDetection}, speech recognition~\citep{6639344}, and medical diagnosis~\citep{FLANNEL, Sleepnet}, have made them essential ingredients within various critical decision-making pipelines. 
In addition to the classification accuracy, a classifier should ideally also generate reliable uncertainty estimates represented in the predicted probability vector. 
An influential study~\citep{ICML2017_Guo} reported that modern DNNs are often overconfident or \emph{miscalibrated}, which could lead to severe consequences in high-stakes applications such as healthcare~\citep{DBLP:journals/jamia/JiangOKO12}. 

%Calibration definition
Calibration is the process of closing the gap between the prediction and the ground truth distribution given this prediction. 
For a $K$-class classification problem, with covariates $X\in \mathcal{X}$ and the label $Y\in\mathcal{Y} = [K]$, denote our classifier $\mathcal{X} \mapsto \Delta^{K-1}$ as $\phat=[\hat{p}_1,\ldots,\hat{p}_K]$, where $\Delta^{K-1}$ is the ($K-1$)-simplex. Then, 
\begin{definition}(Full Calibration~\citep{NeurIPS2019_DirCal,AISTATS2019_CalEval})\label{def:full_cal}
$\phat$ is fully-calibrated if $\forall k\in[K]$:
\vspace{-0.1in}
{\small
\begin{align}%\label{eq:perf_cal}
    \forall \mathbf{q}=[q_1,\ldots,q_K] \in \Delta^{K-1},
    \mathbb{P}\{Y=k | \phat(X)=\mathbf{q}\} = q_k.
\end{align}
}
\end{definition}
It is worth noting that Def.~(\ref{def:full_cal}) implies \textit{nothing about accuracy}. 
In fact, ignoring $X$ and simply predicting $\pi$, the class frequency vector, results in a fully calibrated but inaccurate classifier.
As a result, our goal is always to improve calibration \textit{while maintaining accuracy}. 
Another important requirement is that $\phat\in \Delta^{K-1}$.
Many binary calibration methods such as \cite{KDD_HistogramBinning, KDD_Isotonic} result in vectors that are not interpretable as probabilities, and have to be normalized.

Many existing works only consider {\it confidence calibration}~\citep{ICML2017_Guo,ICML2020_MixNMatch_KDEEval,ICML2020_NonParaCal,ICML2021_MetaCal}, a much weaker notion than that encapsulated by Def.~(\ref{def:full_cal}) and only calibrates the predicted class~\citep{NeurIPS2019_DirCal,AISTATS2019_CalEval}.
\begin{definition}(Confidence Calibration)\label{def:conf_cal}
$\phat$ is confidence-calibrated if:
\begin{align}%\label{eq:conf_cal}
    \forall q\in[0,1],\mathbb{P}\{Y=\argmax_{k}\hat{p}_k(X) | \max_k\hat{p}_k(X) = q\} = q.
\end{align}
\vspace{-4mm}
\end{definition}

\def \FigIntroReliability{
\begin{wrapfigure}[20]{L}{0.33\textwidth}
\vskip -0.3in
\centering
\includegraphics[width=0.33\columnwidth]{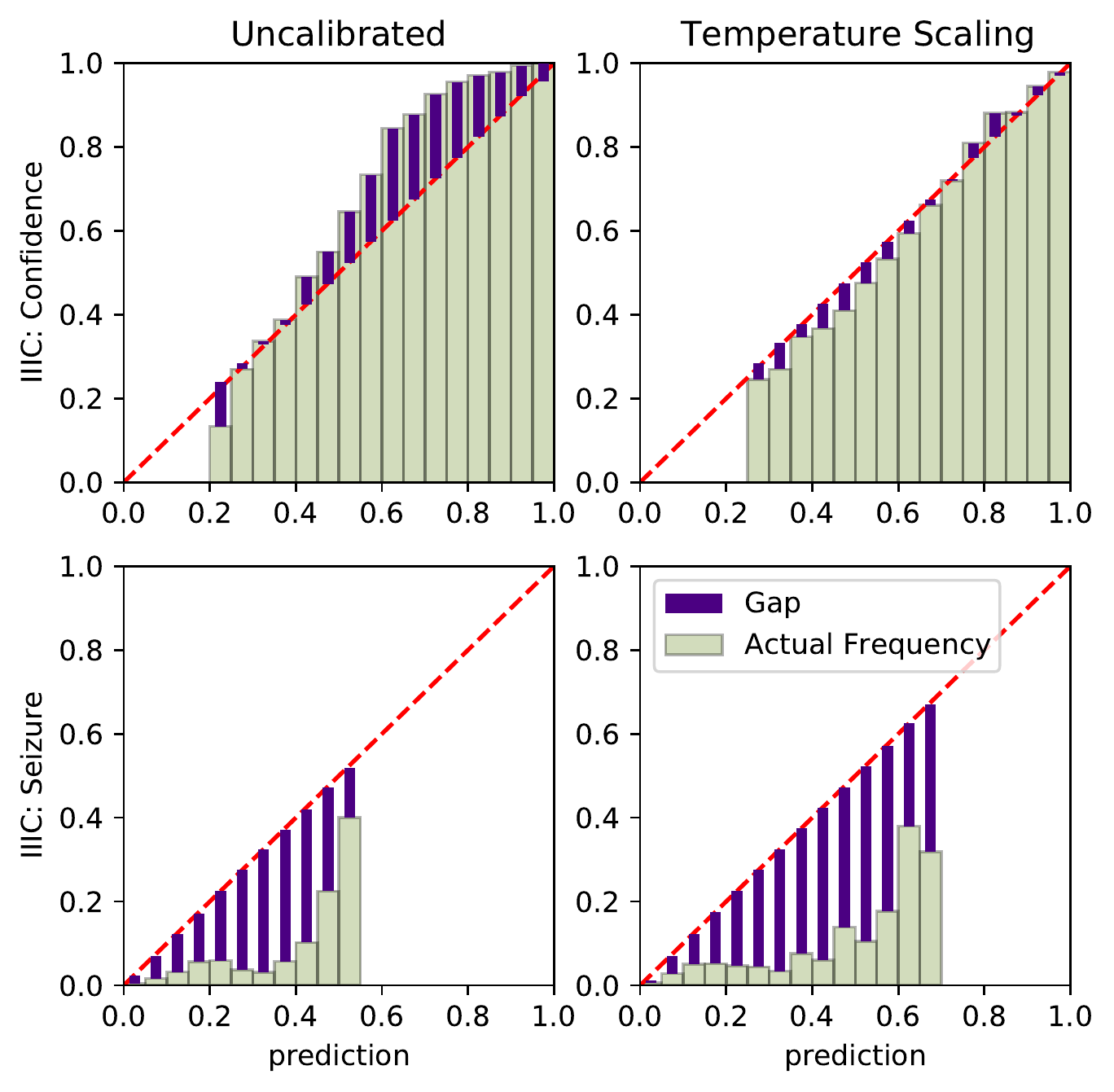}
\vskip -0.1in
\caption{
Reliability diagrams for confidence calibration (top) and Seizure (bottom).
The popular temperature scaling (right) only calibrates the confidence, leaving Seizure poorly calibrated. 
See Figure \ref{fig:reliability:seizure_conf} and the Appendix for complete reliability diagrams.% for \methodname and a suite of other popular methods.
}
\label{fig:reliability:intro}
\end{wrapfigure}
}
\FigIntroReliability
However, confidence calibration is far from sufficient.
Doctors need to perform differential diagnoses on a patient, where multiple possible diseases should be considered with proper probabilities for all of them, not only the most likely diagnosis. 
Figure~\ref{fig:reliability:intro} shows an example where the confidence is calibrated, but prediction for important classes like Seizure is poorly calibrated. 
A classifier can be confidence-calibrated but not useful for differential diagnoses if the probability assignments for most diseases are inaccurate.

Recent research effort has started to focus on full calibration, for example, in~\cite{AISTATS2019_CalEval,NeurIPS2019_DirCal,NeurIPS2019_CalEval,DBLP:SoftCal,NeurIPS2020_FocalCal,ICLR2021_AdaptBinCal_MutualInfo}. 
We approach this problem by leveraging the latent neural network embedding in a nonparametric manner.
Nonparametric methods such as histogram binning (HB)~\citep{KDD_HistogramBinning} and isotonic regression (IR)~\citep{KDD_Isotonic}, are natural for calibration and have become popular.
\cite{ICML2021_GuaranteeHistBin} recently showed a calibration guarantee for HB.
However, HB usually leads to noticeable drops in accuracy~\citep{ICLR2021_AdaptBinCal_MutualInfo}, and IR is prone to overfitting~\citep{IsotonicIssue}.
Unlike existing methods, we take one step back and train a new low-dimensional metric space on the penultimate-layer embeddings of DNNs.
Then, we use a kernel density estimation-based classifier to predict the class probabilities directly.
We refer to our \textbf{K}ernel-based \textbf{Cal}ibration method as \methodname.
Unlike most calibration methods,  \methodname provides high probability error bounds for \textit{full calibration} under standard assumptions.
%enjoys an asymptotic \textit{full calibration} guarantee under standard assumptions with finite sample calibration error bounds. 
Empirically, we show that with little overhead, \methodname outperforms all existing calibration methods in terms of calibration quality, across multiple tasks and DNN architectures, while maintaining and sometimes improving the classification accuracy.

%===========================================================================================

\textbf{Summary of Contributions}:
\begin{itemize}
    \item We propose \methodname, a principled method that calibrates DNNs using kernel density estimation on the \textit{latent embeddings}. 
    %This is different from all existing calibration works that use the final predictions (logits or softmax outputs). 
    
    \item We present an efficient pipeline to train \methodname, including a dimension-reducing projection and a stratified sampling method to facilitate efficient training.
    
    \item We provide finite sample bounds for the calibration error of \methodname-calibrated output under standard assumptions. To the best of our knowledge, this is the first method with a full calibration guarantee.
    %Note that \fontred{no} existing calibration methods do \textit{not} have a provable calibration guarantee.
    
    \item In extensive experiments on multiple datasets and state-of-the-art models, we found that \methodname outperforms existing calibration methods in commonly used evaluation metrics.
    We also show that \methodname provides more reliable predictions for important classes in the healthcare datasets.

\end{itemize}
The code to replicate all our experimental results is submitted along with supplementary materials. 
\vspace{-2mm}

\section{Related Work}\label{sec:related}
Research on calibration originated in the context of meteorology and weather forecasting (see \cite{Murphy1984} for an overview) and has a long history, much older than the field of machine learning~\citep{Brier, Murphy1977, Degroot1983}. We refer to \cite{CalibrationReview} for a holistic overview and focus below on methods proposed in the context of modern neural networks. Based on underlying methodological similarities, we cluster them into distinct categories. 
%\subsection{Calibration}

%Scaling based methods
\noindent{\bf Scaling:} A popular family of calibration methods is based on scaling, in which a mapping is learned from the predicted logits to probability vectors.
Confidence calibration scaling methods include
temperature scaling (\baselineTS)~\citep{ICML2017_Guo} and its antecedent Platt scaling~\citep{Platt99probabilisticoutputs},
an ensemble of \baselineTS~\citep{ICML2020_MixNMatch_KDEEval},
Gaussian-Process scaling~\citep{ICML2020_NonParaCal},
combining a base calibrator (\baselineTS) with a rejection option~\citep{ICML2021_MetaCal}.
Matrix scaling with regularization was also used to perform full calibration~\citep{NeurIPS2019_DirCal}.
While some scaling-based methods can be data-efficient, there are no known theoretical guarantees for them to the best of our knowledge. 

%binning
\noindent{\bf Binning:} Another cluster of solutions relies on binning and its variants, and includes uniform-mass binning~\citep{KDD_HistogramBinning}, scaling before binning~\citep{NEURIPS2019_VerifiedUC}, and mutual-information-maximization-based binning~\citep{ICLR2021_AdaptBinCal_MutualInfo}.
Isotonic regression~\citep{KDD_Isotonic} is also often interpreted as binning.
Uniform-mass binning~\citep{KDD_HistogramBinning} has a distribution-free finite sample calibration guarantee~\citep{ICML2021_GuaranteeHistBin} and asymptotic convergent ECE estimation~\citep{AISTATS2019_CalEval}.
However, in practice, binning tends to decrease accuracy~\citep{ICLR2021_AdaptBinCal_MutualInfo, ICML2017_Guo}.
Binning can also be considered a member of the broader nonparametric calibration family of methods. 
Such methods also include Gaussian Process Calibration~\citep{ICML2020_NonParaCal}, which however also only considers confidence calibration.

\noindent{\bf Loss regularization:} There are also attempts to train a calibrated DNN to begin with.
Such methods typically add a suitable regularizer to the loss function~\citep{DBLP:SoftCal,NeurIPS2020_FocalCal,ICML2018_MMCECal}, which can sometimes result in expensive optimization and reduction in accuracy.

\noindent{\bf Use of Kernels:} Although not directly used for calibration, kernels have also been used for uncertainty quantification for deep learning classification.
In classification with rejection, the k-nearest-neighbors algorithm (kNN), closely related to kernel-based methods, has been used to provide a ``confidence measure'' which is used to make a binary decision (i.e., whether to reject or to predict)~\citep{DBLP:deepKNN,NuerIPS2018_TrustScore}. 
%However, a good ``confidence measure'' is far from being confidence-calibrated, as it only needs to be \textit{monotonic} in accuracy\footnote{The confidence measure could literally be either 0 or 1 (binary).}. 
Recently, continuous kernels have also been used to measure calibration quality or used as regularization during training~\citep{NeurIPS2019_CalEval, ICML2018_MMCECal}.
\cite{ICML2020_MixNMatch_KDEEval} introduced a kernel density estimation (KDE) proxy estimator for estimating ECE.
However, it uses a un-optimized kernel over $\Delta^{K-1}$, and shows the KDE-ECE estimator (but not the calibration map) is consistent.
To the best of our knowledge, use of trained KDE to calibrate predictions hasn't been proposed before. Further, we also provide a bound on the calibration error.

%==============================================================================
%==============================================================================

\section{\methodname: Kernel-based Calibration}\label{sec:method}
In this section, we formally introduce \methodname, study its calibration properties theoretically, and present crucial implementation details and comparisons with other methods. 
Specifically, in Section~\ref{sec:method:inference}, we discuss how to construct (automatically) calibrated predictions for test data using a calibration set $\calibrationset$. 
Doing so requires a well-trained kernel and metric space, and we describe a procedure to train such a kernel in Section~\ref{sec:method:training}. 
In Section~\ref{sec:method:theory}, we show that an appropriate shrinkage rate of the bandwidth ensures that the \methodname prediction is automatically calibrated. Sections~\ref{sec:method:impl} provides implementation details. Finally, in Section~\ref{sec:method:related}, we compare and contrast \methodname with existing methods.

\subsection{Classification with Kernel Density Estimation}\label{sec:method:inference}

Following the calibration literature, we first require a holdout calibration set %\footnote{This is similar to the validation set in many machine learning problems. The model (DNN and the kernel) is not trained on this set of data.}
$\calibrationset=\{X_i, Y_i\}_{i=1}^N$.
In \methodname, we fix a kernel function $\hat{\phi}$ which is learned (the learning procedure is described in Section~\ref{sec:method:training}). For a new datum $X_{N+1}$, the class probability $\phat_k(X_{N+1})$ takes the following form:
{\small
\begin{align}
 \phat_k(X_{N+1};\hat{\phi}, \calibrationset) =\frac{\sum_{(x, y)\in \calibrationset^k} \hat{\phi}(x, X_{N+1})}{\sum_{(x,y)\in \calibrationset} \hat{\phi}(x, X_{N+1})},\label{eq:inference}
\end{align}
}where $\calibrationset^k\defeq \{(x, y) \in \calibrationset| y=k\}$.
The notation $\phat_k(X_{N+1};\hat{\phi}, \calibrationset)$ emphasizes the dependence on $\hat{\phi}$ and $\calibrationset$. However, we will use $\phat_k(X_{N+1})$ when the dependence is clear from context. 

\textbf{Remarks}:
What we have described is essentially the classical nonparametric procedure of applying kernel density estimation for classification.
For a moment, suppose we know the true density function $f_k$ of $\mathcal{P}_k$ (the distribution of all the data in class $k$), and the proportion of class $k$, denoted $\pi_k$ (such that $\sum_{k\in[K]}\pi_k = 1$). 
Then, for any particular $x_0$, using the Bayes rule we get: 
{\small
\begin{align}\label{eq:kde:class_prob}
    \mathbb{P}\{Y=k|X=x_0\} &= \frac{f_k(x_0)\pi_k}{\sum_{{k'}\in[K]} f_{k'}(x_0)\pi_{k'}}.
\end{align}
}Now, replacing $f_k$ with the kernel density estimate $ \hat{f}_k(x_0)\defeq (\sum_{(x,y)\in\calibrationset^k} \hat{\phi}_b(x, x_0))/|\calibrationset^k|$, and the class proportion $\pi_k$ with $\displaystyle \hat{\pi}_k\defeq |\calibrationset^k| / |\calibrationset|$ we get back Eq.~(\ref{eq:inference}).

\subsection{Training}\label{sec:method:training}
Employing an appropriate kernel function $\hat{\phi}$ is crucial for good performance under the kernel density framework. The kernel in turn has a critical reliance on the choice of the underlying metric. To obtain good performance using deep learning learning models, we train a metric space on top of the penultimate layer embeddings. 

%To find the appropriate kernel while keeping the good performance from deep learning models, we train a metric space on top of the penultimate layer embeddings.

To begin, we assume a deep neural network is already trained on $\trainingset=\{X^{train}_i, Y^{train}_i\}_{i=1}^M$. 
We place no limitations on the form of loss function, optimizer, or the model architecture. However, we do require the neural net to compute an embedding before a final prediction layer, which is always the case in modern classification models. 
We denote the embedding function from $\mathcal{X}\mapsto \mathbb{R}^h$ as $\embedder$.%, and the full neural network as $\mathbf{l}^{NN}:\mathcal{X}\to\mathbb{R}^K$ ($\mathbf{l}$ for ``logits'').

Given a base ``mother kernel'' function $\phi$, such as the Radial Basis Function (RBF) kernel, we denote the kernel with bandwidth $b$ as $\phi_b\defeq\frac{1}{b}\phi(\frac{\cdot}{b})$.
We parameterize the learnable kernel as:
\begin{align}
\hat{\phi}(x, x') \defeq    \hat{\phi}_{\proj, \embedder, b}(x, x') \defeq \phi_b(\proj(\embedder(x)) - \proj(\embedder(x'))). \label{eq:training:kernel}
\end{align}
where $\proj:\mathbb{R}^h \mapsto \mathbb{R}^d$ is a dimension-reducing projection parameterized by a shallow MLP (Section~\ref{sec:method:impl}). Since the inference time is linear in $d$, letting $d < h$ also affords computational benefits.

Given that the embedding function $\embedder(x)$ from the neural network is fixed, the only learnable entities are $b$ and $\proj$.
In the training phase, we fix $b=1$, and train $\proj$ using (stochastic) gradient descent and log-loss.
The specific value of $b$ does not matter since it can be folded into $\proj$.
Let us denote $\trainingset^{k} = \{(x,y)\in \trainingset:y=k\}$.
In each iteration, we randomly sample two batches of data from $\trainingset$ - the prediction data, denoted as $\trainingset^B$, to evaluate $\proj$, and ``background'' data for each $k$, denoted as $\mathcal{B}^k$, from $\trainingset^{k}\setminus \trainingset^B$ to construct the KDE classifier.
Then, the prediction for any $x_j$ is given by
{\small
\begin{align}
     \phat_k(x_j;\hat{\phi}, \trainingset \setminus \trainingset^B) \defeq \frac{\sum_{(x, y)\in \mathcal{B}^{k}} \frac{|\trainingset^{k}\setminus \trainingset^B|}{|\mathcal{B}^{k}|} \hat{\phi}(x, x_j)}{\sum_{k', (x,y)\in \mathcal{B}^{k'}} \frac{|\trainingset^{k'}\setminus \trainingset^B|}{|\mathcal{B}^{k'}|} \hat{\phi}(x, x_j)}\label{eq:training:pred}
     \vspace{-1mm}
\end{align}
}where $\hat{\phi}$ is shorthand for $\hat{\phi}_{\proj, \embedder, b=1}$ defined in Eq.~(\ref{eq:training:kernel}).
The log-loss is given formally by
{\small 
\vspace{-2mm}
\begin{align}
    L &= -\frac{1}{B}\sum_{(x,y)\in \trainingset^B} \log \phat_y(x;\hat{\phi}_{\proj, \embedder, 1}, \trainingset \setminus \trainingset^B).
\end{align}
}
\vspace{-5mm}

Finally, we pick a $b=b^*$ on the calibration set $\calibrationset$ using cross-validation. This is because $b$ should be chosen contingent on the sample size (Section~\ref{sec:method:theory}).
Choosing $b$ can be done efficiently (Section~\ref{sec:method:impl}).
Algorithm~\ref{alg:main} summarizes the steps we explicated upon so far.

\subsection{Theoretical Analysis: Calibration Comes Free}\label{sec:method:theory}

In the previous section, we have only described a procedure to improve the prediction accuracy for $\phat$ on $\trainingset$. This section will show that calibration comes free with the $\phat$ obtained using Algorithm~\ref{alg:main}. In particular, we show that as the sample-size for each class in $\calibrationset$ increases, $\phat$ converges to the true frequency vector of $Y$ given the input. 
In interest of smoother presentation, we only state the relevant claims in what follows. Detailed proofs are presented in the Appendix.

To begin, we make a few standard assumptions, such as in~\cite{textbook_MultiVarKDE}, including:
\vspace{-2mm}
\begin{itemize}
    \item
    ($\forall k$) The density on the embedded space, $\proj(\embedder(\mathcal{X}|Y=k))$, denoted as $f_{\proj\circ \embedder, k}$, is square integrable and twice differentiable, with all second order partials bounded, continuous, and square integrable.
    \item $\phi$ is spherically symmetric, with a finite second moment.
\end{itemize}

Lemma~\ref{lemma:admissible} and~\ref{lemma:optimal} focus on an arbitrary class $k$ and ignore the subscript $k$ to the density $f$ for readability. 
We denote the size $|\calibrationset^k|=m$. 
Intuitively, due to the bias-variance trade-off, a suitable bandwidth $b$ will depend on $m$:
A small $b$ reduces bias, but with the finite $m$, a smaller $b$ also leads to increased variance.
Thus, $b$ should go to 0 ``slowly'', which is formally stated below:
\begin{lemma}\label{lemma:admissible}
For almost all $x$, if $b^d m\to \infty$ and $b\to 0$ as $m\to\infty$, then we have
{\small
\begin{align}
    \|\hat{f}_{\proj\circ \embedder, k}(x) - f_{\proj\circ \embedder, k}(x)\|_2 \overset{P}{\to} 0 \text{ as } m \to \infty.\label{eq:admissible}
\end{align}
}
\vspace{-8mm}
\end{lemma}
Here $\hat{f}_{\proj\circ \embedder, k}$ is the estimated $f_{\proj\circ \embedder, k}$ using $\calibrationset$.
Recall that $d$ is the dimension of $\proj(\embedder(\mathcal{X}))$.
We will call such a bandwidth $b$ \textit{admissible}, and we sometimes write $b(m)$ to emphasize the dependence on $m$. The following lemma gives the optimal admissible bandwidth:

\begin{lemma}\label{lemma:optimal}
The optimal bandwidth is $b = \Theta(m^{-\frac{1}{d+4}})$, which leads to the fastest decreasing MSE  (i.e. $\mathbb{E}[\|\hat{f}_{\proj\circ \embedder,k}(x) - f_{\proj\circ \embedder,k}(x)\|_2]$) of $O(m^{-\frac{4}{d+4}})$.
\end{lemma}

Now we are in a position to present the main theoretical results.
In the following, $m$ denotes the rarest class's count ($m\defeq \min_{k}\{|\calibrationset^k|\}$.
Theorem~\ref{thm:main_weak} provides a bound between $\phat$ and the true conditional probability vector on the embedded space $\ptrue(\proj(\embedder(X)))$:
%shows a point-wise convergence of $\phat$ to the true conditional probability vector on the embedded space $\ptrue(\proj(\embedder(X)))$:
\begin{theorem}\label{thm:main_weak}
Fixing $x$ such that the density of $\proj(\embedder(x))$ is positive, with $b(m) = \Theta(m^{-\frac{1}{d+4}})$, for any $\lambda \in (0,2)$:
\vspace{-0.2in}
\begin{align}
    \mathbb{P}\{|\phat_k(x) - \mathbf{p}_k(\proj(\embedder(x)))| &> (3K+1)C m^{\frac{-\lambda}{d+4}}\} \leq Ke^{-Bm^{\frac{4-2\lambda}{d+4}}}\\
    \text{ where } \mathbf{p}_k(\proj(\embedder(x)))&\defeq \mathbb{P}\{Y=k|\proj(\embedder(X)) = \proj(\embedder(x))\}
\end{align}
for some constant $B$ and $C$. 
As a corollary, $\phat(x) \overset{\text{almost surely}}{\to} \ptrue(\proj(\embedder(x))) \text{ as } m \to \infty$.
\end{theorem}
Next, we bound the full calibration error with additional standard assumptions. More specifically, we use and build upon the main uniform convergence result for classical KDE presented in~\cite{ICML2017_UniformConvergence}, to obtain Theorem~\ref{thm:main_strong}:
\begin{theorem}\label{thm:main_strong}
Assume $f_{\proj\circ \embedder,k}$ is $\alpha$-H\"{o}lder continuous and bounded away from $0$ for any $k$. 
For an admissible $b(m)$ with shrinkage rate $\Theta((\frac{\log{m}}{m})^{\frac{1}{d+2\alpha}})$,  for some constants $B$ and $C$ we have:
\vspace{-1mm}
{\small
\begin{align}
    \mathbb{P}\{\sup_{X,k} | \phat_k(X) - \mathbb{P}\{Y=k|\phat(X)\}| > (3K+1) C (\frac{\log{m}}{m})^{\frac{\alpha}{d+2\alpha}}\} \leq K(m^{-1} + m^{-B\frac{2\alpha}{d+2\alpha}m^{\frac{d}{d+2\alpha}}}).
\end{align}
\vspace{-0mm}
}
\end{theorem}

\def \AlgoMainSimplified{
\begin{wrapfigure}[21]{L}{0.5\textwidth}
\begin{minipage}{0.5\textwidth}
\vspace{-0.4in}
\begin{algorithm}[H]
   \caption{Overview of \methodname}
   \label{alg:main}
\textbf{Input}:\\
    $\trainingset$: $\{(X^{train}_i,Y^{train}_i)\}_{i=1}^M$ used to train the NN\\
    $\calibrationset$: $\{(X_i,Y_i)\}_{i=1}^N$ calibration set \\%(not used in NN training)\\
    $\embedder:$ Embedding function $\mathcal{X}\to\mathbb{R}^h$ (trained NN)\\
    $X_{N+1}$: Unseen datum for prediction\\
\textbf{Training (of the projection $\proj$)}:
\begin{algorithmic}
    %\State Initialize $\proj$, a learn-able projection $\mathbb{R}^h\to\mathbb{R}^d$. 
    \State Denote $\trainingset^k\defeq \{(x, y) \in \trainingset| y=k\}$.
    \State Denote $\phi_b$ as a base kernel function (e.g. RBF) with bandwidth $b$.
   \Repeat
   \State Sample $\trainingset^B=\{(x_j, y_j)\}_{j=1}^B$ from $\trainingset$.
   \State Compute $\phat(x_j)$ via Eq.~(\ref{eq:training:pred}).
   \State Loss $l\gets \frac{1}{B} \sum_{j=1}^B  LogLoss(\phat(x_j), y_j)$.
   \State Update $\proj$ with (stochastic) gradient descent.
   \Until{ the loss $l$ does not improve.}
   \State  Set $\hat{\phi}_b \gets \hat{\phi}_{\proj, \embedder, b}$ for inference. % $b=1$ by default.
\end{algorithmic}
\textbf{Inference}:
\begin{algorithmic}
    \State Denote $\calibrationset^k\defeq \{(x, y) \in \calibrationset| y=k\}$.
    \State Tune $b^*$ on $\calibrationset$ by minimizing log loss.
    \State $\phat_k(X_{N+1}) \gets \frac{\sum_{(x, y)\in \calibrationset^k} \hat{\phi}_{b^*}(x, X_{N+1})}{\sum_{(x,y)\in \calibrationset} \hat{\phi}_{b^*}(x, X_{N+1})}$.
\end{algorithmic}
\end{algorithm}
\end{minipage}
\end{wrapfigure}
%\endgroup
%\vspace{-0.25in} 
}
\AlgoMainSimplified

We now proceed to present details pertaining to the efficient implementation of \methodname.

\subsection{Implementation Techniques}\label{sec:method:impl}
\textbf{Efficient Training}: 
As might be immediately apparent, utilizing algorithm~\ref{alg:main} for prediction using full $\trainingset \setminus \trainingset^B$ can be an expensive exercise. In order to afford a training speedup, we consider a random subset from $\trainingset \setminus \trainingset^B$ using a modified stratified sampling. 
Specifically, we take $m$ random samples from each $\trainingset^k$, denoted as $\trainingset^{k, m}$, and
replace the right-hand side of Eq.~\ref{eq:training:pred} with:
%use the following modified prediction formulation :
{\small
\begin{align}
\vspace{-5mm}
    %\phat_k(x_0; \hat{\phi}, \cup_{k'} \trainingset^{k',m}) = 
    \frac{\sum_{(x, y)\in \trainingset^{k, m}} \frac{|\trainingset^k|}{m} \hat{\phi}(x, x_0)}{\sum_{k'\in [K], (x,y)\in \trainingset^{k', m}} \frac{|\trainingset^{k'}|}{m} \hat{\phi}(x, x_0)}.
\end{align}
}
\vskip -0.05in
The re-scaling term $\frac{|\trainingset^k|}{m}$ is crucial to get an unbiased estimate of $\hat{f}_k\hat{\pi}_k$. The stratification employed makes the training more stable, while also reducing the estimation variance for rarer classes (more details in Appendix~\ref{appendix:sec:sampling}). The overall complexity is now $O(KmdhB)$ per batch.
In all experiments, we used $m=20$ and $B=64$.
%Note that compared with the basic stratified sampling, this modified version might slightly increase the overall variance, but will reduce the variance of the rarer classes. 

\textbf{Form of $\proj$}: 
While there is considerable freedom in choosing a suitable form for $\proj$, we parameterize $\proj$ with a two layer MLP with a skip connection. % (with batch normalization before each layer).
%Such a parameterization ensures that%
Consequently, $\proj$ can reduce to linear projection when sufficient, and be more expressive when necessary. 
%We first precompute the neural network embeddings and then train $\proj$ using SGD. 
We also experimented with using only a linear projection, the results for which are included in the appendix. 
We fix the output dimension to $d=\min\{dim(\embedder), 32\}$, except for ImageNet ($d=128$).

\textbf{Bandwidth Selection}: Finally, to find the optimal bandwidth using $\calibrationset$, we use Golden-Section search~\citep{GoldenSection} to find the log-loss-minimizing $b^*$.
This takes $O(\log{\frac{ub-lb}{tol}})$ steps where $[lb,ub]$ is the search space, and $tol$ is the tolerance. Essentially, we assume that the loss is a convex function with respect to $b$, permitting an efficient search (see Appendix \ref{appendix:sec:bandwidthsearch}, which presents empirical evidence that the convexity assumption is valid across datasets).

\subsection{Comparisons with Existing Calibration Methods}\label{sec:method:related}
Most existing calibration methods discussed in Section~\ref{sec:related} and \methodname all utilize a holdout calibration set. However, unlike \methodname, existing works usually fix the last neural network layer. \methodname, on the other hand, ``takes a step back'', and replaces the last prediction layer with a kernel density estimation based classifier. Since the DNN $\embedder$ is fixed regardless of whether we use the original last layer or not, we are really comparing a KDE classifier (\methodname) with linear models trained in various ways, after mapping all the data with $\embedder$.
Note that this characterization is true for most existing methods, with a few exceptions (e.g., those summarized under ``loss regularization'' in Section~\ref{sec:related}).

Employing a KDE classifier affords some clear advantages such as a straightforward convergence guarantee and some interpretability\footnote{That is, one could understand how the prediction is made by examining similar samples.}. Furthermore, \methodname can also be improved in an online fashion, a benefit especially desirable in certain high-stakes applications such as in healthcare. For example, a hospital can calibrate a trained model prior to deployment using its own patient data (which is usually not available to train the original model) as it becomes available.

Another important advantage of \methodname is concerning normalization.
In fact, simultaneously calibrating all classes while satisfying the constraint that $\phat\in\Delta^{K-1}$ is a distinguishing challenge for multi-class calibration.
Many calibration methods perform one-vs-rest calibration for each class, and require a separate normalization step at test time~\citep{KDD_HistogramBinning, KDD_Isotonic, ICLR2021_AdaptBinCal_MutualInfo, SplineBaseline}.
This creates a gap between training and testing and could lead to drastic drop in performance (Section~\ref{sec:exp}).  
On the other hand, \methodname automatically satisfies $\phat\in\Delta^{K-1}$, and the normalization is consistent during training and testing.

%One disadvantage of \methodname, which it also shares with most existing methods, is that it requires at least \textit{some} calibration data.\footnote{Alternatively, one could use the training data, but the theoretical guarantee will likely be not applicable.}
%Another potential 
A disadvantage of \methodname is the need to remember the $\proj(\embedder(\calibrationset))$ used to generate the KDE prediction. This is however mitigated to a large extent by the dimension reduction step, which already reduces the computational overhead significantly\footnote{Experiments about the effect of $d$ on performance and overhead are provided in the Appendix.}. For example, in one of our experiments on CIFAR-100, there are 160K (5K images, $d=32$) scalars to remember, which is only 0.2\% of the parameters (85M+) of the original DNN (ViT-base-patch16).
Moreover, KDE inference is trivial to parallelize on GPUs. There is also a rich, under-explored, literature to further speed up the inference. Examples include, KDE merging~\citep{KDESpeed_Brain}, Dual-Tree~\citep{KDESpeed_DualTree}, and Kernel Herding~\citep{KDESpeed_KernelHerding}. 
%\fontred{Random Sampling, Coresets, etc.}.
%https://arxiv.org/pdf/1912.02283.pdf
These methods can easily be used in conjunction with \methodname.

%\textbf{Overfitting}:

%TODO: Online setting is a benefit

%=================================================================================================
\section{Experiments}\label{sec:exp}

\subsection{Data and Neural Networks}
We utilize two sets of data: computer vision benchmarks on which previous calibration methods were tested, and health monitoring datasets where \textit{full} calibration is crucial for diagnostic applications. 
Table~\ref{tab:main:data} summarizes the datasets and their splits.
\def \TableData{
\begin{table}[ht]
%\vskip 0.2in
\vspace{-0.2in} 
\caption{
Dataset summary: Splits and number of classes ($K$).
}
\label{tab:main:data}
\begin{center}
\begin{footnotesize}

\scalebox{0.8}{
\begin{tabular}{l|ccccc|cccc}
\toprule
%& \multicolumn{5}{c|}{Health}&  \multicolumn{4}{c}{Benchmark}\\
%\midrule
Dataset & IIIC & IIIC(pat) & ISRUC & ISRUC(pat) & PN2017 & C10 & C100 & SVHN & ImageNet\\
\midrule
Train & 103,818 & 1,936 & 61,841 & 69 & 15,087 & 45,000, & 45,000 & 65,931 & 1,281,167\\
Calibration & 1,787 & 77 & 1,372 & 6 & 253 & 5,000 & 5,000 & 7,326 & 25,000\\
Test & 33,953 & 684 & 26,070 & 24 & 4,813 & 10,000 & 10,000 & 26,032 & 25,000\\
$K$ & 6 & 6 & 5 & 5 & 4 & 10 & 100 & 10 & 1,000\\
\bottomrule
\end{tabular}
}
\end{footnotesize}
\end{center}
\vskip -0.2in
\end{table}
}
\TableData

\noindent{\bf Benchmark data}
Following~\cite{NeurIPS2019_DirCal}, we use multiple image benchmark datasets, including CIFAR-10, CIFAR-100, and SVHN~\citep{CIFAR,SVHN}.
We reserve 10\% of the training data as the calibration set. 
We fine-tune pretrained ViT~\citep{ViT} and MLP-Mixer (Mixer)~\citep{MLP_Mixer} from the \texttt{timm} library~\citep{rw2019timm}.
We chose ViT and Mixer because they are the state-of-the-art neural architectures in computer vision, and accuracy should come before calibration quality.
We also included the ImageNet dataset~\citep{ImageNet} and use the pretrained Inception ResNet V2~\citep{InceptV2} following~\cite{ICLR2021_AdaptBinCal_MutualInfo}.

\noindent{\bf Health monitoring data}
We also use three health monitoring datasets for diagnostic tasks:
IIIC~\citep{IIICData}, an ictal-interictal-injury-continuum (IIIC) patterns classification dataset;
ISRUC~\citep{ISRUC}, a sleep staging (classification) dataset using polysomnographic (PSG) recordings;
PN2017 (2017 PhysioNet Challenge)~\citep{Clifford2017AF2017, Goldberger2000PhysioBankSignals}, a public electrocardiogram (ECG) dataset for rhythm (particularly Atrial Fibrillation) classification.
For the training set, we follow~\cite{Hong2019Mina,IIICData} for PN2017 and IIIC, and used 69 patients' data for ISRUC.
For the remaining data, 5\% is used as the calibration set and 95\% for testing. 
We perform additional experiments after splitting into training/calibration/test sets by patients for IIIC and ISRUC\footnote{PN2017 did not provide patient IDs,  so we cannot split by patient.}, marked as the ``pat'' version in tables. 
The calibration/test split is 20/80 in ``IIIC (pat)'' and ``ISRUC (pat)'' because the number of patients is small.
For IIIC and ISRUC, we follow the standard practice and train a CNN (ResNet) on the spectrogram~\citep{Sleepnet,10.3389/fneur.2019.00806,SeizureEEGSpectrogram}.
For PN2017, we used a top-performing model from the 2017 PhysioNet Challenge, MINA~\citep{Hong2019Mina}.

\def \TableExpMainAcc{
\begin{table}[ht]
\vspace{-0.1in}
\caption{Accuracy in \% ($\uparrow$ means higher=better).
Accuracy numbers lower than the uncalibrated predictions are in \accfail{dark red} and the best are in \textbf{bold} (both at p=0.01).
\methodname typically improves or maintains the accuracy.
}
\label{tab:main:acc}
\centering
\vspace{1.5mm}
\begin{footnotesize}
\scalebox{0.65}{
\begin{tabular}{p{0.14\textwidth}c|cccccccc|r}
%\begin{tabular}{lp{0.105\textwidth}|p{0.105\textwidth}p{0.105\textwidth}p{0.105\textwidth}p{0.105\textwidth}p{0.105\textwidth}p{0.105\textwidth}p{0.105\textwidth}|p{0.105\textwidth}}
\toprule
  Accuracy $\uparrow$  & \baselineUnCal & \baselineTS & \baselineDirCal & \baselineIMax & \baselineFocal & \baselineSpline & \baselineIOP & \baselineGP & \baselineMMCE & \methodname\\ 
\midrule
IIIC (pat) & 58.68$\pm$1.42 & 58.68$\pm$1.42 & \textbf{63.17$\pm$1.42} & 57.20$\pm$1.32 & \accfail{54.35$\pm$1.64} & 58.51$\pm$1.32 & 58.68$\pm$1.42 & 58.68$\pm$1.42 & 58.05$\pm$1.37 & \textbf{61.67$\pm$2.22}\\
IIIC & 58.53$\pm$0.06 & 58.53$\pm$0.06 & 63.80$\pm$0.10 & \accfail{56.96$\pm$0.14} & \accfail{54.41$\pm$0.05} & \accfail{58.36$\pm$0.20} & 58.53$\pm$0.06 & 58.52$\pm$0.06 & \accfail{58.06$\pm$0.04} & \textbf{66.32$\pm$0.21}\\
ISRUC (pat) & 75.11$\pm$0.77 & 75.11$\pm$0.77 & \textbf{75.57$\pm$0.91} & \textbf{75.54$\pm$0.68} & \accfail{73.79$\pm$0.72} & 75.11$\pm$0.79 & 75.11$\pm$0.77 & 75.11$\pm$0.76 & \textbf{76.26$\pm$0.59} & \textbf{76.13$\pm$0.89}\\
ISRUC & 74.66$\pm$0.08 & 74.66$\pm$0.08 & 76.08$\pm$0.16 & 75.15$\pm$0.07 & \accfail{73.34$\pm$0.09} & 74.69$\pm$0.09 & 74.66$\pm$0.08 & 74.66$\pm$0.09 & 75.95$\pm$0.07 & \textbf{77.45$\pm$0.16}\\
PN2017 & 54.67$\pm$0.14 & 54.67$\pm$0.14 & \textbf{60.00$\pm$0.22} & 57.55$\pm$0.39 & \accfail{13.78$\pm$0.13} & 55.11$\pm$0.84 & 55.15$\pm$1.48 & 54.69$\pm$0.15 & \accfail{51.90$\pm$0.07} & \textbf{60.36$\pm$0.61}\\
C10 (ViT) & \textbf{98.94$\pm$0.05} & \textbf{98.94$\pm$0.05} & \textbf{98.94$\pm$0.05} & \textbf{98.94$\pm$0.05} & \accfail{98.76$\pm$0.06} & \textbf{98.94$\pm$0.05} & \textbf{98.94$\pm$0.05} & \textbf{98.94$\pm$0.06} & \textbf{98.93$\pm$0.07} & \textbf{98.98$\pm$0.09}\\
C10 (Mixer) & \textbf{98.17$\pm$0.08} & \textbf{98.17$\pm$0.08} & \accfail{98.03$\pm$0.09} & \textbf{98.13$\pm$0.08} & \accfail{96.98$\pm$0.08} & \textbf{98.17$\pm$0.08} & \textbf{98.17$\pm$0.08} & \textbf{98.16$\pm$0.08} & \textbf{98.15$\pm$0.06} & \textbf{98.14$\pm$0.06}\\
C100 (ViT) & 92.09$\pm$0.16 & 92.09$\pm$0.16 & 92.08$\pm$0.14 & 91.95$\pm$0.17 & \accfail{91.21$\pm$0.12} & 92.09$\pm$0.16 & 92.09$\pm$0.16 & 92.09$\pm$0.16 & \textbf{92.41$\pm$0.17} & \textbf{92.37$\pm$0.15}\\
C100 (Mixer) & 87.53$\pm$0.20 & 87.53$\pm$0.20 & \accfail{87.24$\pm$0.22} & \accfail{87.10$\pm$0.21} & \accfail{86.49$\pm$0.23} & 87.53$\pm$0.20 & 87.53$\pm$0.20 & 87.51$\pm$0.20 & \textbf{88.13$\pm$0.25} & 87.55$\pm$0.16\\
SVHN (ViT) & 95.93$\pm$0.05 & 95.93$\pm$0.05 & 95.93$\pm$0.05 & \accfail{95.85$\pm$0.06} & \accfail{95.70$\pm$0.08} & 95.93$\pm$0.05 & 95.93$\pm$0.05 & 95.93$\pm$0.05 & \textbf{96.48$\pm$0.04} & \textbf{96.42$\pm$0.05}\\
SVHN (Mixer) & 95.85$\pm$0.04 & 95.85$\pm$0.04 & 95.98$\pm$0.04 & 95.85$\pm$0.05 & \accfail{95.24$\pm$0.04} & 95.85$\pm$0.04 & 95.85$\pm$0.04 & 95.85$\pm$0.05 & \accfail{95.58$\pm$0.05} & \textbf{96.10$\pm$0.04}\\
ImageNet & \textbf{80.44$\pm$0.24} & \textbf{80.44$\pm$0.24} & \accfail{79.55$\pm$0.24} & \textbf{80.34$\pm$0.28} & \textendash & \textbf{80.22$\pm$0.27} & \textbf{80.44$\pm$0.24} & \textbf{80.44$\pm$0.24} & \textendash & \accfail{79.64$\pm$0.24}\\
\bottomrule
\end{tabular}
}
\end{footnotesize}
\vspace{-0.1in}
\end{table}
}
\TableExpMainAcc

\def \TableExpMainCECEtAdapt{
\begin{table}[ht]
\caption{Class-wise ECE in $10^{-2}$ ($\downarrow$ means lower=better).
%The overall best method is \underline{underscored}. 
The best accuracy-preserving method is in \textbf{bold} (p=0.01).
The lowest but not accuracy-preserving number is \underline{underscored}.
\methodname almost always achieves the lowest class-wise ECE, while maintaining accuracy.
%All numbers are presented in percentage for readability.
}
\label{tab:main:cecet_adapt}
%\begin{center}
\centering
\vspace{1.5mm}
\begin{small}
\scalebox{0.69}{
%\begin{tabular}{lc|ccccccc|r}
\begin{tabular}{p{0.14\textwidth}p{0.1\textwidth}|p{0.1\textwidth}p{0.1\textwidth}p{0.1\textwidth}p{0.1\textwidth}p{0.1\textwidth}p{0.1\textwidth}p{0.1\textwidth}p{0.1\textwidth}|p{0.1\textwidth}}
\toprule
  CECE $\downarrow$  & \baselineUnCal & \baselineTS & \baselineDirCal & \baselineIMax & \baselineFocal & \baselineSpline & \baselineIOP & \baselineGP & \baselineMMCE & \methodname\\ 
\midrule
IIIC (pat) & 8.07$\pm$0.27 & 8.97$\pm$0.85 & \textbf{5.13$\pm$1.48} & 9.23$\pm$0.98 & 8.99$\pm$0.53 & 8.56$\pm$0.62 & 8.33$\pm$0.50 & 7.95$\pm$0.64 & 7.12$\pm$0.43 & \textbf{4.68$\pm$1.27}\\
IIIC & 7.96$\pm$0.02 & 8.96$\pm$0.52 & \textbf{2.24$\pm$0.13} & 8.76$\pm$0.26 & 8.78$\pm$0.02 & 8.43$\pm$0.21 & 8.01$\pm$0.25 & 7.52$\pm$0.23 & 6.70$\pm$0.25 & \textbf{2.03$\pm$0.26}\\
ISRUC (pat) & \textbf{4.48$\pm$0.24} & \textbf{4.69$\pm$0.76} & \textbf{4.18$\pm$0.90} & 8.56$\pm$1.00 & 9.23$\pm$0.21 & \textbf{4.68$\pm$0.46} & \textbf{4.60$\pm$0.60} & \textbf{4.64$\pm$0.43} & \textbf{4.08$\pm$0.36} & \textbf{3.82$\pm$1.24}\\
ISRUC & 4.49$\pm$0.02 & 5.17$\pm$0.77 & 2.71$\pm$0.40 & 9.22$\pm$0.85 & 9.05$\pm$0.03 & 4.73$\pm$0.15 & 4.67$\pm$0.36 & 4.67$\pm$0.27 & 4.10$\pm$0.22 & \textbf{1.90$\pm$0.28}\\
PN2017 & 12.17$\pm$0.07 & 12.31$\pm$0.23 & \textbf{4.30$\pm$0.47} & 9.92$\pm$1.16 & 17.31$\pm$0.09 & 8.61$\pm$0.73 & 12.09$\pm$0.34 & 12.17$\pm$0.07 & 12.35$\pm$0.39 & \textbf{4.25$\pm$1.26}\\
C10 (ViT) & 3.19$\pm$0.01 & 0.76$\pm$0.04 & 0.83$\pm$0.06 & \textbf{0.68$\pm$0.05} & 4.82$\pm$0.07 & 0.90$\pm$0.04 & 0.81$\pm$0.06 & \textbf{0.74$\pm$0.06} & 1.11$\pm$0.27 & \textbf{0.74$\pm$0.07}\\
C10 (Mixer) & 3.11$\pm$0.02 & 1.45$\pm$0.12 & 1.23$\pm$0.10 & \textbf{1.24$\pm$0.17} & 6.70$\pm$0.03 & 1.28$\pm$0.09 & 1.30$\pm$0.07 & \textbf{1.21$\pm$0.07} & 1.43$\pm$0.19 & \textbf{1.17$\pm$0.10}\\
C100 (ViT) & 5.90$\pm$0.05 & 5.27$\pm$0.20 & 4.64$\pm$0.13 & 4.96$\pm$0.17 & 5.53$\pm$0.06 & \textbf{4.41$\pm$0.14} & 4.72$\pm$0.12 & 4.65$\pm$0.16 & \textbf{4.27$\pm$0.23} & \textbf{4.32$\pm$0.10}\\
C100 (Mixer) & 5.39$\pm$0.04 & 5.82$\pm$0.17 & 5.25$\pm$0.14 & 5.79$\pm$0.24 & 5.72$\pm$0.05 & 4.92$\pm$0.18 & 5.34$\pm$0.23 & 5.09$\pm$0.15 & 5.26$\pm$0.19 & \textbf{4.62$\pm$0.10}\\
SVHN (ViT) & 3.37$\pm$0.01 & 2.31$\pm$0.56 & \textbf{1.22$\pm$0.06} & 2.64$\pm$0.20 & 5.89$\pm$0.03 & 1.34$\pm$0.05 & 1.39$\pm$0.06 & 1.40$\pm$0.05 & 1.47$\pm$0.11 & \textbf{1.23$\pm$0.10}\\
SVHN (Mixer) & 3.20$\pm$0.01 & 3.06$\pm$0.61 & \textbf{1.21$\pm$0.12} & 2.64$\pm$0.17 & 5.59$\pm$0.02 & 1.45$\pm$0.09 & 1.44$\pm$0.06 & 1.46$\pm$0.06 & 1.64$\pm$0.13 & 1.40$\pm$0.08\\
ImageNet & 2.96$\pm$0.02 & 3.25$\pm$0.07 & 5.60$\pm$0.23 & 2.82$\pm$0.19 & \textendash & \textbf{2.17$\pm$0.06} & 2.30$\pm$0.14 & 2.42$\pm$0.06 & \textendash & \underline{1.94$\pm$0.04}\\
\bottomrule
\end{tabular}
}
\end{small}
%\end{center}
\end{table}
}
\TableExpMainCECEtAdapt

\def \TableExpMainECEAdapt{
\begin{table}[ht]
\caption{ECE in $10^{-2}$ ($\downarrow$ means lower=better).
%All numbers are presented in percentage for readability.
%The overall best method is \underline{underscored}. 
The best accuracy-preserving method is in \textbf{bold} (p=0.01).
The lowest but not accuracy-preserving number is \underline{underscored}.
\methodname is usually on par or better than the best baseline.
}
\label{tab:main:ece_adapt}
%\begin{center}
\centering
\vspace{1.5mm}
\begin{small}
\scalebox{0.67}{
\begin{tabular}{lc|cccccccc|c}
\toprule
 ECE $\downarrow$  & \baselineUnCal & \baselineTS & \baselineDirCal & \baselineIMax & \baselineFocal & \baselineSpline & \baselineIOP & \baselineGP & \baselineMMCE & \methodname\\ 
\midrule
IIIC (pat) & 9.32$\pm$1.01 & \textbf{5.00$\pm$2.75} & \textbf{2.92$\pm$1.59} & 10.52$\pm$4.05 & 7.53$\pm$0.55 & \textbf{4.58$\pm$2.04} & \textbf{4.57$\pm$2.14} & \textbf{3.86$\pm$1.63} & 6.33$\pm$3.28 & \textbf{4.34$\pm$1.35}\\
IIIC & 9.28$\pm$0.03 & 4.45$\pm$1.52 & \textbf{1.39$\pm$0.19} & 10.16$\pm$0.81 & 7.25$\pm$0.05 & 3.20$\pm$0.64 & 3.50$\pm$0.41 & \textbf{1.80$\pm$0.49} & 4.78$\pm$2.24 & 2.62$\pm$0.59\\
ISRUC (pat) & 3.59$\pm$0.32 & \textbf{2.73$\pm$1.53} & 2.97$\pm$0.97 & 8.86$\pm$1.39 & 14.88$\pm$0.43 & \textbf{1.98$\pm$0.35} & \textbf{2.45$\pm$1.36} & \textbf{2.00$\pm$0.53} & \textbf{2.12$\pm$0.93} & \textbf{2.78$\pm$1.25}\\
ISRUC & 3.46$\pm$0.06 & 3.82$\pm$1.69 & 2.27$\pm$0.69 & 9.58$\pm$1.23 & 14.70$\pm$0.06 & \textbf{1.50$\pm$0.53} & 2.71$\pm$0.96 & 2.09$\pm$0.74 & \textbf{2.12$\pm$1.03} & \textbf{1.36$\pm$0.41}\\
PN2017 & 16.70$\pm$0.22 & 16.99$\pm$0.73 & \textbf{5.64$\pm$0.75} & 10.40$\pm$1.35 & 24.63$\pm$0.13 & \textbf{6.84$\pm$2.09} & 16.07$\pm$2.03 & 16.66$\pm$0.21 & 13.49$\pm$1.07 & \textbf{4.78$\pm$1.48}\\
C10 (ViT) & 9.15$\pm$0.05 & 0.75$\pm$0.11 & 0.40$\pm$0.04 & 0.51$\pm$0.07 & 7.17$\pm$0.07 & 0.39$\pm$0.08 & 0.39$\pm$0.04 & \textbf{0.21$\pm$0.06} & \textbf{0.42$\pm$0.29} & 0.40$\pm$0.05\\
C10 (Mixer) & 9.04$\pm$0.06 & 1.06$\pm$0.12 & 0.61$\pm$0.07 & 0.91$\pm$0.14 & 12.53$\pm$0.06 & \textbf{0.36$\pm$0.06} & 0.66$\pm$0.09 & \textbf{0.34$\pm$0.10} & 0.91$\pm$0.44 & 0.59$\pm$0.09\\
C100 (ViT) & 11.64$\pm$0.14 & 2.77$\pm$0.46 & \textbf{0.74$\pm$0.16} & 3.28$\pm$0.22 & 9.97$\pm$0.09 & 1.08$\pm$0.18 & 1.07$\pm$0.19 & \textbf{0.88$\pm$0.11} & 1.05$\pm$0.30 & 1.50$\pm$0.32\\
C100 (Mixer) & 13.71$\pm$0.15 & 3.03$\pm$0.34 & \underline{1.06$\pm$0.28} & 4.75$\pm$0.27 & 14.35$\pm$0.21 & \textbf{1.25$\pm$0.29} & 1.70$\pm$0.66 & \textbf{1.08$\pm$0.26} & 1.93$\pm$0.49 & 3.07$\pm$0.49\\
SVHN (ViT) & 10.10$\pm$0.05 & \textbf{2.43$\pm$2.72} & \textbf{0.60$\pm$0.07} & 2.05$\pm$0.18 & 12.17$\pm$0.08 & 0.74$\pm$0.10 & \textbf{0.62$\pm$0.08} & \textbf{0.64$\pm$0.07} & \textbf{0.72$\pm$0.21} & \textbf{0.64$\pm$0.12}\\
SVHN (Mixer) & 10.29$\pm$0.04 & 3.19$\pm$2.55 & \textbf{0.66$\pm$0.05} & 2.13$\pm$0.10 & 11.09$\pm$0.06 & 0.78$\pm$0.11 & \textbf{0.60$\pm$0.08} & 0.72$\pm$0.06 & 0.72$\pm$0.28 & 0.73$\pm$0.10\\
ImageNet & 3.21$\pm$0.15 & 3.52$\pm$0.13 & 4.30$\pm$0.68 & 7.97$\pm$0.35 & \textendash & 1.10$\pm$0.20 & 1.31$\pm$0.47 & \textbf{0.87$\pm$0.12} & \textendash & 1.43$\pm$0.34\\
\bottomrule
\end{tabular}
}
\end{small}
%\end{center}
\end{table}
}
\TableExpMainECEAdapt

\def \TableExpMainBrier{
\begin{table}[ht]
\caption{
Brier Score in $10^{-2}$ ($\downarrow$ means lower=better).
The best accuracy-preserving methods are in \textbf{bold} (p=0.01).
The lowest but not accuracy-preserving number is \underline{underscored}.
%\baselineFocal performs well on PN2017 because both the confidence and accuracy is very low.
}
\label{tab:main:brier}
\centering
\vspace{1.5mm}
\begin{small}
\scalebox{0.69}{
%\begin{tabular}{lc|ccccccc|r}
\begin{tabular}{p{0.14\textwidth}p{0.1\textwidth}|p{0.1\textwidth}p{0.1\textwidth}p{0.1\textwidth}p{0.1\textwidth}p{0.1\textwidth}p{0.1\textwidth}p{0.1\textwidth}p{0.1\textwidth}|p{0.1\textwidth}}
\toprule
 Brier $\downarrow$ & \baselineUnCal & \baselineTS & \baselineDirCal & \baselineIMax & \baselineFocal & \baselineSpline & \baselineIOP & \baselineGP & \baselineMMCE & \methodname\\ 
\midrule
IIIC (pat) & 21.30$\pm$0.25 & 20.70$\pm$0.69 & \textbf{18.94$\pm$0.55} & 21.09$\pm$1.29 & 21.48$\pm$0.19 & 20.43$\pm$0.50 & 20.52$\pm$0.58 & 20.33$\pm$0.42 & 21.11$\pm$0.71 & \textbf{19.33$\pm$0.78}\\
IIIC & 21.35$\pm$0.01 & 20.62$\pm$0.27 & 18.33$\pm$0.04 & 20.83$\pm$0.19 & 21.46$\pm$0.01 & 20.21$\pm$0.09 & 20.39$\pm$0.09 & 20.05$\pm$0.08 & 20.86$\pm$0.26 & \textbf{17.54$\pm$0.10}\\
ISRUC (pat) & 15.26$\pm$0.25 & 15.20$\pm$0.31 & 15.37$\pm$0.38 & 16.25$\pm$0.49 & 18.55$\pm$0.18 & 15.11$\pm$0.26 & 15.16$\pm$0.31 & 15.16$\pm$0.29 & \textbf{14.69$\pm$0.22} & \textbf{14.97$\pm$0.29}\\
ISRUC & 15.46$\pm$0.03 & 15.50$\pm$0.19 & 15.07$\pm$0.09 & 16.62$\pm$0.33 & 18.77$\pm$0.01 & 15.31$\pm$0.05 & 15.39$\pm$0.10 & 15.35$\pm$0.06 & 14.91$\pm$0.08 & \textbf{14.28$\pm$0.08}\\
PN2017 & 26.61$\pm$0.05 & 26.74$\pm$0.27 & \textbf{22.44$\pm$0.15} & 24.58$\pm$0.59 & \underline{17.79$\pm$0.03} & 23.28$\pm$0.37 & 26.39$\pm$0.69 & 26.61$\pm$0.05 & 26.41$\pm$0.44 & \textbf{22.56$\pm$0.28}\\
C10 (ViT) & 1.76$\pm$0.03 & 0.89$\pm$0.06 & \textbf{0.78$\pm$0.04} & 0.84$\pm$0.04 & 1.75$\pm$0.03 & \textbf{0.79$\pm$0.04} & \textbf{0.79$\pm$0.04} & \textbf{0.78$\pm$0.04} & 0.85$\pm$0.10 & \textbf{0.75$\pm$0.05}\\
C10 (Mixer) & 2.29$\pm$0.03 & 1.48$\pm$0.07 & 1.42$\pm$0.05 & 1.46$\pm$0.08 & 4.16$\pm$0.04 & 1.39$\pm$0.04 & 1.40$\pm$0.05 & \textbf{1.37$\pm$0.04} & \textbf{1.45$\pm$0.16} & \textbf{1.34$\pm$0.04}\\
C100 (ViT) & 6.94$\pm$0.08 & 5.35$\pm$0.15 & 5.17$\pm$0.10 & 5.48$\pm$0.14 & 6.93$\pm$0.07 & 5.19$\pm$0.09 & 5.18$\pm$0.10 & 5.14$\pm$0.09 & \textbf{4.81$\pm$0.10} & 5.01$\pm$0.08\\
C100 (Mixer) & 10.15$\pm$0.11 & 7.94$\pm$0.17 & 7.82$\pm$0.12 & 8.23$\pm$0.17 & 10.91$\pm$0.08 & 7.76$\pm$0.12 & 7.82$\pm$0.15 & 7.72$\pm$0.13 & \textbf{7.38$\pm$0.16} & 7.61$\pm$0.09\\
SVHN (ViT) & 3.99$\pm$0.03 & 3.03$\pm$0.34 & 2.78$\pm$0.04 & 2.99$\pm$0.07 & 5.03$\pm$0.03 & 2.80$\pm$0.03 & 2.79$\pm$0.04 & 2.79$\pm$0.04 & \textbf{2.43$\pm$0.02} & 2.49$\pm$0.03\\
SVHN (Mixer) & 4.03$\pm$0.03 & 3.21$\pm$0.36 & 2.77$\pm$0.03 & 3.04$\pm$0.04 & 5.06$\pm$0.04 & 2.84$\pm$0.03 & 2.81$\pm$0.04 & 2.81$\pm$0.04 & 3.03$\pm$0.02 & \textbf{2.68$\pm$0.03}\\
ImageNet & 11.15$\pm$0.14 & 11.20$\pm$0.15 & 12.03$\pm$0.21 & 11.93$\pm$0.18 & \textendash & \textbf{10.68$\pm$0.13} & \textbf{10.69$\pm$0.13} & \textbf{10.67$\pm$0.12} & \textendash & 11.14$\pm$0.10\\
\bottomrule
\end{tabular}
}
\end{small}
\end{table}
}
\TableExpMainBrier

\def \TableExpRanks{
\begin{table}[ht]
\caption{
Ranks for different evaluation metrics.
The best rank is \underline{underscored}.
In general, \methodname consistently outperforms baselines on Accuracy, CECE and Brier, and the difference between most methods on ECE is small. 
}
\label{tab:main:ranks}
\centering
\vspace{1.5mm}
\begin{small}
\scalebox{0.73}{
\begin{tabular}{lc|cccccccc|c}
\toprule
Ranking & \baselineUnCal & \baselineTS & \baselineDirCal & \baselineIMax & \baselineFocal & \baselineSpline & \baselineIOP & \baselineGP & \baselineMMCE & \methodname\\ 
\midrule
ECE & 8.42$\pm$1.43 & 6.68$\pm$1.11 & 3.33$\pm$1.80 & 7.73$\pm$1.55 & 9.39$\pm$0.95 & 3.51$\pm$1.06 & 4.25$\pm$1.35 & \underline{2.91$\pm$1.66} & 4.52$\pm$0.98 & 3.84$\pm$1.35\\
Accuracy & 5.03$\pm$1.30 & 5.03$\pm$1.30 & 4.53$\pm$2.69 & 6.41$\pm$2.36 & 9.99$\pm$0.03 & 5.56$\pm$0.93 & 5.01$\pm$1.27 & 5.64$\pm$1.16 & 4.74$\pm$3.30 & \underline{2.70$\pm$2.01}\\
CECE & 6.99$\pm$1.95 & 7.41$\pm$1.60 & 3.31$\pm$2.08 & 6.82$\pm$2.67 & 9.46$\pm$0.61 & 4.59$\pm$2.06 & 5.12$\pm$1.13 & 4.37$\pm$1.27 & 4.69$\pm$1.99 & \underline{1.83$\pm$0.76}\\
Brier & 8.18$\pm$1.52 & 6.91$\pm$0.85 & 3.86$\pm$2.08 & 7.42$\pm$1.06 & 8.98$\pm$2.67 & 4.23$\pm$1.05 & 4.88$\pm$1.24 & 3.89$\pm$1.83 & 4.11$\pm$2.89 & \underline{2.05$\pm$1.17}\\
Average & 7.16 & 6.51 & 3.76 & 7.09 & 9.46 & 4.47 & 4.81 & 4.20 & 4.51 & \underline{2.61}\\
\bottomrule
\end{tabular}
}
\end{small}
\end{table}
}
\TableExpRanks

\subsection{Baselines Methods}
We compare \methodname with the multiple state-of-the-art calibration methods, including 
Temperature Scaling (\baselineTS)~\citep{ICML2017_Guo}, 
Dirichlet Calibration (\baselineDirCal)~\citep{NeurIPS2019_DirCal}, 
Mutual-information-maximization-based Binning (\baselineIMax)~\citep{ICLR2021_AdaptBinCal_MutualInfo}, 
Gaussian Process Calibration (\baselineGP)~\citep{ICML2020_NonParaCal}, 
Intra Order-preserving Calibration (\baselineIOP)~\citep{IOPBaseline}, 
Splines-based Calibration (\baselineSpline)~\citep{SplineBaseline}, 
Focal-loss-based calibration (\baselineFocal)~\citep{NeurIPS2020_FocalCal},
MMCE-based calibration (\baselineMMCE)~\citep{ICML2018_MMCECal}.

\subsection{Evaluation Metrics}
We report standard evaluation metrics: Accuracy, class-wise expected calibration error (CECE)~\citep{NeurIPS2019_DirCal,ICLR2021_AdaptBinCal_MutualInfo,Nixon_2019_CVPR_Workshops}, expected calibration error (ECE)~\citep{ICML2017_Guo}, and Brier score~\citep{Brier}. 
CECE is typically used as a proxy to evaluate full calibration quality, because directly binning basing on the entire vector $\phat$ requires exponentially (in $K$) many bins.
Similar to~\cite{ICLR2021_AdaptBinCal_MutualInfo, Nixon_2019_CVPR_Workshops}, we ignore all predictions with very small probabilities (less than $\max\{0.01, \frac{1}{K}\}$).
ECE, on the other hand, only measures confidence calibration (Def~\ref{def:conf_cal}).
For both ECE and CECE, we use the ``adaptive'' version with equal number of samples in each bin (with 20 bins), because this is shown to measure the calibration quality better than the equal-width version~\citep{Nixon_2019_CVPR_Workshops}.
Brier score can be viewed as the sum of a ``calibration'' term, and a ``refinement'' term measuring how discriminative a model is~\citep{10.1007/978-3-319-23528-8_5}.
Here we focus on the brier score of the top class.
%could be considered a measure of both the (confidence) calibration and accuracy.
We refer to~\citep{ICML2017_Guo,NeurIPS2019_DirCal,Nixon_2019_CVPR_Workshops} for further discussion on these metrics.

\subsection{Results}
The results are presented in Tables~\ref{tab:main:acc}, \ref{tab:main:cecet_adapt}, \ref{tab:main:ece_adapt} and \ref{tab:main:brier}.
All experiments are repeated 10 times by reshuffling calibration and test sets, and the standard deviations are reported.
For ImageNet, we skipped \baselineFocal and \baselineMMCE because the base NN is given and these methods require training from scratch.
Due to space constraints, we include ablation studies in the Appendix.

In general, \methodname has the best CECE, accuracy and Brier score, and is highly competitive in terms of ECE as well.
Note that \methodname is also the only method with provable calibration guarantee.
\baselineTS is effective in controlling overall ECE but shows little improvement on CECE over \baselineUnCal.
\baselineDirCal often ranks high for the calibration quality but tends to decrease accuracy as $K$ increases. %, although the decrements could be minor.
\baselineDirCal's performance also has a higher cost: Every experiment requires training over hundreds of models with SGD and taking the best ensemble, accounting for most of the experiment computation cost in this paper. 
The amount of tuning suggested for good performance indicates sensitivity to the choice of hyper-parameters, which we have indeed observed to be the case. 
%This might be the reason why some existing works reported that \baselineDirCal gave too poor a performance to be considered~\cite{ICML2021_MetaCal}.
\baselineSpline, \baselineIOP and \baselineGP are similar to \baselineDirCal on vision datasets, but generally perform worse on the healthcare datasets.
%I-Max
In~\cite{ICLR2021_AdaptBinCal_MutualInfo}, \baselineIMax lowers ECE and CECE significantly.
However, it has a critical issue - it does \textit{not} produce a valid probability vector\footnote{It generates a vector whose sum ranges from 0.4 to 2.0 in our experiments. The range is wider for a larger $K$.
}. 
Once normalized, as reported in our experiments, the performance worsens. Since calibrating all the classes simultaneously is \textit{the} distinguishing challenge in multiclass classification, we interpret the observation as: If this normalization constraint is removed, the ``optimization problem'' (to lower calibration error) is much simpler, but the results are invalid hence unusable probability vectors.
\baselineSpline also requires a re-normalization step, but its performance stays consistent.
\baselineFocal is worse than the \baselineUnCal in many experiments.
While calibration performance may improve by combing \baselineFocal with other methods, the drop in accuracy is harder to overcome\footnote{In PN2017, rare classes are oversampled during training~\citep{Hong2019Mina}. 
While this did not cause issues for other calibration methods, the distributional shift at test time seems catastrophic for \baselineFocal.  
}. 
We also observed that for healthcare datasets, being able to tune on a different set of patients boosts the performance significantly.
This is reflected in the accuracy gain for \baselineDirCal and \methodname, and suggests that the embeddings/logits are quite transferable, but the prediction criteria itself can vary from patient to patient.

Finally, we summarize the rankings of all datasets in Table~\ref{tab:main:ranks}.
It is clear that \methodname consistently improves calibration quality for all classes and maintains or improves accuracy.
And if we look at only the confidence prediction (Brier or ECE), \methodname is still highly competitive.
%TODO: 
%\fontred{
%It is interesting that despite Brier is, to some extent, a combination of accuracy and ECE, \methodname actually ranks much better in Brier than either of the two. 
%}
\def \FigSeizureREMReliability{

\begin{figure}[H]
\vskip -0.1in
%\begin{center}
\centering
\includegraphics[width=1\columnwidth]{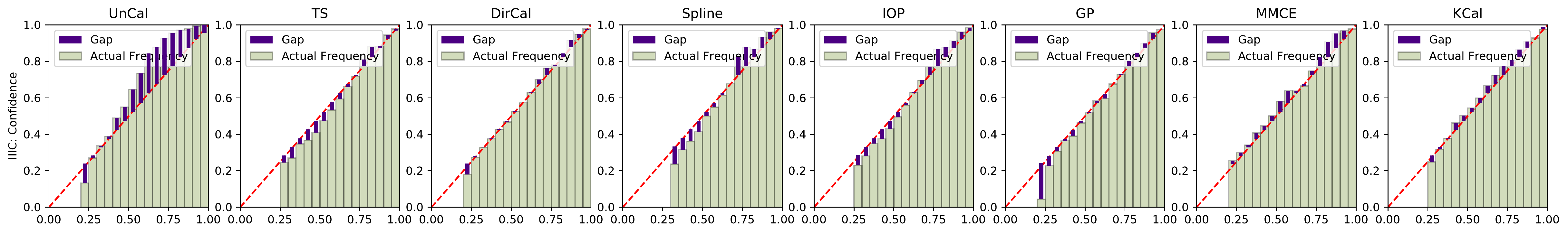}
\includegraphics[width=1\columnwidth]{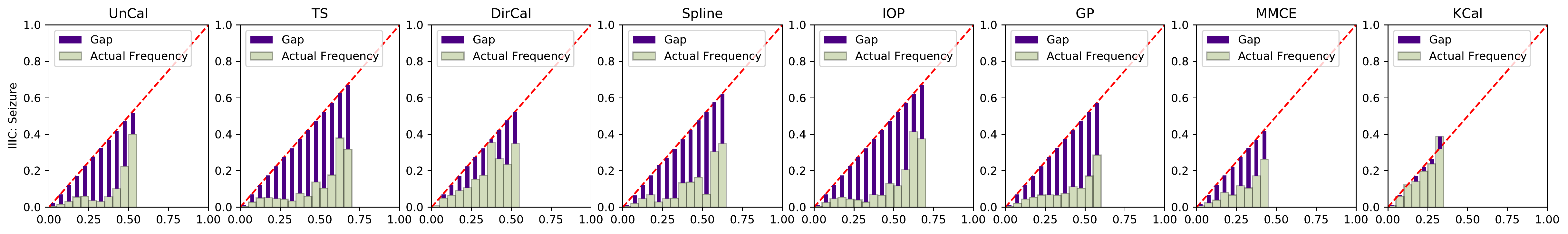}
\vskip -0.1in
\caption{
Reliability diagrams for the predicted class (top) and Seizure (bottom) in IIIC.
All methods calibrate confidence well, but only \methodname achieves reasonable calibration quality for Seizure.
}
\label{fig:reliability:seizure_conf}
%\end{center}
\end{figure}
}
\FigSeizureREMReliability

%====================================================================================================
%\FigIIICSeizureReliability
%\FigISRUCReliability
\subsection{Case Study for Seizure Prediction}
We show the reliability diagrams~\citep{NeurIPS2019_DirCal,ICML2017_Guo} on the IIIC dataset to illustrate the importance of full calibration in Figure~\ref{fig:reliability:seizure_conf}.
We include both the the predicted class (confidence calibration) and Seizure. 
%For ISRUC, we use Rapid Eye Movement (REM), the sleep stage linked to memory consolidation~\citep{REM_Boyce2016CausalEF}.
%For IIIC, we use Seizure, which is the most important but rarest class.
%Figure~\ref{fig:reliability:seizure_rem} shows several reliability diagrams for different predictions for the REM and Seizure class. 
More reliability diagrams can be found in the Appendix, and the results are consistent for all classes. 
The un-calibrated predictions have large gaps for both confidence and Seizure. 
%The un-calibrated predictions tend to be over-confident for lower probability predictions. 
%Further, \baselineTS did not improve the performance much, perhaps because it only has one parameter to tune. 
Most baselines provide calibrated confidence calibration, but fail to calibrated the output for the rare class Seizure.
\methodname, on the other hand, achieves the most consistent results.
We note again that since all competing classes must be considered together for any clinical decision, \textit{full} calibration is indispensable in medical applications.

%==========================================================================
\section{Conclusion}\label{sec:conclusion}

This paper proposed \methodname, a learned-kernel-based calibration method for deep learning models. \methodname consists of a supervised dimensionality reduction step on the penultimate layer neural network embedding to improve efficiency. A KDE classifier using the calibration set is employed in this new metric space. As a natural consequence of the construction, \methodname provides a calibrated probability vector prediction for all classes. Unlike most existing calibration methods, \methodname is also \textit{provably} asymptotically \textit{fully} calibrated with finite sample error bounds. 
We also showed that empirically, it outperforms existing state-of-the-art calibration methods in terms of accuracy and calibration quality.
Moreover, \methodname is more robust to distributional shift, which is common in high-risk applications such as healthcare, where calibration is far more crucial. The major limitation of \methodname is the need to store the entire calibration set, which is a small overhead with the dimension reduction step and potential improvements.%,  as discussed in Sections~\ref{sec:method:impl} and \ref{sec:exp}.

%===========================================End of Main
\bibliography{main}
\bibliographystyle{iclr2023_conference}

%%%%%%%%%%%%%%%%%%%%%%%%%%%%%%%%%%%%%%%%%%%%%%%%%%%%%%%%%%%%
\newpage

%%%%%%%%%%%%%%%%%%%%%%%%%%%%%%%%%%%%%%%%%%%%%%%%%%%%%%%%%%%%%%%%%%%%%%%%%%%%%%%
%%%%%%%%%%%%%%%%%%%%%%%%%%%%%%%%%%%%%%%%%%%%%%%%%%%%%%%%%%%%%%%%%%%%%%%%%%%%%%%
% APPENDIX
%%%%%%%%%%%%%%%%%%%%%%%%%%%%%%%%%%%%%%%%%%%%%%%%%%%%%%%%%%%%%%%%%%%%%%%%%%%%%%%
%%%%%%%%%%%%%%%%%%%%%%%%%%%%%%%%%%%%%%%%%%%%%%%%%%%%%%%%%%%%%%%%%%%%%%%%%%%%%%%
\newpage
\appendix
\onecolumn
%\section{Overview}
\textbf{\begin{center} \LARGE Appendix \end{center}}\hfill \\ \\ 
\textbf{Overview of Appendices:}
Appendix~\ref{appendix:sec:proof} contains proofs for the lemmata and theorems that appear in Section~\ref{sec:method:theory}. 
Appendix~\ref{appendix:sec:sampling} clarifies the benefits of the sampling method (equal-size stratified sampling) described in Section~\ref{sec:method:impl}.
Appendix~\ref{appendix:sec:exp:more} contains more details on the experiments in this paper. Appendix~\ref{appendix:sec:2layer_vs_linear} compares \methodname with a simpler variant, namely \methodname-Linear, which uses a linear layer as the $\proj$.
Appendix~\ref{appendix:sec:dim} explores the effect of $d$, the projected dimension, on the performance and computational overhead.
Finally, Appendix~\ref{appendix:sec:bandwidth} compares the cross-validation-selected bandwidth vs the analytically computed bandwidth, which shows that it is possible to avoid most of the bandwidth selection steps if we use \methodname in an online manner. 

\section{Detailed Assumptions and Proofs}\label{appendix:sec:proof}
\subsection{Assumptions and Definitions}

Denote $Z_i\defeq \proj(\embedder(X_i))$ for $i\in[N+1]$. 
We assume $\{Z_i\}_{i=1}^{N+1}$ are i.i.d.
Since fixing $\proj$ and $\embedder$ using $\trainingset$, all data will now live in $\mathcal{Z}\defeq \proj(\embedder(\mathcal{X}))$. 
We are just performing a standard (multivariate) kernel density estimation with only one parameter $b$ on the calibration set. 
We will use $\hat{g}$ and $g$ to denote the estimation and density in $\mathcal{Z}$, instead of the more cumbersome $\hat{f}_{\proj\circ \embedder,k}$ and $f_{\proj\circ \embedder,k}$.

Like in~\cite{textbook_MultiVarKDE}, we make the following standard assumptions for $g_k(\mathcal{Z})$:
\vspace{-1mm}
\begin{itemize}[nolistsep]
    \item (For any $k$) $g_k$ is square integrable and twice differentiable, with all second order partials bounded, continuous and square integrable. 
\end{itemize}
\vspace{-1mm}
The base ``mother kernel'' function should satisfy the following (true for the RBF kernel):
\vspace{-1mm}
\begin{itemize}[nolistsep]
    \item $\phi$ is spherical symmetric and has a finite second moment.
    Formally, this means $\int_{\mathbb{R}^d} x \phi(x)dx = 
    \mathbf{0}$ and $\forall i\in[d], \int_{\mathbb{R}^d} x_i x_j \phi(x)dx=\mu_{2,\phi}\mathbf{1}\{i=j\}$ where $\mu_{2,\phi}$ is a fixed finite constant.
    %\item $\int_{\mathbb{R}^d} \phi(x) dx = 1$.
    %\item \fontred{$\lim_{\|x\|\to\infty} \phi(x) = 0$}
\end{itemize}
\vspace{-1mm}
In the proof for Lemma~\ref{lemma:admissible} and Lemma~\ref{lemma:optimal}, for simplicity, we ignore the subscript $_k$ and write $g$ instead of $g_k$ where there is no confusion.
%Most kernels widely used, including the RBF kernel we use in this paper, satisfies the above.

\subsection{Proof of Lemma~\ref{lemma:admissible}}\label{appendix:sec:proof:admissible}
Rewriting Eq.~(\ref{eq:admissible}), we want to show $\|\hat{g}(\mathbf{z}) - g(\mathbf{z})\|_2$ converges to $0$ in probability with an admissible $b(m)$, as $m\to\infty$. 
We first derive the expression of the bias and variance of $\hat{g}$. 
For the bias, we have:
\begin{align}
    \mathbb{E}[\hat{g}(\mathbf{z})] - g(\mathbf{z}) &= \frac{1}{b^d}\mathbb{E}\Bigg[\phi\Bigg(\frac{\mathbf{z}-Z}{b}\Bigg)\Bigg] - g(\mathbf{z})\\
    &= \frac{1}{b^d}\int \phi\Bigg(\frac{\mathbf{z}'-\mathbf{z}}{b}\Bigg) g(\mathbf{z}')d\mathbf{z}' - g(\mathbf{z})\\
    &= \int \phi(\mathbf{u}) g(\mathbf{z} + b \mathbf{u}) d\mathbf{u} - g(\mathbf{z}) \\%\text{ (Let $\mathbf{u}=\frac{\mathbf{z}'-\mathbf{z}}{b}$)}\\%Taylor
    &= \int \phi(\mathbf{u}) \Big(
    g(\mathbf{z}) + b(D_g(\mathbf{z}))^\top \mathbf{u} + \frac{1}{2} b^2 \mathbf{u}^\top H_g(\mathbf{z}) \mathbf{u} + o(\|b\mathbf{u}\|^2)
    \Big)d\mathbf{u} - g(\mathbf{z}) \\
    %&\approx \int \phi(\mathbf{u}) \Big(g(\mathbf{z}) + b(D_g(\mathbf{z}))^\top \mathbf{u} + \frac{1}{2} b^2 \mathbf{u}^\top H_g(\mathbf{z}) \mathbf{u}\Big)d\mathbf{u} - g(\mathbf{z}) \\
    &= \int \phi(\mathbf{u})\frac{1}{2} b^2 \mathbf{u}^\top H_g(\mathbf{z}) \mathbf{u}d\mathbf{u}  + o(\|b\mathbf{u}\|^2)\\
    &= \int \phi(\mathbf{u})\frac{1}{2} b^2 \sum_{i,j}u_i u_j H_{g,i,j}(\mathbf{z}) d\mathbf{u}  + o(b^2)\\
    &= \sum_{i} H_{g,i,i}(\mathbf{z}) \mu_{2,\phi} \frac{b^2}{2} + o(b^2)\\
    %https://bookdown.org/egarpor/NP-UC3M/kde-ii-asymp.html
    &=\frac{b^2}{2}  \mu_{2,\phi} tr(H_g(\mathbf{z})) + o(b^2)\\
    \implies |\mathbb{E}[\hat{g}(\mathbf{z})] - g(\mathbf{z})| &\leq C_{\phi,b} b^2
\end{align}
for some constant $C_{\phi,b}$.

For the variance, 
\begin{align}
    Var(\hat{g}(z)) &= Var\Bigg(\frac{1}{mb^d} \sum_{i=1}^m \phi\Bigg(\frac{\mathbf{z} - Z_i}{b}\Bigg)\Bigg) = \frac{1}{mb^{2d}} Var\Bigg(\phi\Bigg(\frac{\mathbf{z} - Z}{b}\Bigg)\Bigg) \\
    &\leq  \frac{1}{mb^{2d}} \mathbb{E}\Bigg[\phi^2\Bigg(\frac{\mathbf{z} - Z}{b}\Bigg)\Bigg] = \frac{1}{mb^{2d}} \int \phi^2\Bigg(\frac{\mathbf{z} - \mathbf{z}'}{b}\Bigg) g(\mathbf{z}')d\mathbf{z}'\\
    &= \frac{1}{mb^{d}} \int \phi^2(\mathbf{u}) g(\mathbf{z}+b\mathbf{u})d\mathbf{u}\\
    &= \frac{1}{mb^{d}}\int  \phi^2(\mathbf{u}) \Big(
    g(\mathbf{z}) + b(D_g(\mathbf{z}))^\top \mathbf{u} + o(\|b\mathbf{u}\|^1)
    \Big)d\mathbf{u}\\
    &= \frac{1}{mb^{d}}\int  \phi^2(\mathbf{u}) g(\mathbf{z}) d\mathbf{u} + o(\frac{1}{mb^d})\\
    &=\frac{1}{mb^{d}}g(\mathbf{z}) \int \phi^2(\mathbf{u})d\mathbf{u} + o(\frac{1}{mb^d}) \leq C_{\phi,v}\frac{1}{mb^d}.
\end{align}
for some constant $C_{\phi,v}$.

As a result, for any $\mathbf{z}\in\mathcal{Z}$, we have the MSE as:
\begin{align}
    \mathbb{E}[\|\hat{g}(\mathbf{z}) - g(\mathbf{z})\|^2] &= \mathbb{E}[\|\hat{g}(\mathbf{z}) - \mathbb{E}[\hat{g}(\mathbf{z})] + \mathbb{E}[\hat{g}(\mathbf{z})] - g(\mathbf{z})\|^2]\\
    &= \mathbb{E}[\|\hat{g}(\mathbf{z}) - \mathbb{E}[\hat{g}(\mathbf{z})]\|^2] + \mathbb{E}[\|\mathbb{E}[\hat{g}(\mathbf{z})] - g(\mathbf{z})\|^2]\\%Since Cov(\hat{g}(z) - \mathbb{E}[\hat{g}(z)], \mathbb{E}[\hat{g}(z)] - g(z)) =0
    &= \underbrace{Var(\hat{g}(\mathbf{z}))}_{variance} + \underbrace{(\mathbb{E}[\hat{g}(\mathbf{z})] - g(\mathbf{z}))^2}_{bias^2} \\
    &\leq C_{\phi,v}\frac{1}{mb^d} + C_{\phi,b}b^4. \label{eq:biasvariance:final}
    %&= \Theta\Big(\frac{1}{mb^d}\Big) + \Theta(b^4)\label{eq:biasvariance:bigtheta}
\end{align}
This means the MSE goes to $0$ as long as $b^dm\to \infty$ and $b\to 0$. As $m\to\infty$, we have $\lim_{m\to\infty} \mathbb{E}[\|\hat{g}(\mathbf{z}) - g(\mathbf{z})\|^2] = 0$.

Now, note that $\hat{g}(\mathbf{z}) = \frac{1}{m}\sum_{j=1}^m V_j$ where $V_j = \frac{1}{b^d}\phi(\frac{\mathbf{z} - Z_j}{b})$.
By Bernstein's inequality, we have
\begin{align}
    \mathbf{P}\{|\hat{g}(\mathbf{z}) - \mathbb{E}[\hat{g}(\mathbf{z})]| > \epsilon \}\leq 2e^{-\frac{(m\epsilon^2)/2}{mC_{\phi,v}b^{-d}+\frac{1}{3}m\epsilon\phi(0)b^{-d}}} \leq e^{-B mb^d \epsilon^2}
\end{align}
for some constant $B$ as long as $\epsilon$ is smaller than a constant (say $1$). 
With triangular inequality, we have
\begin{align}
    \mathbf{P}\{|\hat{g}(\mathbf{z}) - g(\mathbf{z})| > \epsilon + C_{\phi,b} b^2 \} &\leq \mathbf{P}\{|\hat{g}(\mathbf{z}) - \mathbb{E}[\hat{g}(\mathbf{z})]| > \epsilon \} \leq e^{-B mb^d \epsilon^2} \label{eq:biasvariance:finalbound}
\end{align}
which gives us the conclusion as the RHS goes to $0$ as $m\to\infty$. 

%The convergence in probability follows from Markov's inequality:
%\begin{align}
%    \mathbb{P}\{\|\hat{g}(\mathbf{z}) - g(\mathbf{z})\|^2 > \epsilon \} \leq \frac{\mathbb{E}[\|\hat{g}(\mathbf{z}) - g(\mathbf{z})\|^2]}{\epsilon } \leq \frac{C_{1}\frac{1}{mb^d} + C_{2}b^4}{\epsilon }
%    \label{eq:biasvariance:markov}
%\end{align}

\subsection{Proof of Lemma~\ref{lemma:optimal}}
Lemma~\ref{lemma:optimal} says that $b=\Theta(m^{-\frac{1}{d+4}})$ is the optimal shrinkage rate to minimize $\mathbb{E}[\|\hat{g}(\mathbf{z}) - g(\mathbf{z})\|^2]$. 
Following Eq.~(\ref{eq:biasvariance:final}), by letting $C_{\phi,b}\frac{1}{mb^d} = C_{\phi,v}b^4$, we get $b = \Theta(m^{-\frac{1}{d+4}})$.
We can also derive this formula by taking the derivative of Eq.~(\ref{eq:biasvariance:final}) with respect to $b$ and setting it to $0$, which gives us (asymptotically):
\begin{align}
    \frac{-dC_{\phi,b}}{m}b^{-(d+1)} + 4C_{\phi,v}b^3 = 0 \Rightarrow b^* = C' m^{-\frac{1}{d+4}}
\end{align}
for some constant $C'$. 
The optimal MSE is thus $O(m^{-\frac{4}{d+4}})$.

%new finite sample
\subsection{Proof of Theorem~\ref{thm:main_weak}}\label{appendix:sec:proof:main_weak}
Denote $m\defeq \min_{k}\{m_k\}$. 
$\forall k\in [K]$, Bernstein's inequality\footnote{
One could apply Bennett's inequality to get: %https://en.wikipedia.org/wiki/Bennett%27s_inequality
%\begin{align}
%    \mathbb{P}\{N (\hat{\pi}_k - \pi_k) \geq \epsilon'_1 \} &\leq e^{-\frac{N\pi_k(1-\pi_k)}{a^2}h(\frac{a\epsilon'_1}{N\pi_k(1-\pi_k)})}
%\end{align}
%where $h(u) \defeq (1+u)\log{(1+u)} - u$, for $a$ such that  $\mathbbm{1}\{Y = k\} - \pi_k < a$ almost surely. 
%Applying this in both directions and choose $a=1-\pi_k$ and $\pi_k$ respectively, we have
\begin{align}
    \mathbb{P}\{|\hat{\pi}_k - \pi_k| \geq \epsilon_1 \} &\leq  e^{-N\frac{\pi_k}{1-\pi_k}h(\frac{\epsilon_1}{\pi_k})} + e^{-N\frac{1-\pi_k}{\pi_k}h(\frac{\epsilon_1}{1-\pi_k})} \label{eq:finite:pi_bound:Bennett}
\end{align}
and repeat the following proof for a slightly tigher bound.
However, the notation is much more complicated.
} gives us:
\begin{align}
    \mathbb{P}\{|\hat{\pi}_k - \pi_k| \geq \epsilon_1 \} &\leq 2e^{-\frac{N \epsilon_1^2}{2v_{min} + \frac{2}{3}\epsilon_1}}\leq e^{-B_2 N \epsilon_1^2}
\end{align}
where $v_{min} = \min_{k}\{\pi_k (1-\pi_k)\}$ and some constant $B_2$ (we find the smallest such constant among all classes), {as long as $\epsilon_1$ is smaller than a constant (e.g. 1)}.

%https://www.stat.cmu.edu/~larry/=sml/densityestimation.pdf
From Eq.~\ref{eq:biasvariance:finalbound}, with $b=C' m^{\frac{-1}{d+4}}$, and let $\epsilon = m^\frac{-\lambda}{d+4}$ {for $\lambda \in (0,2)$}, we have, for some constants $C_1, B_1$:
\begin{align}
    %\mathbb{P}\{|\hat{g}(\mathbf{z}) - g(\mathbf{z})| > \epsilon_2 + C_1 C_3 m^{\frac{2}{d+4}}\} \leq e^{-B C_3^d m^{\frac{d}{d+4}}\epsilon_2^2}
    \mathbb{P}\{|\hat{g}(\mathbf{z}) - g(\mathbf{z})| > C_1 m^{\frac{-\lambda}{d+4}}\} \leq e^{-B_1 m^{\frac{4-2\lambda}{d+4}}}.
\end{align}

%Define $\delta_k\defeq \Big(e^{-N\frac{\pi_k}{1-\pi_k}h(\frac{\epsilon_1}{\pi_k})} + e^{-N\frac{1-\pi_k}{\pi_k}h(\frac{\epsilon_1}{1-\pi_k})} + e^{-C_2 m_k^{\frac{4-2\lambda}{d+4}}}\Big)$.
Define $\delta_k \defeq  e^{-B_1 m_k^{\frac{4-2\lambda}{d+4}}} + e^{-B_2 N \epsilon_1^2} \leq e^{-B_1 m^{\frac{4-2\lambda}{d+4}}} + e^{-B_2 N \epsilon_1^2} $.
With probability $\geq 1-\sum_{k} \delta_k$ (union bound), for all $k$:
\begin{align}
    |\hat{g}_k(\mathbf{z}) \hat{\pi}_k - g_k(\mathbf{z})\pi_k| &\leq |\hat{g}_k(\mathbf{z}) \hat{\pi}_k - g_k(\mathbf{z})\hat{\pi}_k| + |g_k(\mathbf{z}) \hat{\pi}_k - g_k(\mathbf{z})\pi_k| \\
    &\leq C_1 m_k^{\frac{-\lambda}{d+4}} + g_k(\mathbf{z})\epsilon_1. \label{eq:finite:eachgpi}
\end{align}

Denote $g^+(\mathbf{z})= \max_{k} g_k(\mathbf{z}), g^-(\mathbf{z})= \min_{k} g_k(\mathbf{z})$, and $\overline{g}(\mathbf{z}) = \sum_{k} g_k(\mathbf{z})\pi_k$.
Denote $\epsilon_{k,2}\defeq C_1 m_k^{\frac{-\lambda}{d+4}}$, Eq.~\ref{eq:finite:eachgpi} means:
\begin{align}
    \hat{p}_k(\mathbf{z}) &= \frac{\hat{g}_k(\mathbf{z})\hat{\pi}_k}{\sum_{k'} \hat{g}_{k'}(\mathbf{z}) \hat{\pi}_{k'}} 
    \geq \frac{g_k(\mathbf{z})\pi_k - \epsilon_{k,2} - g_k(\mathbf{z})\epsilon_1}{\overline{g}(\mathbf{z}) + \sum_{k'} [\epsilon_{k',2} + g_{k'}(\mathbf{z})\epsilon_1]}\\
    &= \frac{g_k(\mathbf{z})\pi_k - \epsilon_{k,2} - g_k(\mathbf{z})\epsilon_1}{\overline{g}(\mathbf{z})} 
    \frac{1}{1 + \frac{\sum_{k'} [\epsilon_{k',2} + g_{k'}(\mathbf{z})\epsilon_1]}{\overline{g}(\mathbf{z})}}\\
    &\geq \frac{g_k(\mathbf{z})\pi_k - \epsilon_{k,2} - g_k(\mathbf{z})\epsilon_1}{\overline{g}(\mathbf{z})} (1-\frac{\sum_{k'} [\epsilon_{k',2} + g_{k'}(\mathbf{z})\epsilon_1]}{\overline{g}(\mathbf{z})}) \\
    &\geq p_k(\mathbf{z})
    -\frac{\epsilon_{k,2} + g_k(\mathbf{z})\epsilon_1}{\overline{g}(\mathbf{z})}
    - \frac{\sum_{k'} [\epsilon_{k',2} + g_{k'}(\mathbf{z})\epsilon_1]}{\overline{g}(\mathbf{z})}\label{eq:finite:geq:end} %the thing before parenthesis in the previous line is \leq 1
\end{align}
Similarly,
\begin{align}
    \hat{p}_k(\mathbf{z}) &\leq \frac{g_k(\mathbf{z})\pi_k + \epsilon_{k,2} + g_k(\mathbf{z})\epsilon_1}{\overline{g}(\mathbf{z})} 
    \frac{1}{1 - \frac{\sum_{k'} [\epsilon_{k',2} + g_{k'}(\mathbf{z})\epsilon_1]}{\overline{g}(\mathbf{z})}} \label{eq:finite:leq:1/1-x} \\  
    &\leq (p_k(\mathbf{z}) +\frac{\epsilon_{k,2} + g_k(\mathbf{z})\epsilon_1}{\overline{g}(\mathbf{z})})(1+2\frac{\sum_{k'} [\epsilon_{k',2} + g_{k'}(\mathbf{z})\epsilon_1]}{\overline{g}(\mathbf{z})}) \label{eq:finite:leq:1+2x}\\
    &\leq p_k(\mathbf{z}) 
    + \frac{\epsilon_{k,2} + g_{k}(\mathbf{z})\epsilon_1}{\overline{g}(\mathbf{z})}
    + \frac{3\sum_{k'} [\epsilon_{k',2} + g_{k'}(\mathbf{z})\epsilon_1]}{\overline{g}(\mathbf{z})}
    \label{eq:finite:leq:end}
\end{align}
We can proceed from Eq.\ref{eq:finite:leq:1/1-x} to \ref{eq:finite:leq:1+2x} and from \ref{eq:finite:leq:1+2x} to \ref{eq:finite:leq:end} when {
$\frac{\sum_{k'} [\epsilon_{k',2} + g_{k'}(\mathbf{z})\epsilon_1]}{\overline{g}(\mathbf{z})} \leq 0.5$
, which is achievable for a large $m$ (the smallest $m_k$, thus $N$) given any $\mathbf{z}$ as long as $\overline{g}(\mathbf{z}) > 0$. 
}.

With Eq.~\ref{eq:finite:geq:end} and \ref{eq:finite:leq:end}, with probability $\geq 1- K (e^{-B_1 m^{\frac{4-2\lambda}{d+4}}} + e^{-B_2 N \epsilon_1^2})$:
\begin{align}
    |p_k(\mathbf{z}) - \hat{p}_k(\mathbf{z})| &\leq \frac{(3K+1)(\epsilon_{k,2} + g^+(\mathbf{z}) \epsilon_1)}{\overline{g}(\mathbf{z})}\\
    &=\frac{(3K+1)(C_1 m^{\frac{-\lambda}{d+4}} + g^+(\mathbf{z})\epsilon_1)}{\overline{g}(\mathbf{z})}  
\end{align}
If we let $\epsilon_1 = \Theta(m^{\frac{-\lambda}{d+4}})$ and merge the constants, we have, with probability $\geq 1-Ke^{-B m^{\frac{4-2\lambda}{d+4}}}$ (note that $N\geq Km$):
\begin{align}
    |p_k(\mathbf{z}) - \hat{p}_k(\mathbf{z})| &\leq (3K+1)C m^{\frac{-\lambda}{d+4}}\\
    \implies  |\mathbf{p}(\mathbf{z}) - \phat(\mathbf{z})|_1 &\leq K(3K+1)C m^{\frac{-\lambda}{d+4}}
\end{align}
with some constant $C$ and $B$ that depends on $\{g_k(\mathbf{z})\}_{k\in[K]}$. 

\subsection{Proof of Theorem~\ref{thm:main_strong}}
If we assume $g_k$ is $\alpha$-H\"older continuous for all $k$ %with constant $C_\alpha$ (sup over all $k$), 
then by Theorem 2 in~\cite{ICML2017_UniformConvergence}, there exists positive constant $C'$ independent of $b$ and $m$, such that the following holds with probability $\geq 1-\frac{1}{m_k}$
\begin{align}
    \sup_{\mathbf{z}} | \hat{g}_k(\mathbf{z}) - g_k(\mathbf{z})| < C'\Big( b^\alpha + \sqrt{\frac{\log{m_k}}{m_k b^d}}\Big). \label{eq:uniformbound}
\end{align}
Furthermore, we assume that all the densities are bounded from below (see, for example, Section 3 in~\cite{10.2307/43818918}).
{Denote $U\defeq \max_{k} \sup_{\mathbf{z}} g_k(\mathbf{z})$ and $L\defeq \min_{k}\inf_{\mathbf{z}} g_k(\mathbf{z})$.}

We could replace $\epsilon_{k,2}$ in the previous section with $\epsilon_{k,2} = C_1(b^\alpha + \sqrt{\frac{\log{m_k}}{m_k b^d}})$.
Following similar steps leading towards Eq.~\ref{eq:finite:geq:end} and Eq.~\ref{eq:finite:leq:end}, we have, with probability $\geq 1-K(\frac{1}{m} + e^{-B_2 N \epsilon_1^2})$, for any $\mathbf{z}$:
\begin{align}
    |\hat{p}_k(\mathbf{z}) - p_k(\mathbf{z})| &\leq  \frac{(3K+1)(\epsilon_{k,2} + g^+(\mathbf{z}) \epsilon_1)}{\overline{g}(\mathbf{z})}\\
    &\leq \frac{(3K+1)}{L} \Big( C_1(b^\alpha + \sqrt{\frac{\log{m}}{m b^d}}) + U\epsilon_1 \Big)
\end{align}
Note that we still need $\frac{\sum_{k'} [\epsilon_{k',2} + g_{k'}(\mathbf{z})\epsilon_1]}{\overline{g}(\mathbf{z})} \leq 0.5$, which is satisfied as $N$ increases because $g_k(\mathbf{z}) >= L$. 
Now, we let $b = \Theta((\frac{\log{m}}{m})^{\frac{1}{d+2\alpha}})$, and  $\epsilon_1 = \Theta((\frac{\log{m}}{m})^{\frac{\alpha}{d+2\alpha}})$, we have with probability 
$
\geq 1-K(m^{-1} + e^{-Bm^{\frac{d}{d+2\alpha}}(\log{m})^{\frac{2\alpha}{d+2\alpha}}}) 
= 1-K(m^{-1} + m^{-B\frac{2\alpha}{d+2\alpha}m^{\frac{d}{d+2\alpha}}})
$:
\begin{align}
    |\hat{p}_k(\mathbf{z}) - p_k(\mathbf{z})| &\leq (3K+1) C (\frac{\log{m}}{m})^{\frac{\alpha}{d+2\alpha}}.
\end{align}

%Let $b$ be an admissible bandwidth of order $\Theta((\frac{\log{m}}{m})^{\frac{1}{d}})$, which means the RHS goes to 0 as $m\to\infty$.
%Like in Section~\ref{appendix:sec:proof:main_weak}, $\forall \epsilon>0$, $\delta>0$, we could find a $M$ s.t. $m>M$ means with probability $\geq 1-\delta$,
%\begin{align}
%    \sup_{\mathbf{z}}\|\phat(\mathbf{z}) - \ptrue(\mathbf{z})\|_\infty < \epsilon.
%\end{align}
%(Probability is taken over the training samples.)

Finally, with probability $\geq 1-K(m^{-1} + m^{-B\frac{2\alpha}{d+2\alpha}m^{\frac{d}{d+2\alpha}}})$, for any $\mathbf{q}$ in $\Delta^{K-1}$:
\begin{align}
    \sup_{\mathbf{z}: \phat(\mathbf{z})=\mathbf{q}}|\mathbb{P}\{Y=k|\phat (\mathbf{z}) = \mathbf{q}\} - q_k| = \sup_{\mathbf{z}: \phat(\mathbf{z})=\mathbf{q}}|p_k(\mathbf{z}) - \hat{p}_k(\mathbf{z}) |  
    %\leq \sup_{\mathbf{z}}\|\phat(\mathbf{z}) - \ptrue(\mathbf{z})\|_\infty 
    \leq \sup_{\mathbf{z}}|p_k(\mathbf{z}) - \hat{p}_k(\mathbf{z}) |
    \leq (3K+1) C (\frac{\log{m}}{m})^{\frac{\alpha}{d+2\alpha}}.
\end{align}
%Since $\epsilon$ and $\delta$ are arbitrary, we are done.

%%%%%%%%%%%%%%%%%%%%%%%%%%%%%%%%%%%%%%%%%%%%%%%%%%%%%%%%%%%%%%%%%%%%%%%%%%%%%%%
%%%%%%%%%%%%%%%%%%%%%%%%%%%%%%%%%%%%%%%%%%%%%%%%%%%%%%%%%%%%%%%%%%%%%%%%%%%%%%%

\section{Theoretical Analysis of Equal-sized Stratified Sampling in Training}\label{appendix:sec:sampling}
We adopted equal-sized stratified sampling to facilitate efficient training. Here we provide some theoretical justification of this choice. 

After fixing a $x_0$ whose label $y_0$ is the prediction target, the problem is essentially estimating $\frac{\mu_k p_k}{\sum_{k'} \mu_{k'} p_{k'}}$ for all $k$, where $p_k$ denotes the frequency of class $k$ in the population\footnote{In our case, this population is the large training set.} and $\mu_k$ denotes $\mathbb{E}[\phi(X, x_0) | Y=k]$. 
Note that we know $p_k$, but not $\mu_k$, since $p_k$ is fixed for our training set, but $\mu_k$ depends on $x_0$ and $\proj$, which is what we are training.
%Furthermore, we have $\mu\in[0,1]$.
Suppose we can afford to use $M$ samples in total to make the prediction, the question is: How do we distribute these $M$ samples to different classes?

What sampling method to use will depend on many factors, although a stratified sampling strategy tends to be more efficient in sample size. 
The sampling method we use (sample the same number of samples for each class $k$) intuitively will improve the estimation quality of the rarer class.
Here, we will elaborate why we chose this sampling method, the assumptions behind it, and why it helps training.

Denoting $S_k = \mu_k p_k$ and $S_{-k} = \sum_{k'\neq k}\mu_{k'} p_{k'}$, we can apply Taylor expansion to get an approximation of the variance\footnote{Such a derivation could be found in \url{https://www.stat.cmu.edu/~hseltman/files/ratio.pdf}}:
\begin{align}
    Var(\frac{S_k}{S_{-k}+S_k}) &\approx \frac{1}{\mathbb{E}[S_{-k} + S_{k}]^2} Var(S_k) - 2\frac{\mathbb{E}[S_k]}{\mathbb{E}[S_{-k}+S_{k}]^3} Cov(S_k, S_{-k}+S_k) \\ &+ \frac{\mathbb{E}[S_k]^2}{\mathbb{E}[S_{-k}+S_{k}]^4} Var(S_{-k} + S_k)\label{eq:appendix:variance:1}
\end{align}
%\url{https://www.stat.cmu.edu/~hseltman/files/ratio.pdf}
If we perform stratified sampling of any kind, then $Cov(S_k, S_{-k}) =0$, and Eq.~(\ref{eq:appendix:variance:1}) becomes:
\begin{align}
    Var(\frac{S_k}{S_{-k}+S_k}) &\approx 
     \frac{1}{\mathbb{E}[S_{-k} + S_{k}]^2} Var(S_k) - 2\frac{\mathbb{E}[S_k]}{\mathbb{E}[S_{-k}+S_{k}]^3} Var(S_k) \\ &+ \frac{\mathbb{E}[S_k]^2}{\mathbb{E}[S_{-k}+S_{k}]^4} [Var(S_{-k}) + Var(S_k)]\\
     &=\frac{\mathbb{E}[S_k]^2}{\mathbb{E}[S_{-k} + S_{k}]^4}  \Bigg( 
      \Bigg[\frac{\mathbb{E}[S_{-k}]}{\mathbb{E}[S_k]}\Bigg]^2 Var(S_k) +   Var(S_{-k})
     \Bigg)\label{eq:appendix:variance:2}
\end{align}

To further analyze Eq.~(\ref{eq:appendix:variance:2}) and gain more intuition, we make the following assumptions:
\begin{itemize}
    \item For any $k'\neq y_0$, $\mu_{k'}$ has the same value denoted as $\mu_{-y_0}$ (and is smaller than $\mu_{y_0}$).
    Intuitively, this is like considering a one-vs-rest classification problem, and we are just saying data from the same class will look more similar according to our kernel.
    
    \item The standard deviation for a single observation is directly proportional to the mean. 
    Namely, for all $k$, $\displaystyle \frac{\sqrt{Var(\phi(X, x_0) | Y=k)}}{\mathbb{E}[\phi(X, x_0)|Y=k]} \equiv r$ for a fixed number $r$.
\end{itemize}
%assume the 
If we assign $m_k$ samples to estimate $\mu_k$ then we have $Var(S_k) = r^2\frac{\mathbb{E}[S_k]^2}{m_k}$ and $Var(S_{-k}) = r^2\frac{\mathbb{E}[S_{-k}]^2}{M-m_k}$, where $M=\sum_{k'=1}^K m_{k'}$ ($M\gg m_k$ when $K$ is large).
This transforms Eq.~(\ref{eq:appendix:variance:2}) into:
\begin{align}
    Var\Bigg(\frac{S_k}{S_{-k}+S_k}\Bigg) \approx \frac{\mathbb{E}[S_k]^2\mathbb{E}[S_{-k}]^2}{\mathbb{E}[S_{-k}+S_k]^4}r^2\Bigg(
    \frac{1}{m_k} + \frac{1}{M-m_k}\Bigg) = C \Bigg(\frac{1}{m_k} + \frac{1}{M-m_k}\Bigg)
\end{align}
where $C$ is a constant that does not depend on $m_k$. 

Without prior information, it is natural to assume $C$ is class-independent (or at least relatively constant across classes).
Now, if our goal is to minimize the average variance, by Cauchy-Schwarts inequality we have:
\begin{align}
    \sum_{k=1}^K \frac{1}{m_k} &\geq \frac{K^2}{M}\\
    \sum_{k=1}^K \frac{1}{M-m_k} &\geq \frac{K^2}{(K-1)M}
\end{align}
The equality in both cases is achieved if and only if $m_k\equiv \frac{M}{K}$ for all $k$.
This means, to minimize the average variance $\frac{C}{K}\sum_{k=1}^K (\frac{1}{m_k} +\frac{1}{M-m_k})$, we need to choose $m_k$ to be the same for all class $k$.

%It is worth noting that the base version of stratified sampling, where $m_k \propto p_k$, 
%
It is worth noting that the discussion above is about training (and how to get better estimation therein). 
This is not referring to errors of the final $\proj$.
Given enough time, different ways to sample data lead to similar performance. 

%This means the mean of $n$ samples is $\sigma_n^2 = \frac{\sigma^2}{n}$.
%TODO: \fontred{Argument about the denominator's variance is small using  \url{http://www.est.uc3m.es/icascos/eng/simulation_notes/variance-reduction-techniques.html}}

%If we look at the numerator's variance, and we just want to reduce the overall variance, then it is clear that the vanilla stratified sampling is the best choice, because by Cauchy-Schwartz inequality, we have:
%\begin{align}
%    \sum_{k} \frac{\sigma^2}{m_k}p_k^2 \geq \frac{(\sum_{k} p_k)^2}{\sum_{k}m_k}\sigma^2 = \frac{\sigma^2}{M}
%\end{align}
%where $M$ is the total number of samples drawn and the equality holds when $\frac{p_k}{m_k}$ is the same for all $k$.

%However, if our goal is to reduce the variance for every class, then we will be looking at a different ``average'' variance, expressed as:
%\begin{align}
%    \fontred{placeholder}
%\end{align}

%%%%%%%%%%%%%%%%%%%%%%%%%%%%%%%%%%%%%%%%%%%%%%%%%%%%%%%%%%%%%%%%%%%%%%%%%%%%%%%
%%%%%%%%%%%%%%%%%%%%%%%%%%%%%%%%%%%%%%%%%%%%%%%%%%%%%%%%%%%%%%%%%%%%%%%%%%%%%%%
\newpage

\section{Additional Experimental Details}\label{appendix:sec:exp:more}
\subsection{Datasets}
This section provides more detail on the healthcare datasets, which might be less familiar to readers.

\textbf{IIIC}~\citep{IIICData, IIICData2} is an electroencephalography (EEG) dataset from the Massachusetts General Hospital EEG Archive. 
It is collected for the purpose of automated ictal-interictal-injury-continuum (IIIC) detection/monitoring.
IIIC patterns include seizure and seizure-like patterns designated Lateralized Periodic Discharges (LPDs), Generalized
Periodic Discharges (GPDs), Lateralized Rhythmic Delta Activity (LRDA), and Generalized
Rhythmic Delta Activity (GRDA)\citep{IIICData2}.
%The EEG recordings consist of 16 channels, 
The training data has been enriched with ``label spreading''~\citep{IIICData2}, whereas the test (and calibration) data consists of only labels from medical experts. 
To improve stability (because IIIC labeling is a challenging task for even experts), any sample with less than 3 labels are dropped. 
The majority label is then used as the truth for the test and calibration ses.
For more details on how the data was collected and labeled, please refer to~\cite{IIICData,IIICData2}.

\textbf{ISRUC}~\citep{ISRUC} is a public polysomnographic (PSG) dataset for the sleep staging task. 
It has three groups of data, with the first group having the most data and most widely used.
The (group 1) dataset contains 100 subjects with one recording session per subject.
Every 30 second of the recording is considered an ``epoch'' and is rated independently by two human experts.
We use the label from the first expert as the gold label.
The five classes of ISRUC correspond to five different stages of sleep, including Rapid Eye Movement (REM), Non-REM Stage 1 (N1), Non-REM Stage 2 (N2), Non-REM Stage 3 (N3), and Wake (Wake).
%From N1 to N3, we have progressively deeper sleep.
For more details, please refer to~\cite{ISRUC}.

\textbf{PhysioNet Callenge 2017 (PN2017)}~\citep{Clifford2017AF2017, Goldberger2000PhysioBankSignals} is a public (upon request) electrocardiogram (ECG) dataset for ECG rhythm classification.
The ECG recordings are sampled at 300Hz.
The original dataset contains four classes: Normal sinus rhythm (N), Atrial Fibrillation (AF), Other cardiac rhythms (O) and Noise segment.
Among these patterns, AF is an abnormal heart rhythm, and is the ``important class''.
We used the same processing method as~\cite{Hong2019Mina}, which cuts one segment into several shorter segments with data augmentation during the training phase.

A summary of the classes can be found below in Table~\ref{appendix:tab:main:healthdata}.
\AppendixTabHealthData

\textbf{Data Licenses and Consent}:
\begin{itemize}
    \item ISRUC:
    We could not find the license. 
    Per~\cite{ISRUC}, “All patients referred were submitted to an initial briefing with the support of an informed consent document. The ethics committee of CHUC approved the use of the data of the referred patients as anonymous for the research purposes”.
    
    \item PN2017: The license is Open Data Commons Attribution License v1.0. 
    The dataset is donated by AliveCor.
    
    \item IIIC: 
    We could not find the license. 
    Per~\cite{IIICData} ``the local IRB waived the requirement for informed consent for this retrospective analysis of EEG data''.
    
    \item CIFAR-100/CIFAR-10: We could not find the license. 
    They are publicly available.
    
    \item SVHN: Under CC0: Public Domain license. It is publicly available.

\end{itemize}

\subsection{Baseline Implementation}
\begin{itemize}
    \item Temperature Scaling: 
    We used the github repository accompanying~\cite{ICML2017_Guo}, \url{https://github.com/gpleiss/temperature_scaling}.
    
    \item Dirichlet Calibration:
    We used the code at \url{https://github.com/dirichletcal/experiments_dnn}.%to PyTorch (which, as pointed out by the original paper, is really just one simple linear layer.)
    
    \item Focal Loss~\citep{NeurIPS2020_FocalCal}:
    We used the loss function and the gamma schedule provided in \url{https://github.com/torrvision/focal_calibration}, and replaced our CrossEntropy loss function in all experiments during training. 
    
    \item Mutual-information-maximization-based Binning (\baselineIMax): 
    We use the official github implementation \url{https://github.com/boschresearch/imax-calibration}. 
    To normalize and get valid probability vectors, we used softmax on the log-odds given by \baselineIMax. 
    
    \item Gaussian Process Calibration:
    We use the official github implementation \url{https://github.com/JonathanWenger/pycalib}.
    
    \item Splines-based Calibration:
    We use the official github implementation \url{https://github.com/kartikgupta-at-anu/spline-calibration}.
    
    \item Intra Order-preserving Calibration:
    We use the official github implementation  \url{https://github.com/AmirooR/IntraOrderPreservingCalibration}.

    \item MMCE:
    We use the official github implementation  \url{https://github.com/aviralkumar2907/MMCE} with additional temperature scaling on the calibration set as suggested in the original paper.
    
\end{itemize}

\subsection{Training Details}
For CIFAR-10, CIFAR-100, SVHN, and ISRUC, the models are trained for 50 epochs, using a one-cycle Cosine scheduler with 3 warm-up and 10 cool-down epochs (the other parameters are default in \texttt{timm}).
The exact ViT and Mixer are \texttt{vit\_base\_patch16\_224\_in21k} and \texttt{mixer\_b16\_224\_in21k} implemented and pretrained by \texttt{timm} .
For PN2017, the number of epochs is 100, and we use a \texttt{ReduceLROnPlateau} scheduler that halves the learning rate with the patience parameter set to 10 epochs. 
We use a batch size of 128, SGD optimizer and weight decay rate of 1e-4.
For IIIC dataset, we use a AdamW optimizer with a weight decay rate of 1e-5, and no scheduler.
The learning rates are 2e-4 for CIFAR-10, CIFAR-100 and SVHN, 5e-3 for ISRUC, 1e-2 for PN2017 and 1e-3 for IIIC.
For all datasets except for IIIC, we used \texttt{LabelSmoothingCrossEntropy} in \texttt{timm} with smoothing being 0.1. 
For IIIC, since the original dataset contains pseudo-labels that form a distribution, we use a cross entropy loss.
The experiments for the \baselineFocal baseline replace all loss functions with the proposed focal loss.

To train $\proj$, we use an SGD optimizer with a learning rate of 4e-4 for CIFAR-10, CIFAR-100, SVHN and IIIC, 1e-3 for ISRUC and PN2017.
We use \texttt{ReduceLROnPlateau} scheduler that halves the learning rate with the patience parameter set to 10 epochs, and trains for 100 epochs.
Each epoch has a fixed number of 5000 batches (regardless of the size of the training set) and each batch consists of $B=64$ prediction samples and a ``background'' set used to construct KDE with $m_k\equiv m=20$ for all $k$.
The exact details could be found in our code.
Training time for the largest dataset (except for ImageNet), SVHN, is 3 hours for the base neural network, and 1 hour for $\proj$ on a machine with Nvidia RTX 3090 GPU. 
Inference time is much shorter.
%==============================================================
\subsection{Additional Evaluation Metrics}
In this section, we compute the following variants of the evaluation metrics presented in the main text.
The conclusion stays very similar across all methods. 
\begin{itemize}
    \item The static (equal-width bins) version of CECE, in Table~\ref{tab:main:cecet}.
    \item The static (equal-width bins) version of ECE, in Table~\ref{tab:main:ece}.
    \item The multi-class version of Brier score, in Table~\ref{tab:main:brier:mult}. 
    To be specific, the brier score in the main text is $\frac{1}{N}\sum_{i=1}^N (\phat_{k_i^*}(x_i) - \mathbbm{1}\{y_i=k_i^*\})^2$ where $k_i^* = \argmax_{k} \phat_k(x_i)$.
    The multi-class version of Brier score is $\frac{1}{NK}\sum_{i=1}^N \sum_{k=1}^K (\phat_k(x_i) - \mathbbm{1}\{y_i=k\})^2$.

    \item NLL Loss, in Table~\ref{tab:main:nll:torch}.
\end{itemize}

\def \TableExpMainCECEt{
\begin{table}[ht]
\caption{(Static) Class-wise ECE in $10^{-2}$ ($\downarrow$ means lower=better).
The best accuracy-preserving method is in \textbf{bold} (p=0.01).
The otherwise lowest number is \underline{underscored}. 
\methodname almost always achieves the lowest class-wise ECE, while maintaining accuracy.
}
\label{tab:main:cecet}
%\begin{center}
\centering
\begin{small}
\scalebox{0.69}{
%\begin{tabular}{lc|ccccccc|r}
\begin{tabular}{p{0.14\textwidth}p{0.1\textwidth}|p{0.1\textwidth}p{0.1\textwidth}p{0.1\textwidth}p{0.1\textwidth}p{0.1\textwidth}p{0.1\textwidth}p{0.1\textwidth}p{0.1\textwidth}|p{0.1\textwidth}}
\toprule
  CECE $\downarrow$  & \baselineUnCal & \baselineTS & \baselineDirCal & \baselineIMax & \baselineFocal & \baselineSpline & \baselineIOP & \baselineGP & \baselineMMCE & \methodname\\ 
\midrule
IIIC (pat) & 8.01$\pm$0.27 & 8.94$\pm$0.86 & \textbf{5.11$\pm$1.49} & 9.17$\pm$0.99 & 8.95$\pm$0.52 & 8.55$\pm$0.63 & 8.30$\pm$0.53 & 7.94$\pm$0.65 & 7.09$\pm$0.44 & \textbf{4.66$\pm$1.30}\\
IIIC & 7.89$\pm$0.02 & 8.96$\pm$0.50 & \textbf{2.13$\pm$0.13} & 8.77$\pm$0.24 & 8.76$\pm$0.02 & 8.41$\pm$0.23 & 7.97$\pm$0.26 & 7.51$\pm$0.24 & 6.66$\pm$0.26 & \textbf{2.04$\pm$0.27}\\
ISRUC (pat) & \textbf{4.51$\pm$0.25} & \textbf{4.68$\pm$0.77} & \textbf{4.19$\pm$0.89} & 8.65$\pm$0.99 & 9.24$\pm$0.20 & \textbf{4.67$\pm$0.46} & \textbf{4.59$\pm$0.59} & \textbf{4.63$\pm$0.42} & \textbf{4.06$\pm$0.35} & \textbf{3.84$\pm$1.22}\\
ISRUC & 4.53$\pm$0.02 & 5.18$\pm$0.79 & 2.73$\pm$0.38 & 9.29$\pm$0.86 & 9.07$\pm$0.02 & 4.75$\pm$0.16 & 4.69$\pm$0.37 & 4.71$\pm$0.25 & 4.07$\pm$0.21 & \textbf{1.93$\pm$0.27}\\
PN2017 & 12.20$\pm$0.07 & 12.32$\pm$0.19 & \textbf{4.04$\pm$0.54} & 9.70$\pm$1.19 & 16.70$\pm$0.10 & 8.42$\pm$0.73 & 12.10$\pm$0.37 & 12.20$\pm$0.07 & 12.20$\pm$0.32 & \textbf{3.83$\pm$1.27}\\
C10 (ViT) & 3.42$\pm$0.01 & 1.39$\pm$0.08 & 1.25$\pm$0.08 & \textbf{1.15$\pm$0.06} & 5.19$\pm$0.03 & 1.36$\pm$0.06 & 1.25$\pm$0.07 & 1.23$\pm$0.06 & 1.52$\pm$0.22 & \textbf{1.18$\pm$0.08}\\
C10 (Mixer) & 3.36$\pm$0.02 & 2.11$\pm$0.11 & 1.64$\pm$0.08 & \textbf{1.76$\pm$0.24} & 7.02$\pm$0.03 & 1.71$\pm$0.09 & 1.78$\pm$0.10 & 1.75$\pm$0.10 & 1.95$\pm$0.27 & \textbf{1.59$\pm$0.06}\\
C100 (ViT) & 6.33$\pm$0.05 & 6.43$\pm$0.29 & 5.44$\pm$0.14 & 5.96$\pm$0.21 & 6.07$\pm$0.05 & \textbf{5.16$\pm$0.17} & 5.58$\pm$0.14 & 5.54$\pm$0.09 & 5.30$\pm$0.22 & \textbf{5.06$\pm$0.11}\\
C100 (Mixer) & 5.60$\pm$0.05 & 6.75$\pm$0.25 & 5.87$\pm$0.20 & 6.64$\pm$0.29 & 6.08$\pm$0.06 & 5.56$\pm$0.13 & 6.09$\pm$0.32 & 5.80$\pm$0.14 & 6.15$\pm$0.21 & \textbf{5.16$\pm$0.07}\\
SVHN (ViT) & 3.50$\pm$0.01 & 2.56$\pm$0.58 & \textbf{1.40$\pm$0.06} & 2.98$\pm$0.22 & 6.11$\pm$0.02 & 1.51$\pm$0.07 & \textbf{1.47$\pm$0.07} & 1.51$\pm$0.05 & 1.63$\pm$0.11 & \textbf{1.46$\pm$0.08}\\
SVHN (Mixer) & 3.36$\pm$0.02 & 3.38$\pm$0.67 & \textbf{1.39$\pm$0.11} & 3.00$\pm$0.16 & 5.79$\pm$0.02 & 1.66$\pm$0.07 & 1.54$\pm$0.06 & 1.58$\pm$0.06 & 1.73$\pm$0.09 & 1.57$\pm$0.11\\
ImageNet & 3.70$\pm$0.03 & 3.99$\pm$0.07 & 6.11$\pm$0.22 & 3.29$\pm$0.21 & \textendash & \textbf{2.80$\pm$0.07} & \textbf{2.93$\pm$0.16} & 3.05$\pm$0.08 & \textendash & \underline{2.40$\pm$0.04}\\
\bottomrule
\end{tabular}
}
\end{small}
\end{table}
}
\TableExpMainCECEt

\def \TableExpMainECE{
\begin{table}[ht]
\caption{(Static) ECE in $10^{-2}$ ($\downarrow$ means lower=better).
%All numbers are presented in percentage for readability
The best accuracy-preserving method is in \textbf{bold} (p=0.01).
%The otherwise lowest number is \underline{underscored}. 
\methodname is usually on par or better than the best baseline.
}
\label{tab:main:ece}
%\begin{center}
\centering
\begin{small}
\scalebox{0.67}{
%\begin{tabular}{lc|ccccccc|r}
\begin{tabular}{lc|cccccccc|c}
%\begin{tabular} {p{0.14\textwidth}p{0.1\textwidth}|p{0.1\textwidth}p{0.1\textwidth}p{0.1\textwidth}p{0.1\textwidth}p{0.1\textwidth}p{0.1\textwidth}p{0.1\textwidth}p{0.1\textwidth}|p{0.1\textwidth}}
\toprule
 ECE $\downarrow$  & \baselineUnCal & \baselineTS & \baselineDirCal & \baselineIMax & \baselineFocal & \baselineSpline & \baselineIOP & \baselineGP & \baselineMMCE & \methodname\\ 
\midrule
IIIC (pat) & 9.18$\pm$1.08 & \textbf{4.95$\pm$2.77} & \textbf{2.87$\pm$1.62} & 10.56$\pm$4.05 & 7.37$\pm$0.53 & \textbf{4.54$\pm$2.07} & \textbf{4.56$\pm$2.15} & \textbf{3.84$\pm$1.63} & 6.34$\pm$3.30 & \textbf{4.28$\pm$1.42}\\
IIIC & 9.13$\pm$0.04 & 4.42$\pm$1.53 & \textbf{1.22$\pm$0.17} & 10.17$\pm$0.81 & 7.10$\pm$0.04 & 3.08$\pm$0.65 & 3.44$\pm$0.38 & 1.68$\pm$0.55 & 4.78$\pm$2.26 & 2.55$\pm$0.61\\
ISRUC (pat) & 3.60$\pm$0.32 & \textbf{2.70$\pm$1.56} & 2.91$\pm$1.02 & 8.82$\pm$1.41 & 14.95$\pm$0.40 & \textbf{1.99$\pm$0.36} & \textbf{2.40$\pm$1.43} & \textbf{1.94$\pm$0.62} & \textbf{2.09$\pm$0.97} & \textbf{2.74$\pm$1.29}\\
ISRUC & 3.46$\pm$0.06 & 3.81$\pm$1.67 & 2.20$\pm$0.68 & 9.58$\pm$1.26 & 14.76$\pm$0.05 & \textbf{1.48$\pm$0.55} & 2.69$\pm$0.94 & 2.04$\pm$0.76 & \textbf{2.08$\pm$1.06} & \textbf{1.34$\pm$0.41}\\
PN2017 & 17.10$\pm$0.14 & 17.34$\pm$0.42 & \textbf{5.46$\pm$0.66} & 8.97$\pm$1.85 & 24.65$\pm$0.13 & \textbf{6.10$\pm$2.22} & 16.55$\pm$2.03 & 17.13$\pm$0.15 & 13.21$\pm$1.08 & \textbf{4.56$\pm$1.41}\\
C10 (ViT) & 9.17$\pm$0.05 & 0.76$\pm$0.11 & 0.44$\pm$0.08 & 0.61$\pm$0.06 & 7.19$\pm$0.06 & 0.49$\pm$0.10 & 0.38$\pm$0.05 & \textbf{0.28$\pm$0.07} & 0.65$\pm$0.15 & 0.41$\pm$0.10\\
C10 (Mixer) & 9.06$\pm$0.05 & 1.11$\pm$0.12 & 0.51$\pm$0.05 & 1.04$\pm$0.17 & 12.54$\pm$0.06 & 0.48$\pm$0.08 & 0.56$\pm$0.12 & \textbf{0.34$\pm$0.06} & 1.01$\pm$0.40 & 0.65$\pm$0.09\\
C100 (ViT) & 11.65$\pm$0.14 & 2.81$\pm$0.44 & \textbf{0.77$\pm$0.12} & 3.39$\pm$0.23 & 9.98$\pm$0.09 & 1.07$\pm$0.24 & 1.24$\pm$0.27 & 0.92$\pm$0.12 & 1.21$\pm$0.36 & 1.58$\pm$0.33\\
C100 (Mixer) & 13.71$\pm$0.15 & 3.18$\pm$0.35 & 1.17$\pm$0.26 & 4.82$\pm$0.25 & 14.36$\pm$0.20 & \textbf{1.20$\pm$0.35} & 1.82$\pm$0.72 & \textbf{1.15$\pm$0.22} & 2.14$\pm$0.49 & 3.11$\pm$0.48\\
SVHN (ViT) & 10.11$\pm$0.05 & \textbf{2.44$\pm$2.72} & \textbf{0.61$\pm$0.09} & 2.08$\pm$0.18 & 12.17$\pm$0.08 & \textbf{0.64$\pm$0.14} & \textbf{0.55$\pm$0.11} & \textbf{0.61$\pm$0.10} & \textbf{0.66$\pm$0.15} & 0.71$\pm$0.13\\
SVHN (Mixer) & 10.30$\pm$0.04 & 3.19$\pm$2.55 & \textbf{0.57$\pm$0.08} & 2.21$\pm$0.10 & 11.09$\pm$0.06 & 0.67$\pm$0.13 & \textbf{0.49$\pm$0.10} & 0.62$\pm$0.08 & 0.69$\pm$0.21 & 0.74$\pm$0.11\\
ImageNet & 3.06$\pm$0.13 & 3.26$\pm$0.13 & 4.26$\pm$0.74 & 8.05$\pm$0.32 & \textendash & 1.13$\pm$0.15 & 1.38$\pm$0.46 & \textbf{0.95$\pm$0.16} & \textendash & 1.30$\pm$0.28\\
\bottomrule
\end{tabular}
}
\end{small}
%\end{center}
\end{table}
}
\TableExpMainECE

\def \TableExpMainBrierMult{
\begin{table}[ht]
\caption{
Brier Score (multi-class) in $10^{-2}$ ($\downarrow$ means lower=better).
The best accuracy-preserving methods are in \textbf{bold} (p=0.01).
}
\label{tab:main:brier:mult}
\centering
\begin{small}
\scalebox{0.69}{
%\begin{tabular}{lc|ccccccc|r}
\begin{tabular}{p{0.14\textwidth}p{0.1\textwidth}|p{0.1\textwidth}p{0.1\textwidth}p{0.1\textwidth}p{0.1\textwidth}p{0.1\textwidth}p{0.1\textwidth}p{0.1\textwidth}p{0.1\textwidth}|p{0.1\textwidth}}
\toprule
 Brier $\downarrow$ & \baselineUnCal & \baselineTS & \baselineDirCal & \baselineIMax & \baselineFocal & \baselineSpline & \baselineIOP & \baselineGP & \baselineMMCE & \methodname\\ 
\midrule
IIIC (pat) & 9.23$\pm$0.18 & 9.11$\pm$0.31 & \textbf{8.13$\pm$0.26} & 9.22$\pm$0.44 & 9.69$\pm$0.20 & 9.05$\pm$0.25 & 9.07$\pm$0.27 & 9.01$\pm$0.24 & 9.13$\pm$0.25 & \textbf{8.38$\pm$0.36}\\
IIIC & 9.25$\pm$0.01 & 9.10$\pm$0.06 & 7.86$\pm$0.02 & 9.17$\pm$0.05 & 9.68$\pm$0.00 & 9.00$\pm$0.01 & 9.05$\pm$0.03 & 8.95$\pm$0.03 & 9.07$\pm$0.06 & \textbf{7.40$\pm$0.04}\\
ISRUC (pat) & 6.84$\pm$0.17 & 6.83$\pm$0.17 & 6.83$\pm$0.21 & 7.15$\pm$0.21 & 7.97$\pm$0.14 & 6.79$\pm$0.17 & 6.82$\pm$0.18 & 6.80$\pm$0.17 & \textbf{6.59$\pm$0.14} & \textbf{6.67$\pm$0.18}\\
ISRUC & 6.95$\pm$0.02 & 6.97$\pm$0.06 & 6.66$\pm$0.04 & 7.31$\pm$0.11 & 8.07$\pm$0.01 & 6.90$\pm$0.02 & 6.94$\pm$0.04 & 6.90$\pm$0.03 & 6.68$\pm$0.02 & \textbf{6.30$\pm$0.03}\\
PN2017 & 14.92$\pm$0.02 & 14.97$\pm$0.11 & \textbf{12.85$\pm$0.09} & 14.03$\pm$0.20 & 17.64$\pm$0.01 & 13.78$\pm$0.13 & 14.84$\pm$0.24 & 14.92$\pm$0.02 & 15.26$\pm$0.17 & \textbf{12.81$\pm$0.13}\\
C10 (ViT) & 0.27$\pm$0.01 & 0.18$\pm$0.01 & \textbf{0.16$\pm$0.01} & 0.17$\pm$0.01 & 0.31$\pm$0.01 & \textbf{0.16$\pm$0.01} & \textbf{0.16$\pm$0.01} & \textbf{0.16$\pm$0.01} & 0.18$\pm$0.01 & \textbf{0.15$\pm$0.01}\\
C10 (Mixer) & 0.39$\pm$0.01 & 0.30$\pm$0.01 & 0.30$\pm$0.01 & 0.30$\pm$0.02 & 0.74$\pm$0.01 & \textbf{0.29$\pm$0.01} & \textbf{0.29$\pm$0.01} & \textbf{0.28$\pm$0.01} & \textbf{0.30$\pm$0.03} & \textbf{0.28$\pm$0.01}\\
C100 (ViT) & 0.14$\pm$0.00 & 0.12$\pm$0.00 & 0.12$\pm$0.00 & 0.12$\pm$0.00 & 0.14$\pm$0.00 & 0.12$\pm$0.00 & 0.12$\pm$0.00 & 0.12$\pm$0.00 & \textbf{0.11$\pm$0.00} & 0.11$\pm$0.00\\
C100 (Mixer) & 0.21$\pm$0.00 & 0.19$\pm$0.00 & 0.18$\pm$0.00 & 0.19$\pm$0.00 & 0.23$\pm$0.00 & 0.18$\pm$0.00 & 0.18$\pm$0.00 & 0.18$\pm$0.00 & \textbf{0.17$\pm$0.00} & 0.18$\pm$0.00\\
SVHN (ViT) & 0.76$\pm$0.01 & 0.65$\pm$0.04 & 0.62$\pm$0.01 & 0.65$\pm$0.01 & 0.95$\pm$0.01 & 0.62$\pm$0.01 & 0.62$\pm$0.01 & 0.62$\pm$0.01 & \textbf{0.54$\pm$0.00} & 0.55$\pm$0.01\\
SVHN (Mixer) & 0.77$\pm$0.01 & 0.68$\pm$0.05 & 0.62$\pm$0.01 & 0.66$\pm$0.01 & 0.97$\pm$0.01 & 0.63$\pm$0.01 & 0.63$\pm$0.01 & 0.63$\pm$0.01 & 0.68$\pm$0.01 & \textbf{0.60$\pm$0.01}\\
ImageNet & 0.03$\pm$0.00 & 0.03$\pm$0.00 & 0.03$\pm$0.00 & 0.03$\pm$0.00 & \textendash & \textbf{0.03$\pm$0.00} & \textbf{0.03$\pm$0.00} & \textbf{0.03$\pm$0.00} & \textendash & 0.03$\pm$0.00\\
\bottomrule
\end{tabular}
}
\end{small}
\end{table}
}
\TableExpMainBrierMult

\def \TableExpAppendixNLL{
\begin{table}[ht]
\caption{
NLL ($\downarrow$ means lower=better).
The best accuracy-preserving methods are in \textbf{bold} (p=0.01).
}
\label{tab:main:nll:torch}
\centering
\begin{small}
\scalebox{0.69}{
%\begin{tabular}{lc|ccccccc|r}
\begin{tabular}{p{0.14\textwidth}p{0.1\textwidth}|p{0.1\textwidth}p{0.1\textwidth}p{0.1\textwidth}p{0.1\textwidth}p{0.1\textwidth}p{0.1\textwidth}p{0.1\textwidth}p{0.1\textwidth}|p{0.1\textwidth}}
\toprule
 NLL $\downarrow$ & \baselineUnCal & \baselineTS & \baselineDirCal & \baselineIMax & \baselineFocal & \baselineSpline & \baselineIOP & \baselineGP & \baselineMMCE & \methodname\\ 
\midrule
IIIC (pat) & 1.09$\pm$0.02 & 1.08$\pm$0.05 & \textbf{0.97$\pm$0.05} & 1.11$\pm$0.07 & 1.11$\pm$0.03 & 1.07$\pm$0.03 & 1.07$\pm$0.04 & 1.06$\pm$0.03 & 1.07$\pm$0.03 & \textbf{1.00$\pm$0.05}\\
IIIC & 1.09$\pm$0.00 & 1.08$\pm$0.01 & 0.92$\pm$0.00 & 1.10$\pm$0.01 & 1.11$\pm$0.00 & 1.06$\pm$0.00 & 1.06$\pm$0.00 & 1.05$\pm$0.00 & 1.06$\pm$0.01 & \textbf{0.87$\pm$0.01}\\
ISRUC (pat) & 0.63$\pm$0.02 & 0.62$\pm$0.02 & \textbf{0.62$\pm$0.02} & 0.69$\pm$0.03 & 0.72$\pm$0.01 & \textbf{0.61$\pm$0.02} & 0.62$\pm$0.02 & \textbf{0.61$\pm$0.02} & \textbf{0.60$\pm$0.02} & \textbf{0.61$\pm$0.02}\\
ISRUC & 0.64$\pm$0.00 & 0.63$\pm$0.01 & 0.60$\pm$0.00 & 0.71$\pm$0.02 & 0.73$\pm$0.00 & 0.63$\pm$0.00 & 0.63$\pm$0.00 & 0.62$\pm$0.00 & 0.61$\pm$0.00 & \textbf{0.57$\pm$0.00}\\
PN2017 & 1.00$\pm$0.00 & 1.00$\pm$0.00 & \textbf{0.86$\pm$0.01} & 0.96$\pm$0.02 & 1.19$\pm$0.00 & 0.95$\pm$0.01 & 0.99$\pm$0.02 & 1.00$\pm$0.00 & 1.04$\pm$0.03 & \textbf{0.86$\pm$0.01}\\
C10 (ViT) & 0.12$\pm$0.00 & 0.05$\pm$0.01 & \textbf{0.03$\pm$0.00} & 0.04$\pm$0.00 & 0.10$\pm$0.00 & 0.04$\pm$0.00 & \textbf{0.03$\pm$0.00} & \textbf{0.03$\pm$0.00} & 0.05$\pm$0.00 & \textbf{0.03$\pm$0.00}\\
C10 (Mixer) & 0.15$\pm$0.00 & 0.07$\pm$0.01 & 0.06$\pm$0.00 & 0.07$\pm$0.00 & 0.20$\pm$0.00 & \textbf{0.06$\pm$0.00} & 0.06$\pm$0.00 & \textbf{0.06$\pm$0.00} & \textbf{0.07$\pm$0.02} & \textbf{0.06$\pm$0.00}\\
C100 (ViT) & 0.38$\pm$0.00 & 0.29$\pm$0.01 & 0.28$\pm$0.01 & 0.32$\pm$0.01 & 0.36$\pm$0.00 & 0.28$\pm$0.01 & 0.28$\pm$0.01 & 0.27$\pm$0.01 & \textbf{0.25$\pm$0.01} & 0.27$\pm$0.00\\
C100 (Mixer) & 0.54$\pm$0.01 & 0.43$\pm$0.01 & 0.43$\pm$0.01 & 0.47$\pm$0.01 & 0.54$\pm$0.01 & 0.43$\pm$0.01 & 0.43$\pm$0.01 & 0.42$\pm$0.01 & \textbf{0.39$\pm$0.01} & 0.44$\pm$0.01\\
SVHN (ViT) & 0.23$\pm$0.00 & 0.16$\pm$0.02 & 0.15$\pm$0.00 & 0.17$\pm$0.00 & 0.26$\pm$0.00 & 0.15$\pm$0.00 & 0.15$\pm$0.00 & 0.15$\pm$0.00 & \textbf{0.13$\pm$0.00} & 0.13$\pm$0.00\\
SVHN (Mixer) & 0.23$\pm$0.00 & 0.19$\pm$0.03 & 0.15$\pm$0.00 & 0.18$\pm$0.00 & 0.26$\pm$0.00 & 0.16$\pm$0.00 & 0.15$\pm$0.00 & 0.15$\pm$0.00 & 0.16$\pm$0.00 & \textbf{0.15$\pm$0.00}\\
ImageNet & 0.84$\pm$0.01 & 0.83$\pm$0.01 & 0.90$\pm$0.02 & 0.87$\pm$0.02 & \textendash & 0.77$\pm$0.01 & \textbf{0.75$\pm$0.01} & \textbf{0.75$\pm$0.01} & \textendash & 0.87$\pm$0.01\\
\bottomrule
\end{tabular}
}
\end{small}
\end{table}
}
\TableExpAppendixNLL

%==============================================================
\clearpage
\subsection{Reliability Diagrams}
Figure~\ref{fig:appendix:reliability:iiic}, \ref{fig:appendix:reliability:isruc}, and \ref{fig:appendix:reliability:ecg} are the reliability diagrams for the IIIC, ISRUC and PN2017 dataset, respectively.
We keep only bins with at least 15 samples, because otherwise the ``gap'' is  misleading due to small sample and big variance. 
The count of samples in each bin is plotted on the right axis (log-scale). 
The conclusion is similar.
In all cases, \baselineTS seems to calibrate the overall ECE but fails on some classes.
\baselineDirCal tends to improve on all classes, but \methodname usually closes the gap between actual frequency and the prediction further.

\def \AppendixFigIIICReliability{
\begin{figure}[H]
\begin{center}
\centerline{\includegraphics[width=1\columnwidth]{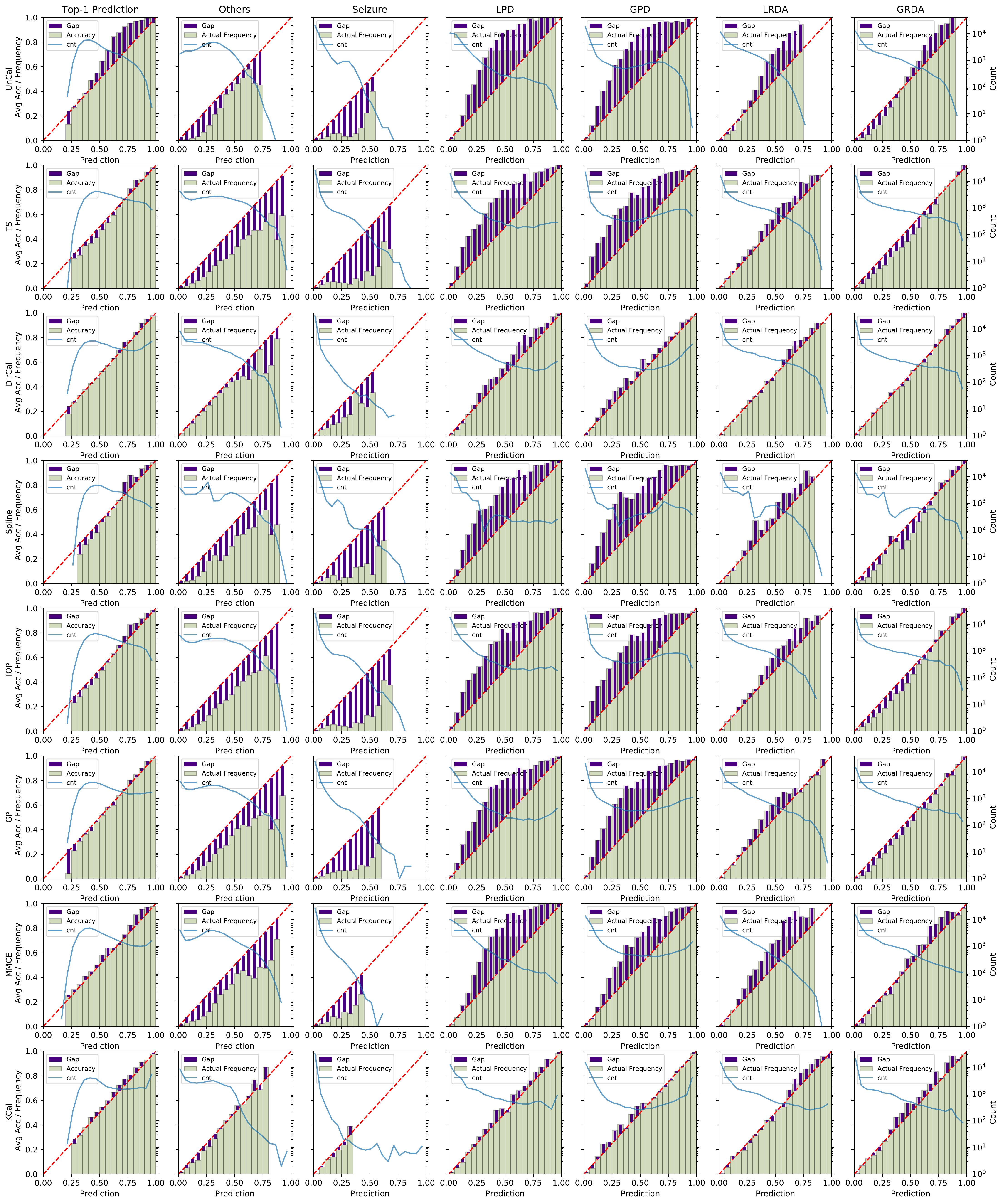}}
\caption{
Reliability diagrams for the IIIC dataset.
}
\label{fig:appendix:reliability:iiic}
\end{center}
\end{figure}
}
\def \AppendixFigISRUCReliability{
\begin{figure}[H]
\begin{center}
\centerline{\includegraphics[width=1\columnwidth]{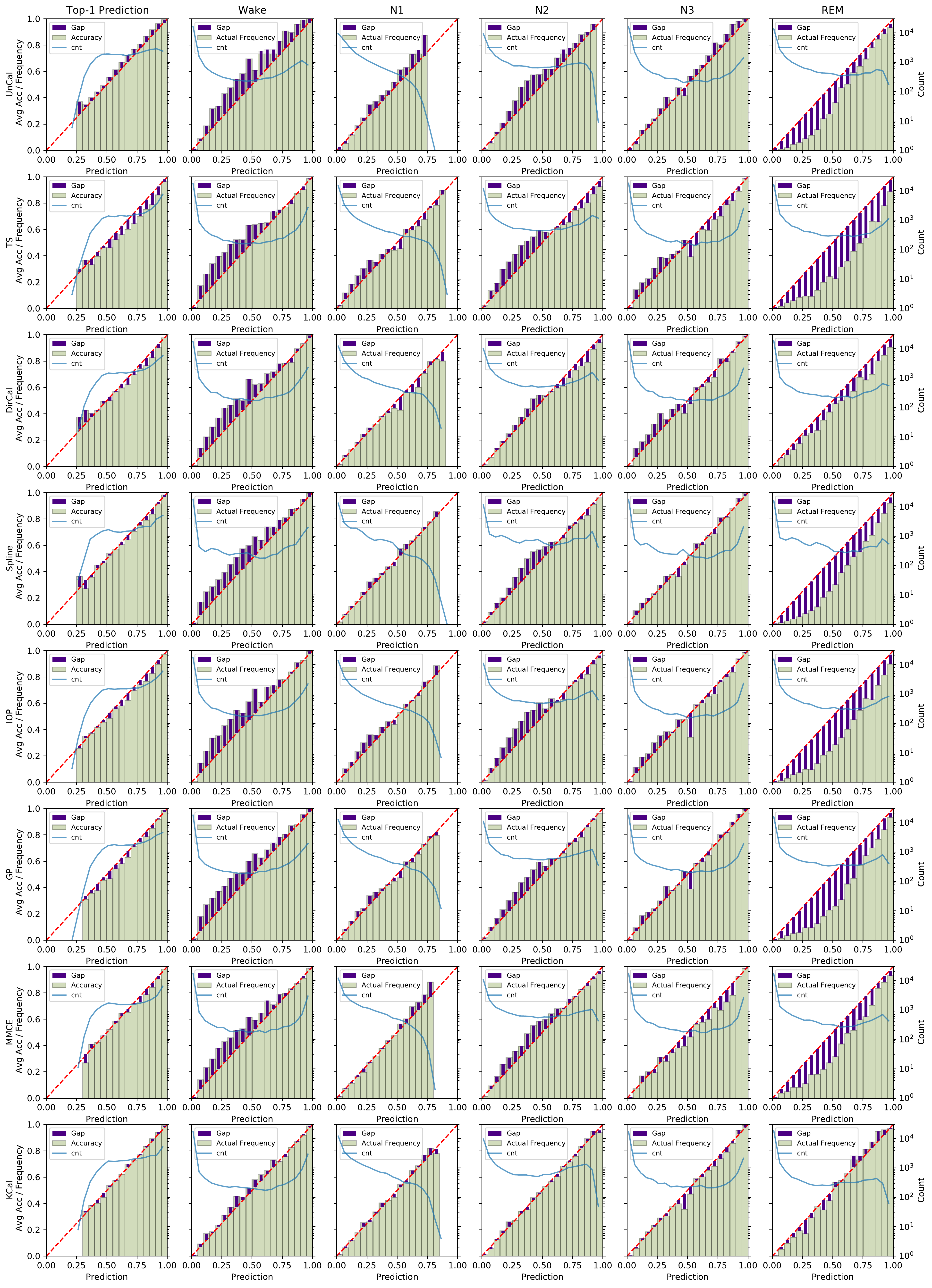}}
\caption{
Reliability diagrams for the ISRUC dataset.
}
\label{fig:appendix:reliability:isruc}
\end{center}
\end{figure}
}

\def \AppendixFigECGReliability{
\begin{figure}[H]
\begin{center}
\centerline{\includegraphics[width=1\columnwidth]{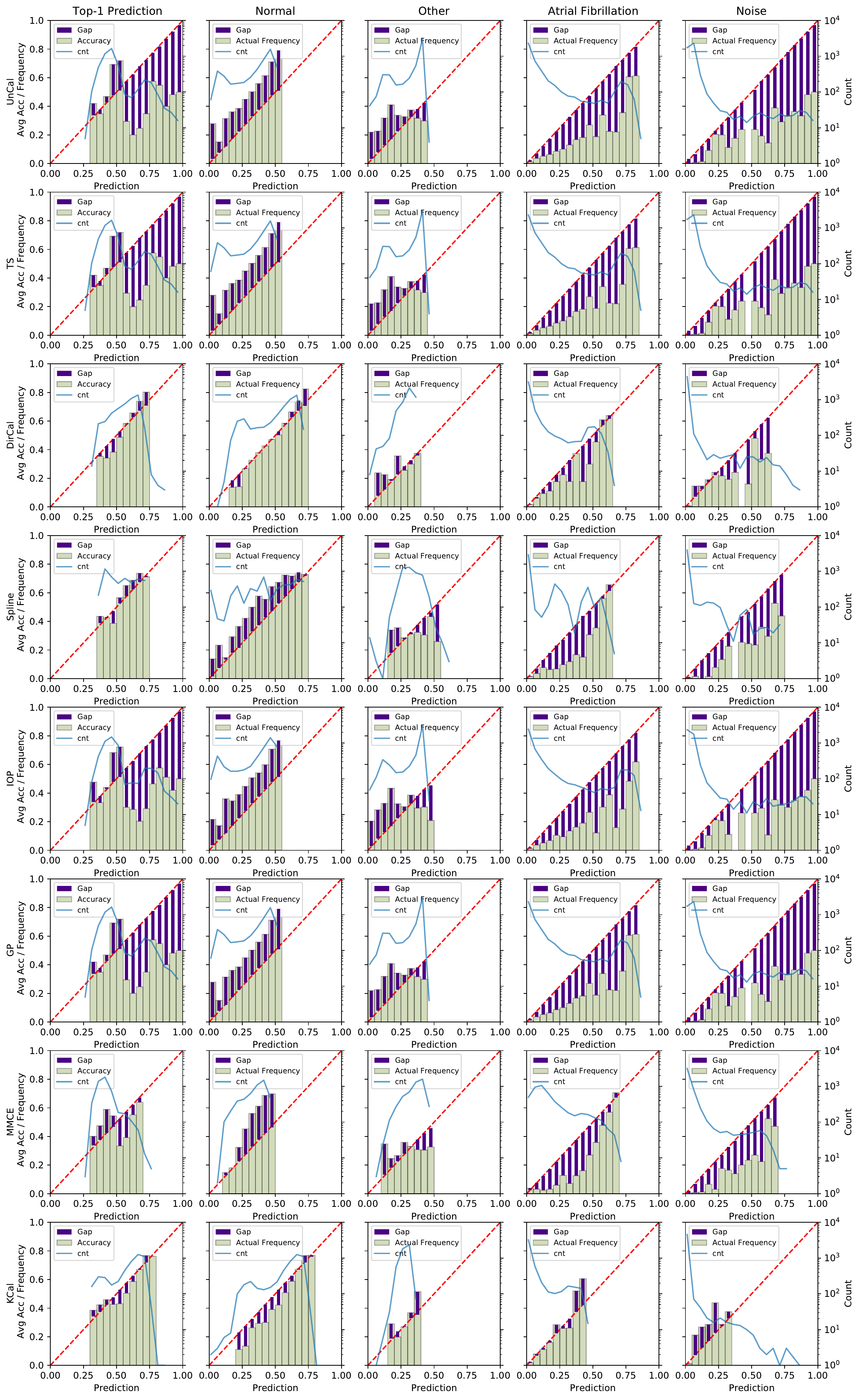}}
\caption{
Reliability diagrams for the PN2017 dataset.
}
\label{fig:appendix:reliability:ecg}
\end{center}
\end{figure}
}
\AppendixFigIIICReliability
\AppendixFigISRUCReliability
\AppendixFigECGReliability

\section{Ablation Study: Linear Projection}\label{appendix:sec:2layer_vs_linear}
A natural first architecture to try for $\proj$ is a simple linear layer. 
It is however not clear whether a linear projection can learn the best metric space due to its simplicity. 
We introduced a mild complexity by having two layers in $\proj$, yet the skip connection should help it learn well when a linear projection is the most desirable as well.
\KernelArchitecture 
We empirically compared both versions: \methodname, with the architecture showed in Figure~\ref{fig:mlp2}, and \methodname-Linear, which only uses one linear layer with the same output dimension ($d$). 
Both $\proj$ normalized $\embedder(\cdot)$ automatically with a Batch Normalization layer. 
The results are in Table~\ref{tab:appendix:linear}. 
As we can see, \methodname is generally better than the linear version, but the gap is generally small.
The additional computation time is smaller than 1x the computation time for \methodname-Linear, because the second layer has only $d^2$ parameters rather than $hd$ in the first layer ($h>d$). 
Both have negligible computation overhead compared with calling $\embedder$ (see Appendix~\ref{appendix:sec:dim}).

\def \AppendixTabLinear{

\begin{table*}[ht]
\caption{
Comparison between the architecture described in Figure~\ref{fig:mlp2} (\methodname) and a simple linear projection with the same input and output dimensions (\methodname-Linear).
On average, \methodname adapts to different datasets and architectures better than (\methodname-Linear), although the performance is generally similar. 
}
\label{tab:appendix:linear}
\begin{center}
\begin{small}
\scalebox{0.8}{
\begin{tabular}{l|cc|cc|cc|cc}
\toprule
   & \multicolumn{2}{c}{Accuracy$\uparrow$}& \multicolumn{2}{c}{CECE$\downarrow$}& \multicolumn{2}{c}{ECE$\downarrow$}&\multicolumn{2}{c}{Brier $\downarrow$}\\
   & \methodname & \methodname-Lienar & \methodname & \methodname-Lienar& \methodname & \methodname-Lienar & \methodname & \methodname-Lienar\\
\midrule
IIIC(pat) & \textbf{61.67$\pm$2.22} & 61.51$\pm$2.46 & \textbf{4.68$\pm$1.27} & 4.68$\pm$1.41 & \textbf{4.34$\pm$1.35} & 4.48$\pm$1.99 & 19.33$\pm$0.78 & \textbf{19.28$\pm$0.82}\\
IIIC & \textbf{66.32$\pm$0.21} & 65.59$\pm$0.20 & \textbf{2.03$\pm$0.26} & 2.08$\pm$0.23 & \textbf{2.62$\pm$0.59} & 3.12$\pm$0.68 & \textbf{17.54$\pm$0.10} & 17.88$\pm$0.09\\
ISRUC(pat) & \textbf{76.13$\pm$0.89} & 76.02$\pm$1.08 & \textbf{3.82$\pm$1.24} & 3.96$\pm$1.34 & \textbf{2.78$\pm$1.25} & 2.87$\pm$1.53 & \textbf{14.97$\pm$0.29} & 15.04$\pm$0.30\\
ISRUC & \textbf{77.45$\pm$0.16} & 77.19$\pm$0.19 & \textbf{1.90$\pm$0.28} & 2.01$\pm$0.31 & \textbf{1.36$\pm$0.41} & 1.69$\pm$0.46 & \textbf{14.28$\pm$0.08} & 14.37$\pm$0.08\\
PN2017 & \textbf{60.36$\pm$0.61} & 60.15$\pm$0.56 & 4.25$\pm$1.26 & \textbf{4.21$\pm$1.26} & \textbf{4.78$\pm$1.48} & 5.41$\pm$1.14 & \textbf{22.56$\pm$0.28} & 22.69$\pm$0.32\\
C10 (ViT) & \textbf{98.98$\pm$0.09} & 98.96$\pm$0.07 & 0.74$\pm$0.07 & \textbf{0.72$\pm$0.06} & 0.40$\pm$0.05 & \textbf{0.31$\pm$0.06} & \textbf{0.75$\pm$0.05} & 0.75$\pm$0.05\\
C10 (Mixer) & \textbf{98.14$\pm$0.06} & 98.12$\pm$0.09 & \textbf{1.17$\pm$0.10} & 1.18$\pm$0.07 & \textbf{0.59$\pm$0.09} & 0.61$\pm$0.13 & \textbf{1.34$\pm$0.04} & 1.34$\pm$0.05\\
C100 (ViT) & 92.37$\pm$0.15 & \textbf{92.47$\pm$0.14} & \textbf{4.32$\pm$0.10} & 4.37$\pm$0.08 & 1.50$\pm$0.32 & \textbf{1.43$\pm$0.33} & 5.01$\pm$0.08 & \textbf{4.93$\pm$0.08}\\
C100 (Mixer) & 87.55$\pm$0.16 & \textbf{88.00$\pm$0.24} & \textbf{4.62$\pm$0.10} & 4.73$\pm$0.12 & 3.07$\pm$0.49 & \textbf{2.78$\pm$0.45} & 7.61$\pm$0.09 & \textbf{7.39$\pm$0.07}\\
SVHN (ViT) & \textbf{96.42$\pm$0.05} & 96.36$\pm$0.06 & \textbf{1.23$\pm$0.10} & 1.32$\pm$0.08 & \textbf{0.64$\pm$0.12} & 0.65$\pm$0.09 & \textbf{2.49$\pm$0.03} & 2.49$\pm$0.03\\
SVHN (Mixer) & 96.10$\pm$0.04 & \textbf{96.13$\pm$0.04} & \textbf{1.40$\pm$0.08} & 1.49$\pm$0.08 & 0.73$\pm$0.10 & \textbf{0.61$\pm$0.09} & \textbf{2.68$\pm$0.03} & 2.69$\pm$0.03\\
%ImageNet & \textbf{79.64$\pm$0.24} & 79.55$\pm$0.28 & \textbf{1.94$\pm$0.04} & 2.05$\pm$0.05& 1.43$\pm$0.34 & \textbf{1.35$\pm$0.35}& \textbf{11.14$\pm$0.10} & 11.16$\pm$0.12 \\
\bottomrule
\end{tabular}
}
\end{small}
\end{center}
\end{table*}
}
\AppendixTabLinear

\section{Ablation Study: Using the classification logits}\label{appendix:sec:embed_vs_logits}
We empirically compared using the penultimate-layer embeddings and the predicted logits in Table~\ref{tab:appendix:logits}.
As we can see, \methodname is generally better than the alternative that uses the logits.

\def \AppendixTabLogit{

\begin{table*}[ht]
\caption{
Comparison between using the penultimate layer embedding vs the prediction logits as the input to $\proj$ (\methodname-Logits).
Overall, \methodname is significantly better than \methodname-Logits, but \methodname-Logits also has competitive performance. 
}
\label{tab:appendix:logits}
\begin{center}
\begin{small}
\scalebox{0.8}{
\begin{tabular}{l|cc|cc|cc|cc}
\toprule
   & \multicolumn{2}{c}{Accuracy$\uparrow$}& \multicolumn{2}{c}{CECE$\downarrow$}& \multicolumn{2}{c}{ECE$\downarrow$}&\multicolumn{2}{c}{Brier $\downarrow$}\\
   & \methodname & \methodname-Logits & \methodname & \methodname-Logits& \methodname & \methodname-Logits & \methodname & \methodname-Logits\\
\midrule
IIIC(pat) & \textbf{61.67$\pm$2.22} & 61.21$\pm$2.66 & 4.68$\pm$1.27 & \textbf{4.26$\pm$1.30} & 4.34$\pm$1.35 & \textbf{4.02$\pm$1.51} & 19.33$\pm$0.78 & \textbf{19.07$\pm$0.77}\\
IIIC & \textbf{66.32$\pm$0.21} & 65.26$\pm$0.20 & \textbf{2.03$\pm$0.26} & 2.11$\pm$0.27 & \textbf{2.62$\pm$0.59} & 2.77$\pm$0.37 & \textbf{17.54$\pm$0.10} & 17.90$\pm$0.05\\
ISRUC(pat) & \textbf{76.13$\pm$0.89} & 75.57$\pm$1.02 & \textbf{3.82$\pm$1.24} & 3.95$\pm$1.44 & 2.78$\pm$1.25 & \textbf{2.75$\pm$1.27} & \textbf{14.97$\pm$0.29} & 15.30$\pm$0.31\\
ISRUC & \textbf{77.45$\pm$0.16} & 76.75$\pm$0.12 & \textbf{1.90$\pm$0.28} & 1.97$\pm$0.32 & \textbf{1.36$\pm$0.41} & 1.62$\pm$0.48 & \textbf{14.28$\pm$0.08} & 14.60$\pm$0.09\\
PN2017 & \textbf{60.36$\pm$0.61} & 59.99$\pm$0.56 & 4.25$\pm$1.26 & \textbf{4.13$\pm$1.22} & \textbf{4.78$\pm$1.48} & 5.18$\pm$0.96 & \textbf{22.56$\pm$0.28} & 22.64$\pm$0.34\\
C10 (ViT) & \textbf{98.98$\pm$0.09} & 98.94$\pm$0.06 & \textbf{0.74$\pm$0.07} & 0.79$\pm$0.07 & \textbf{0.40$\pm$0.05} & 0.43$\pm$0.05 & \textbf{0.75$\pm$0.05} & 0.79$\pm$0.04\\
C10 (Mixer) & \textbf{98.14$\pm$0.06} & 98.11$\pm$0.06 & \textbf{1.17$\pm$0.10} & 1.21$\pm$0.07 & 0.59$\pm$0.09 & \textbf{0.54$\pm$0.06} & \textbf{1.34$\pm$0.04} & 1.37$\pm$0.04\\
C100 (ViT) & \textbf{92.37$\pm$0.15} & 91.11$\pm$0.14 & \textbf{4.32$\pm$0.10} & 4.67$\pm$0.10 & \textbf{1.50$\pm$0.32} & 1.95$\pm$0.37 & \textbf{5.01$\pm$0.08} & 5.55$\pm$0.08\\
C100 (Mixer) & \textbf{87.55$\pm$0.16} & 85.07$\pm$0.26 & \textbf{4.62$\pm$0.10} & 4.98$\pm$0.13 & \textbf{3.07$\pm$0.49} & 3.73$\pm$0.54 & \textbf{7.61$\pm$0.09} & 8.84$\pm$0.06\\
SVHN (ViT) & \textbf{96.42$\pm$0.05} & 96.05$\pm$0.05 & \textbf{1.23$\pm$0.10} & 1.53$\pm$0.12 & \textbf{0.64$\pm$0.12} & 0.91$\pm$0.08 & \textbf{2.49$\pm$0.03} & 2.76$\pm$0.03\\
SVHN (Mixer) & \textbf{96.10$\pm$0.04} & 95.90$\pm$0.05 & \textbf{1.40$\pm$0.08} & 1.65$\pm$0.11 & \textbf{0.73$\pm$0.10} & 0.88$\pm$0.09 & \textbf{2.68$\pm$0.03} & 2.84$\pm$0.03\\
%ImageNet & \textbf{79.64$\pm$0.24} & 79.20$\pm$0.25 & \textbf{1.94$\pm$0.04} & 2.97$\pm$0.03 & \textbf{1.43$\pm$0.34} & 2.08$\pm$0.32 & \textbf{11.14$\pm$0.10} & 11.21$\pm$0.09 \\
\bottomrule
\end{tabular}
}
\end{small}
\end{center}
\end{table*}
}
\AppendixTabLogit

\def \AppendixTabRawKernel{

\begin{table*}[ht]
\caption{
\fontred{Placeholder}
}
\label{tab:appendix:untrained}
\begin{center}
\begin{small}
\scalebox{0.8}{
\begin{tabular}{l|cc|cc|cc|cc}
\toprule
   & \multicolumn{2}{c}{Accuracy$\uparrow$}& \multicolumn{2}{c}{CECE$\downarrow$}& \multicolumn{2}{c}{ECE$\downarrow$}&\multicolumn{2}{c}{Brier $\downarrow$}\\
   & \methodname & Untrained & \methodname & Untrained& \methodname & Untrained & \methodname & Untrained\\
\midrule
IIIC(pat) & \textbf{58.13$\pm$3.20} & 57.51$\pm$3.97 & \textbf{5.53$\pm$1.35} & 8.92$\pm$0.96 & \textbf{4.48$\pm$1.75} & 16.59$\pm$4.33 & \textbf{19.80$\pm$1.26} & 22.69$\pm$1.63\\
IIIC & \textbf{69.31$\pm$0.36} & 62.36$\pm$0.79 & \textbf{3.41$\pm$0.23} & 8.63$\pm$0.15 & \textbf{5.94$\pm$0.61} & 20.99$\pm$1.09 & \textbf{17.09$\pm$0.14} & 23.96$\pm$0.38\\
ISRUC(pat) & 74.23$\pm$0.89 & \textbf{75.55$\pm$1.17} & \textbf{5.01$\pm$1.77} & 7.10$\pm$1.12 & \textbf{2.92$\pm$1.72} & 11.67$\pm$1.43 & \textbf{15.92$\pm$0.46} & 16.80$\pm$0.46\\
ISRUC & \textbf{77.68$\pm$0.18} & 75.93$\pm$0.21 & \textbf{3.23$\pm$0.38} & 6.64$\pm$0.35 & \textbf{3.49$\pm$0.89} & 11.16$\pm$0.99 & \textbf{14.28$\pm$0.13} & 16.68$\pm$0.20\\
PN2017 & 57.59$\pm$0.66 & \textbf{58.80$\pm$0.98} & \textbf{5.12$\pm$0.68} & 5.17$\pm$0.59 & \textbf{7.65$\pm$1.26} & 8.18$\pm$1.03 & \textbf{22.41$\pm$0.27} & 22.70$\pm$0.15\\
C10 (ViT) & 98.91$\pm$0.05 & \textbf{98.94$\pm$0.05} & \textbf{1.01$\pm$0.14} & 2.95$\pm$0.07 & \textbf{0.27$\pm$0.07} & 6.47$\pm$0.12 & \textbf{0.83$\pm$0.03} & 1.24$\pm$0.03\\
C10 (Mixer) & 98.00$\pm$0.08 & \textbf{98.16$\pm$0.09} & \textbf{1.49$\pm$0.18} & 3.63$\pm$0.04 & \textbf{0.29$\pm$0.06} & 11.68$\pm$0.16 & \textbf{1.50$\pm$0.04} & 2.92$\pm$0.03\\
C100 (ViT) & \textbf{91.47$\pm$0.11} & 70.94$\pm$3.04 & \textbf{4.94$\pm$0.10} & 10.83$\pm$1.81 & \textbf{1.43$\pm$0.29} & 69.25$\pm$3.05 & \textbf{5.40$\pm$0.06} & 68.44$\pm$2.96\\
C100 (Mixer) & \textbf{85.78$\pm$0.24} & 65.40$\pm$4.54 & \textbf{5.42$\pm$0.12} & 12.85$\pm$1.47 & \textbf{2.64$\pm$0.37} & 63.73$\pm$4.56 & \textbf{8.49$\pm$0.11} & 63.11$\pm$4.41\\
SVHN (ViT) & 95.36$\pm$0.07 & \textbf{95.77$\pm$0.06} & \textbf{1.76$\pm$0.13} & 4.39$\pm$0.07 & \textbf{0.56$\pm$0.06} & 20.70$\pm$0.20 & \textbf{3.12$\pm$0.04} & 7.64$\pm$0.09\\
SVHN (Mixer) & 95.41$\pm$0.05 & \textbf{95.77$\pm$0.04} & \textbf{1.85$\pm$0.09} & 4.75$\pm$0.06 & \textbf{0.46$\pm$0.09} & 23.05$\pm$0.19 & \textbf{3.10$\pm$0.03} & 8.83$\pm$0.10\\
%ImageNet & \textbf{79.64$\pm$0.24} & 79.20$\pm$0.25 & \textbf{1.94$\pm$0.04} & 2.97$\pm$0.03 & \textbf{1.43$\pm$0.34} & 2.08$\pm$0.32 & \textbf{11.14$\pm$0.10} & 11.21$\pm$0.09 \\
\bottomrule
\end{tabular}
}
\end{small}
\end{center}
\end{table*}
}
%\AppendixTabRawKernel

%===============================================================
\newpage

\section{Ablation Study: Effect of $d$}\label{appendix:sec:dim}
To investigate the effect of $d$, we tried $d=$8, 16, 32, 64, and 128 and repeat the experiments.
The performance and the inference time (overhead) can be found in Figure~\ref{appendix:fig:dim}.
The inference time depends on the size of the calibration set, which is specified in Section~\ref{sec:exp}.

Generally speaking, we can only tell for sure that increasing $d$ increases the overhead, although the overhead is always small compared with calling $\embedder$.
The effect on other metrics, including accuracy, ECE and CEC, is not monotonic, and the best $d$ probably depends on many factors. 
\def \AppendixFigDim{
\begin{figure}[ht]
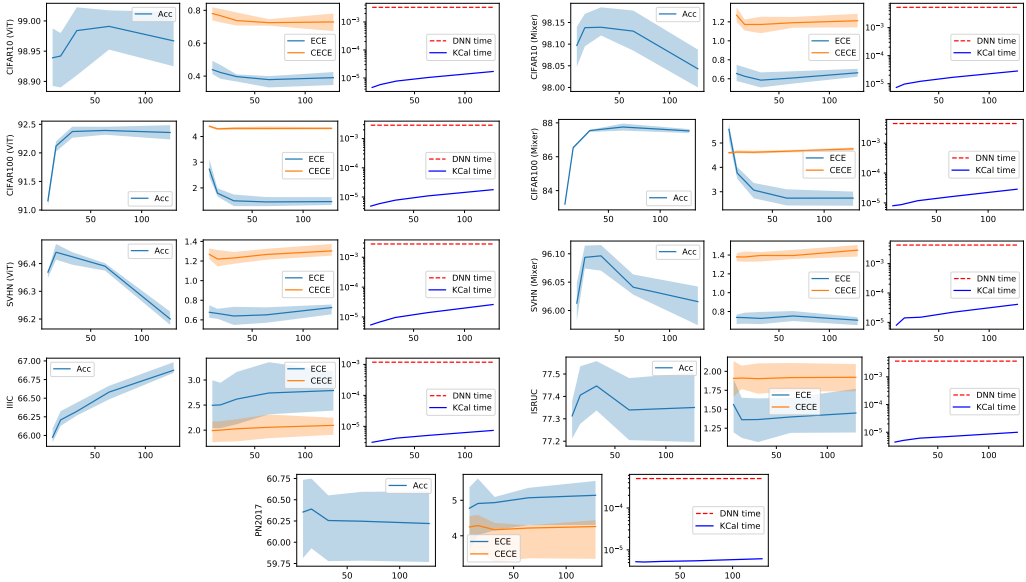

\vskip 0.2in
\begin{center}
\includegraphics[width=0.48\columnwidth]{Figures/Appendix/dim/appendix_dim_CIFAR10_ViT.pdf}\hspace{5pt}
\includegraphics[width=0.48\columnwidth]{Figures/Appendix/dim/appendix_dim_CIFAR10_Mixer.pdf}
\includegraphics[width=0.48\columnwidth]{Figures/Appendix/dim/appendix_dim_CIFAR100_ViT.pdf}\hspace{5pt}
\includegraphics[width=0.48\columnwidth]{Figures/Appendix/dim/appendix_dim_CIFAR100_Mixer.pdf}
\includegraphics[width=0.48\columnwidth]{Figures/Appendix/dim/appendix_dim_SVHN_ViT.pdf}\hspace{5pt}
\includegraphics[width=0.48\columnwidth]{Figures/Appendix/dim/appendix_dim_SVHN_Mixer.pdf}
\includegraphics[width=0.48\columnwidth]{Figures/Appendix/dim/appendix_dim_IIIC.pdf}\hspace{5pt}
\includegraphics[width=0.48\columnwidth]{Figures/Appendix/dim/appendix_dim_ISRUC.pdf}
\includegraphics[width=0.5\columnwidth]{Figures/Appendix/dim/appendix_dim_PN2017.pdf}
\caption{
Change in performance and inference time if we we change $d$ (the output embedding size of $\proj$. 
``DNN time'' refers to the average time running $\embedder$ for one input $x$, and ``\methodname time'' refers to the average time transforming $\embedder(x)$ to $\phat(x)$ using \methodname. 
For Accuracy, ECE and CECE, the unit is percentage. 
The band represents the median 50\% among 10 experiments.
For time, the unit is second.
Performance is not always improving as $d$ increases, but a larger $d$ naturally leads to larger overhead. 
It is however worth noting that in all experiment, the overhead (``\methodname time'') is negligible compared with the `DNN time''. 
}
\label{appendix:fig:dim}
\end{center}
\vskip -0.2in
\end{figure}
}

\AppendixFigDim

%===============================================================
\newpage
\section{Computing bandwidth}\label{appendix:sec:bandwidth}
As suggested in the main text, although there is a bandwidth selection step that seemingly prevents \methodname from efficiently updating predictions in an online manner, we could actually leverage Lemma~\ref{lemma:optimal} to compute $b$ as opposed to actually performing cross-validation.
To verify empirically that this is feasible in practice, we perform experiments where we vary the size of the calibration set, and plot the cross-validation-selected bandwidth $b$ against the predicted value $\Theta(m^{-\frac{1}{d+4}})$. 
The results are in Figure~\ref{fig:bandwidth}.
\AppendixFigBW

If everything is perfect, we should see a linear relation in all plots, and we can use this relationship to compute $b^*$ when we gradually add samples to the calibration set.
It is clear that if we use the estimated constant in the $\Theta(\cdot)$ and the calibration set size (per class) $m$ to set the bandwidth, we are still very close to the empirically selected value most of the time. 
In practice, this means that we only need to perform the actual cross validation occasionally, and predict the $b^*$ in between.
Note that from left to right, $m$ decreases, so the optimal $b^*$ increases and the variance increases greatly due to $m$ being small. 
In practice, one might keep updating $b^*$ using cross validation when $m$ is small (and cross-validation takes very little time) and only compute $b^*$ when $m$ is already large.

While computation will give good estimates for $b^*$ for most datasets, especially when $m$ is large and the estimate of $b^*$ is relatively stable (towards the left ends of plots), IIIC (and ISRUC to some extent) seems to show two different slopes. 
As $m$ increases, from right to left, $b^*$ seems to first decrease, and then stop decreasing. 
While a detailed analysis for this are beyond the scope of this paper, there are a few possible reasons.
\begin{enumerate}
    \item First, and most importantly, the optimal bandwidth derived in Lemma~\ref{lemma:optimal} is ``best'' for estimating the density, $f_k$ (in Eq.~(\ref{eq:kde:class_prob}), not $\mathbb{P}\{\cdot | X\}$. 
$b^*$ is however chosen according to the log-loss of the KDE classifier. 
As a result, the formula should be more relevant when $K$ is large and the difference between $\phat_k(X)$ and $\mathbb{P}\{Y=k| X\}$ is essentially linear in $\hat{f}_k - f_k$ (as the denominator is much more accurate than the numerator).
The experiment does support this point, since CIFAR100, with 100 classes, exhibits the clearest linear relationship. 

    \item Lemma~\ref{lemma:optimal} is not applicable if $f_{\proj\circ \embedder}$ violates the assumptions.
    For example, if $\embedder$ creates a discontinuity in the density, with a lot of data from different classes mapped to the same embedding. 
    This means decreasing $b$ might not decrease the bias term in Section~\ref{appendix:sec:proof:admissible}, and only increases variance. 
    This could be what is happening in CIFAR10-ViT (with 99\% accuracy) and in the left end of IIIC: decreasing $b$ might not improve log-loss as we have exhausted the discriminative power of $\embedder$. 
\end{enumerate}

\section{Bandwidth Selection}\label{appendix:sec:bandwidthsearch}
In Section~\ref{sec:method:impl}, we stated that we use Golden-Section search because we assume the cross entropy loss is convex in bandwidth $b$. 
While the convexity is expected from the bias-variance trade-off, we show in Figure~\ref{appendix:fig:bw_convexity} that this is indeed the case.
\AppendixFigBWConvexity

\section{Effect of $|\calibrationset|$}\label{appendix:sec:calsize}
In Figure~\ref{appendix:fig:size_of_scal}, we plot the accuracy, Brier score, CECE and ECE as a function of $|\calibrationset|$ for different datasets.
As expected, as $|\calibrationset|$ increases, the performance of \methodname increases and then stabilizes. 
\AppendixFigSizeSCal

\end{document}